\documentclass{article} 
\usepackage{iclr2027_conference,times}

\usepackage{amsmath,amsfonts,bm}

\def\eqref#1{equation~\ref{#1}}

\def\1{\bm{1}}

\DeclareMathAlphabet{\mathsfit}{\encodingdefault}{\sfdefault}{m}{sl}
\SetMathAlphabet{\mathsfit}{bold}{\encodingdefault}{\sfdefault}{bx}{n}

\usepackage{hyperref}
\usepackage{url}

\usepackage{microtype}
\usepackage{graphicx}
\usepackage{subcaption}
\usepackage{booktabs}

\usepackage{amsmath}
\usepackage{amssymb}
\usepackage{mathtools}
\usepackage{amsthm}

\usepackage[capitalize,noabbrev]{cleveref}

\theoremstyle{plain}

\theoremstyle{definition}

\theoremstyle{remark}

\usepackage{resizegather}
\usepackage{wrapfig}
\usepackage{tikz}
\usepackage{multirow}
\usepackage{color}
\usepackage{xcolor}
\usepackage{url}
\usepackage{xurl}
\usepackage{caption}
\usepackage{subcaption}
\usepackage{float}
\usepackage{balance}
\usepackage{graphicx}
\usepackage{textcomp}
\usepackage{booktabs}
\usepackage{colortbl}
\usepackage{enumerate}

\newcommand\shortsection[1]{\vspace{6pt}{\noindent\bf #1.}}

\title{No Free Efficiency: Revisiting the Trade-off Between Training Efficiency and Model Vulnerability}

\author{Yiyong Liu\textsuperscript{1}, Jun Sakuma\textsuperscript{2}, Michael Backes\textsuperscript{1}, Rui Wen\textsuperscript{2} \\
\textsuperscript{1}CISPA Helmholtz Center for Information Security \\
\textsuperscript{2}Institute of Science Tokyo 
}

\iclrfinalcopy
\begin{document}

\maketitle

\begin{abstract}
Training efficiency has become the central driver of recent progress in foundation models. To overcome the massive computational and data requirements of large-scale training, researchers increasingly adopt strategies such as selective data sampling, efficient pre-training, and simplified reinforcement learning pipelines. While these strategies drastically reduce overhead, they prompt a critical, yet neglected question: Is efficiency achieved at the expense of model robustness and security?
To our knowledge, we present the first systematic cross-domain investigation of the efficiency--vulnerability trade-off. Across vision and language models, we show that efficiency-oriented training increases susceptibility to adversarial and privacy attacks. We characterize this vulnerability by analyzing the models’ internal geometry and functional representations, demonstrating that the evaluated efficient variants consistently exhibit sharper loss geometry together with systematic changes in representational structure.
We further extend our analysis to ``zero RL training'', finding that models trained using simplified RL recipes exhibit substantially greater susceptibility to catastrophic forgetting and more pronounced overconfidence than those trained through conventional alignment pipelines. Our findings suggest that training efficiency is rarely a ``free lunch''; rather, the mechanisms that minimize computation can inadvertently compromise safety. We conclude by calling for a paradigm shift toward multi-objective training that jointly optimizes for performance, cost, and security.
\end{abstract}  

\section{Introduction}
\label{sec:intro}
In recent years, foundation models have reshaped the world at large, becoming defining technologies embedded in search engines and creative tools. However, their success remains tied to scale, requiring massive datasets and tens of millions of dollars in compute. 

This financial burden has sparked an ``efficiency revolution,'' with researchers and companies exploring approaches to reach similar performance at a fraction of the cost. Recent examples include DeepSeekV3~\cite{D24} and Kimi 2 Thinking~\cite{K25}, which have drawn wide attention for their highly efficient training strategies. While exact costs are not public, GPU usage for DeepSeekV3 suggests a compute cost of roughly 5.6 million US dollars~\cite{D24}. This is a dramatic reduction from conventional models. These results have even influenced the market value of hardware companies such as NVIDIA~\cite{NVIDIA-DeepSeek}, which sparks an optimistic question: \textit{can we truly build powerful AI systems without paying the traditional computational and financial cost?}

However, as the ``no free lunch'' principle suggests~\cite{WM97}, efficiency improvements in complex systems can carry hidden trade-offs. While current discussions emphasize performance and scalability, the potential security and privacy implications of these efficiency techniques remain largely unexplored. This gap motivates our work: we aim to understand whether the pursuit of efficiency in foundation model training introduces new vulnerabilities that may compromise robustness and safety.

Our findings demonstrate that the representative efficiency strategies studied here carry measurable security costs. Through comprehensive cross-domain experiments, we find that the evaluated efficiency-oriented variants exhibit systematic changes in their internal geometry and functional representations, alongside greater susceptibility to adversarial and privacy attacks. In short, the same strategies that reduce training requirements can also introduce new avenues for exploitation.

We make the following contributions:
\vspace{-2mm}
\begin{itemize}
    \item First systematic cross-domain analysis of the efficiency--vulnerability trade-off: We analyze multiple stages of the training pipeline, including data selection, pre-training, and alignment, across computer vision and language domains.
    \item Theoretical intuition for the observed trade-offs: We identify links between efficiency choices and measurable changes in a model's geometric and functional properties. This provides conceptual insights into why efficient models may become more fragile.
    \item Hidden security costs of efficient LLM alignment: We reveal that zero-RL models can preserve reasoning utility while exhibiting substantially greater structural brittleness and overconfidence, uncovering a previously overlooked tension between alignment efficiency and model security.
\end{itemize}

Our study demonstrates that improving training efficiency is not a free gain. As foundation models continue to shape the technological landscape, understanding this hidden cost is essential for building systems that are not only efficient and powerful but also secure and trustworthy.

\section{Related Work}
\label{sec:related}

\subsection{Training-Efficiency Strategies}
\label{sec:effi_train}
The substantial cost of modern model training has led to a wide range of techniques for reducing data and computational requirements. One prominent direction reduces training data volume through high-importance core-set selection~\cite{GZ19,JDWHGLZSS19,ZLXISMMEYYZGLZL23,JXLXLDWZ25}, which identifies a compact subset that preserves task utility. Beyond data reduction, pre-training followed by downstream fine-tuning~\cite{DBKWZUDMHGUH21,DCLT19,D24} amortizes representation learning across tasks, while parameter-efficient fine-tuning~\cite{HSWALWWC22,ZZGZSZZ24} further lowers adaptation costs by updating only a small portion of the model. A third direction concerns post-training alignment. Conventional approaches such as RLHF~\cite{CLBMLA17,OWJAWMZASRSHKMSAWCLL22} and DPO~\cite{RSMMEF23} rely on preference data and additional optimization; recent ``zero RL training'' methods~\cite{ZHLLHMH25,CYWWZCLHFYXCYCZLWYHPCLSZD26} reduce this burden by starting directly from capable base models without initial SFT. These three lines of work define the training-efficiency landscape examined in our study.

\subsection{Efficiency and Model Trustworthiness}
\label{sec:trade-off}
A growing body of work shows that choices made to improve or modify training can also reshape model trustworthiness. Differentially private fine-tuning~\cite{LTLH22,BJWXLLD25} can reduce memorization and membership-inference risk, whereas model distillation~\cite{GWZZLSW21,MVSV26} can transfer or amplify backdoor behavior. Data selection~\cite{WBZ25,RRRRS25} may likewise influence privacy leakage and stability. Related studies consider efficiency after training, showing that pruning~\cite{YLXLCLZZMW19,ESGVV25}, quantization~\cite{YCGF24}, and LLM compression more broadly~\cite{HDZLXLDBJXKHSWL24} can degrade robustness and other dimensions of trustworthiness. Collectively, these findings establish important connections between individual efficiency mechanisms and security, but they remain fragmented across particular methods, stages, and domains. Our work provides the first systematic cross-domain analysis that connects these questions across multiple stages of model training.

\section{Empirical Framework}
\label{sec:framework}
To systematically investigate the efficiency-vulnerability trade-off across model families and training stages, we combine controlled, utility-matched studies of vision models with cross-scale evaluations of representative efficient alignment recipes for LLMs.
\subsection{Defining Training Efficiency}
\label{sec:define_efficiency}
In foundation model development, training efficiency means reducing data, computation, or alignment overhead while maintaining comparable task utility. Our work focuses on three major efficiency paradigms that correspond to different stages of development, illustrated in Figure~\ref{fig:lifecycle}:
\begin{minipage}[t]{0.49\textwidth}
\vspace{0pt}
\shortsection{Data Selection} This stage minimizes the amount of training data used to save on memory, storage, and collection costs. We use strategies like high-importance core-set selection to find the most informative data samples. This often involves data valuation metrics like Shapley values to prioritize ``high-importance'' records~\cite{GZ19}. This approach is ideal when collecting data is expensive or when the goal is to train on only the most critical information.
\end{minipage}
\hfill
\begin{minipage}[t]{0.48\textwidth}
\vspace{0pt}
\centering
\includegraphics[width=\linewidth]{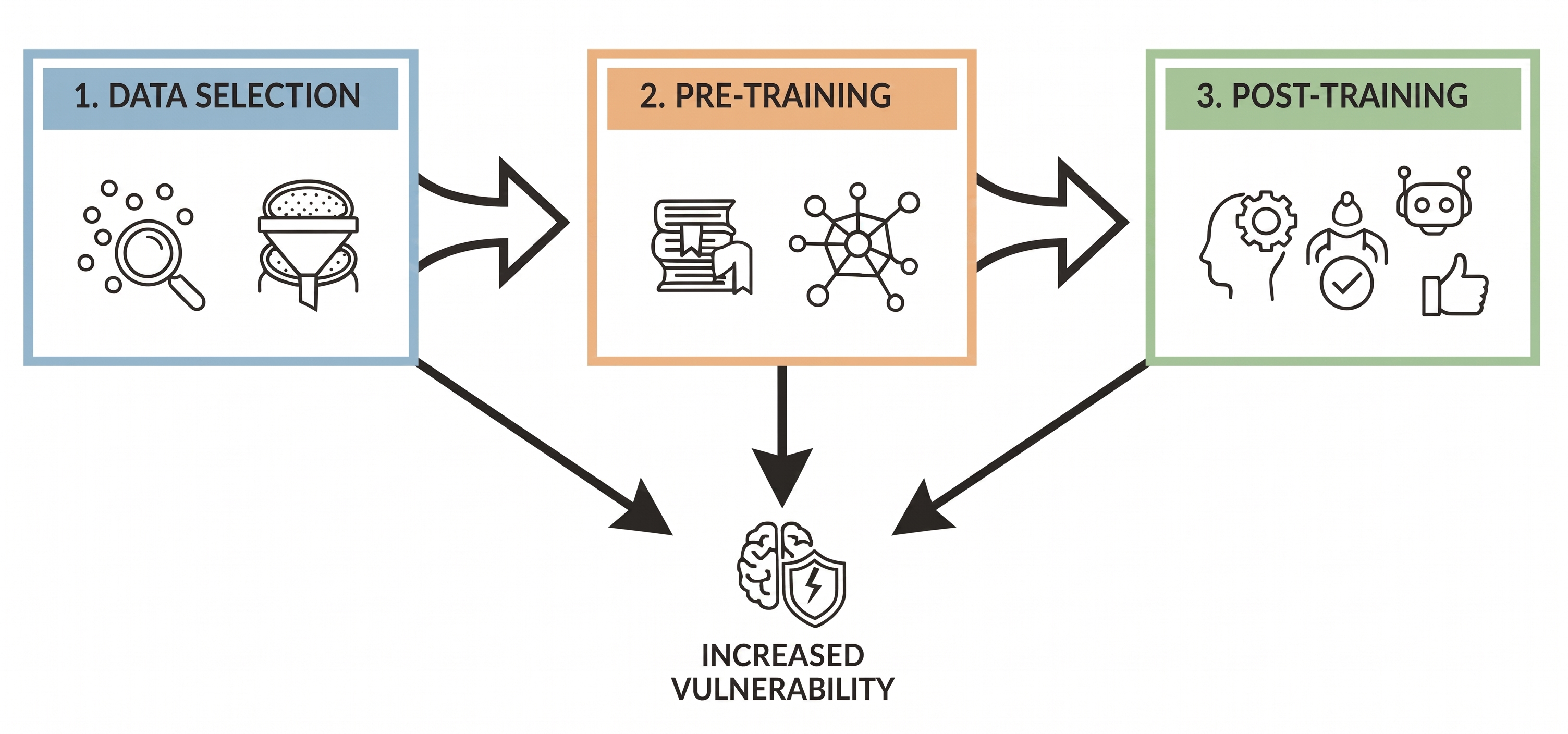}
\captionof{figure}{Lifecycle of model training.}
\label{fig:lifecycle}
\end{minipage}

\shortsection{Pre-training} This paradigm uses a two-stage Pre-training and Fine-tuning pipeline. Pre-training uses the most resources to build general knowledge from a massive dataset. Fine-tuning then uses this knowledge base to adapt the model to specific tasks with very little additional compute. Our study explores the trade-off between using this efficient pre-trained base versus the higher cost of training a model from scratch.

\shortsection{Post-training Alignment}
Alignment elicits specialized behaviors, such as mathematical reasoning. Conventional alignment pipelines typically build on SFT and may regularize policy updates relative to a reference policy. For example, KL-regularized RLHF maximizes
\begin{equation}
\mathcal{J}_{\mathrm{RLHF}}(\theta)
=
\mathbb{E}_{x\sim\mathcal{D},\,y\sim\pi_\theta(\cdot|x)}
\left[
r(x,y)
-
\beta\log
\frac{\pi_\theta(y|x)}
{\pi_{\mathrm{ref}}(y|x)}
\right],
\end{equation}
where $\mathcal{D}$ is the prompt distribution, $\pi_\theta$ is the trainable policy, $r(x,y)$ is the reward, $\pi_{\mathrm{ref}}$ is the reference policy, and $\beta$ controls the strength of the deviation penalty. The evaluated ``zero RL training'' approaches reduce alignment overhead by starting directly from capable base models without preliminary SFT and simplifying conventional alignment components, including the omission of explicit KL regularization. Our comparison therefore examines whether these utility-preserving approaches incur greater security risks, treating each training pipeline as a whole rather than isolating the effect of any single omitted component.

\subsection{Measuring Vulnerability}
\label{sec:measure_vul}
To quantify the cost of efficiency, we adopt a comprehensive approach to measure model vulnerability across three core security concerns: data privacy, robustness, and proprietary security.

\shortsection{Data Privacy} This dimension measures whether training data can be inferred from the model. We evaluate this risk using membership inference attacks (MIA), where an adversary attempts to determine whether a specific sample was included in the training set. For vision tasks, we adopt three computationally efficient and widely used MIAs: prediction confidence-based attack~\cite{SSM19}, entropy-based attack~\cite{SZHBFB19}, and modified prediction entropy-based attack~\cite{SM21}. For math reasoning LLMs, we construct source-defined candidate-member and candidate-non-member groups and evaluate answer likelihood~\cite{FWGLLJ24} (analogous to loss-based attacks), model self-confidence~\cite{YKYKKKCKS25} (analogous to confidence-based attacks), and the Min-$k$\% probability attack, which provide complementary membership signals for discrete reasoning tasks.

\begin{table}[t]
\centering
\begin{minipage}[t]{0.48\textwidth}
\vspace{0pt}
\centering
\caption{Accuracy (training accuracy in brackets) and TPR at 0.1\% FPR of membership inference for data selection with Shapley values. Higher TPR indicates greater privacy leakage.}
\vspace{-2mm}
\scriptsize
\setlength{\tabcolsep}{1.5pt}
\begin{tabular*}{\linewidth}{@{\extracolsep{\fill}}ll|c|ccc@{}}
\toprule
\rowcolor{white}
\multirow{2}{*}{Dataset} & \multirow{2}{*}{Value} & \multirow{2}{*}{Acc} & \multicolumn{3}{c}{MIA} \\
\cmidrule(l{3pt}r{0pt}){4-6}
& & & conf & entr & modi \\
\midrule
\multirow{2}{*}{CIFAR-10} & Low & 48.07 (99.97) & 0.12\% & 0.05\% & 0.11\% \\
& High & 47.97 (99.97) & 0.16\% & 0.09\% & 0.16\% \\
\midrule
\multirow{2}{*}{CIFAR-10 (swin)} & Low & 37.01 (99.95) & 0.08\% & 0.05\% & 0.09\% \\
& High & 36.94 (99.93) & 0.29\% & 0.30\% & 0.30\% \\
\midrule
\multirow{2}{*}{TinyImageNet} & Low & 36.09 (100.00) & 0.08\% & 0.06\% & 0.10\% \\
& High & 36.17 (100.00) & 0.22\% & 0.35\% & 0.27\% \\
\midrule
\multirow{2}{*}{PubFig83} & Low & 44.01 (100.00) & 0.32\% & 0.32\% & 0.32\% \\
& High & 44.26 (100.00) & 1.51\% & 1.43\% & 1.51\% \\
\bottomrule
\end{tabular*}
\label{tab:mia_shapley}
\end{minipage}\hfill
\begin{minipage}[t]{0.48\textwidth}
\vspace{0pt}
\centering
\caption{Adversarial robustness for data selection and pre-training in vision models, measured by the average number of correctly classified PGD iterates (higher is better).}
\vspace{-2mm}
\scriptsize
\setlength{\tabcolsep}{1.5pt}
\begin{tabular*}{\linewidth}{@{\extracolsep{\fill}}l|lc|lc@{}}
\toprule
\rowcolor{white}
\multirow{2}{*}{Dataset} & \multicolumn{2}{c}{Data Selection} & \multicolumn{2}{c}{Pre-training} \\
\cmidrule(l{0pt}r{2pt}){2-3}\cmidrule(l{2pt}r{0pt}){4-5}
& Value & Steps & Paradigm & Steps \\
\midrule
\multirow{2}{*}{CIFAR-10} & Low & 3.85 & Scratch & 3.58 \\
& High & 1.33 & Pre-train & 1.00 \\
\midrule
\multirow{2}{*}{CIFAR-10 (swin)} & Low & 0.70 & Scratch & 3.22 \\
& High & 0.18 & Pre-train & 0.11 \\
\midrule
\multirow{2}{*}{TinyImageNet} & Low & 20.89 & Scratch & 14.03 \\
& High & 10.46 & Pre-train & 3.64 \\
\midrule
\multirow{2}{*}{PubFig83} & Low & 2.34 & Scratch & 1.83 \\
& High & 1.30 & Pre-train & 0.41 \\
\bottomrule
\end{tabular*}
\label{tab:distance}
\end{minipage}
\end{table}

\shortsection{Model Robustness and Resilience}
This category assesses whether the model is easily manipulated by external perturbations or prone to functional degradation. We evaluate this across two dimensions: adversarial resistance and adaptation stability.
To quantify susceptibility to adversarial manipulation (being ``tricked''), we employ domain-specific attacks. For vision tasks, we measure how many Projected Gradient Descent (PGD)~\cite{MMSTV18} iterates remain correctly classified on average; a lower score indicates a more fragile robustness margin. For math reasoning LLMs, we use the RobustMath dataset~\cite{ZWJYYLWHH24} to evaluate resilience against semantic perturbations and AutoDAN~\cite{LXCX24} on AdvBench~\cite{ZWKF23} to evaluate jailbreak susceptibility under adversarial prompting.
Additionally, we evaluate fine-tuning stability by adapting the model to a disjoint domain using LoRA~\cite{HSWALWWC22}, measuring if the model maintains its integrity or suffers catastrophic performance loss under distribution shift.

\shortsection{Proprietary Security} This dimension concerns the risk of unauthorized extraction of a model's proprietary knowledge and structure. We evaluate this threat using Model Stealing Attacks (MS), in which an adversary attempts to reconstruct a functional equivalent of the target model by querying its outputs. We focus on a logit-query setting, using data from the target distribution for vision tasks.

\section{Controlled Study in Vision Tasks}
\label{sec:exp_cv}

\subsection{Experimental setup}
\label{sec:cv_setup}
To examine how efficiency choices affect vulnerability at comparable utility, we compare a higher-cost baseline with an \textit{Efficient Variant}. We calibrate dataset sizes and configurations so that each pair achieves nearly identical training and testing accuracy, preventing task-performance differences from confounding the security comparison.
All vision results are means over three seeds; complete mean $\pm$ standard deviation results are reported in Appendix~\ref{sec:three_seed_vision_results}.

\shortsection{Datasets and Architecture} 
We employ ResNet-18 and Swin Transformer architectures to evaluate vulnerability across distinct inductive biases, using ResNet-18 by default unless ``Swin'' is specified. Our experiments span three datasets: CIFAR-10 and TinyImageNet for standard object classification, and PubFig83 for fine-grained face recognition. This diversity allows us to test whether the observed patterns persist across architectures and tasks involving different levels of intra-class variation.

\shortsection{Model Training} We study efficiency through data selection and pre-training. For data selection, KNN-Shapley~\cite{JDWHGLZSS19} ranks the training samples: \textit{High} uses a smaller subset with the highest values, whereas \textit{Low} uses a larger subset with the lowest values. The two subsets are disjoint, and their sizes are intentionally calibrated to match clean utility, with all other training hyperparameters fixed within each comparison. For pre-training, the Efficient Variant fine-tunes a pre-trained model on less task-specific data, whereas the higher-cost baseline is trained from scratch on a larger dataset. We thus operationalize efficiency through reduced task-specific data and reuse of pre-trained representations.

\begin{table}[t]
\centering
\caption{Accuracy (training accuracy in brackets), membership inference (MIA), and model stealing (MS) under pre-trained fine-tuning and scratch training. Higher MIA indicates greater privacy leakage. MS reports stolen-model accuracy with target--surrogate agreement in parentheses; higher values indicate more successful extraction.}
\vspace{-2mm}
\setlength{\tabcolsep}{4.pt}
\scalebox{0.90}{
\begin{tabular}{ll|c|ccc|cc}
\toprule
\rowcolor{white}
\multirow{2}{*}{Dataset} & \multirow{2}{*}{Paradigm} & \multirow{2}{*}{Acc} & \multicolumn{3}{c}{MIA} & \multicolumn{2}{c}{MS} \\
\cmidrule(l{3pt}r{3pt}){4-6}\cmidrule(l{3pt}r{0pt}){7-8}
& & & conf & entr & modi & 1000 & 1000 (Pre-train) \\
\midrule
\multirow{2}{*}{CIFAR-10} & Scratch & 67.71 (99.91) & 0.05\% & 0.04\% & 0.05\% & 62.60 (0.748) & 65.22 (0.758) \\
 & Pre-train & 67.88 (100.00) & 0.21\% & 0.16\% & 0.21\% & 58.05 (0.669) & 67.25 (0.785) \\
\midrule
\multirow{2}{*}{CIFAR-10 (swin)} & Scratch & 73.85 (100.00) & 0.12\% & 0.12\% & 0.11\% & 42.41 (0.446) & 57.70 (0.536) \\
& Pre-train & 74.01 (100.00) & 0.54\% & 0.54\% & 0.54\% & 39.01 (0.437) & 60.06 (0.651) \\
\midrule
\multirow{2}{*}{TinyImageNet} & Scratch & 34.91 (99.98) & 0.13\% & 0.07\% & 0.14\% & 24.65 (0.381) & 27.53 (0.379) \\
 & Pre-train & 34.93 (100.00) & 0.75\% & 0.75\% & 0.67\% & 17.65 (0.305) & 29.82 (0.529) \\
\midrule
\multirow{2}{*}{PubFig83} & Scratch & 35.19 (100.00) & 0.01\% & 0.01\% & 0.01\% & 17.81 (0.358) & 18.59 (0.299) \\
 & Pre-train & 34.79 (100.00) & 0.36\% & 0.46\% & 0.37\% & 7.89 (0.183) & 25.26 (0.533) \\
\bottomrule
\end{tabular}
}
\label{tab:mis_ms_pretrain}
\end{table}

\begin{figure}[t]
\centering
\begin{subfigure}{0.245\columnwidth}
\includegraphics[width=\columnwidth]{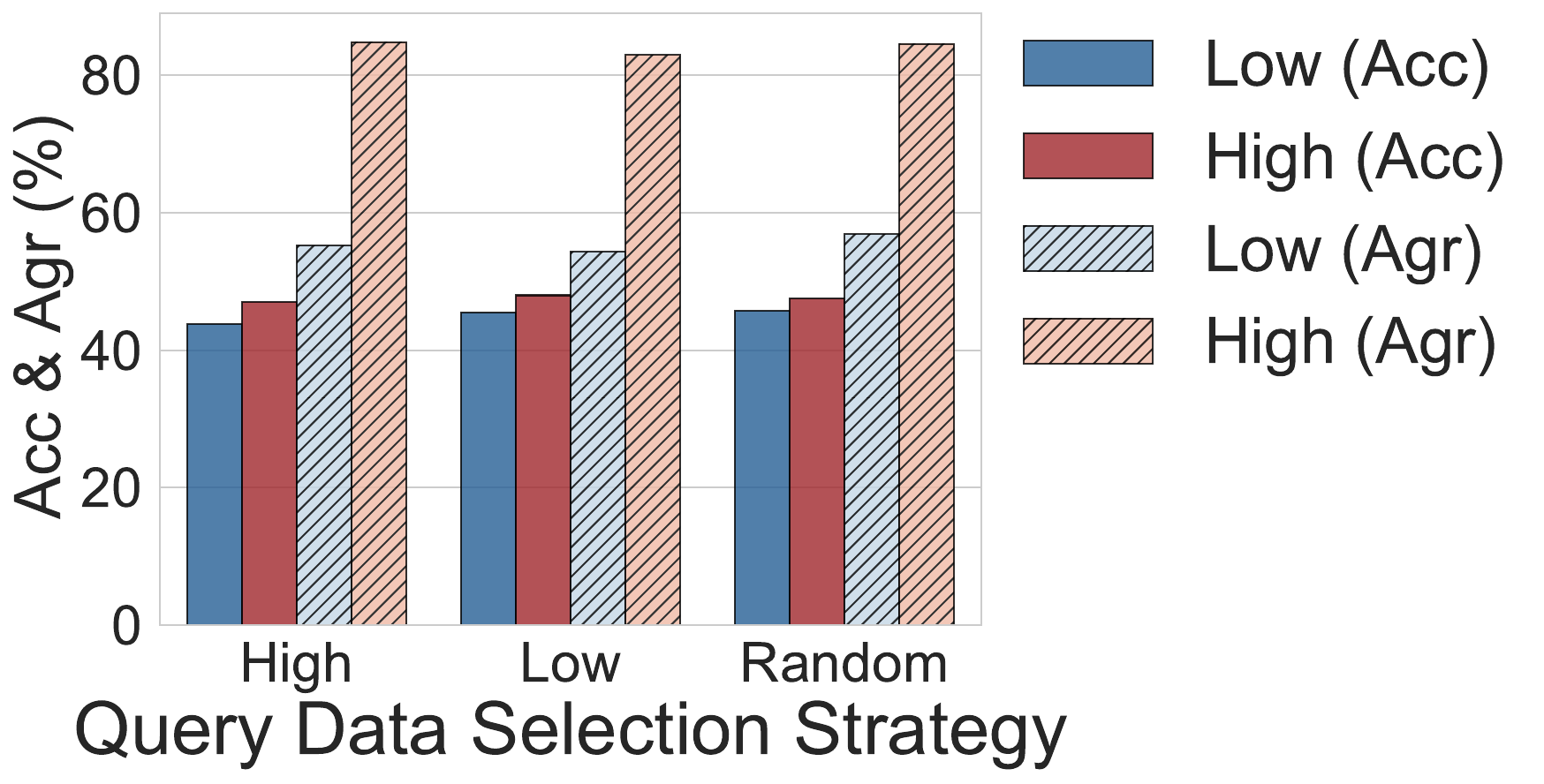}
\caption{CIFAR-10}
\end{subfigure}
\begin{subfigure}{0.245\columnwidth}
\includegraphics[width=\columnwidth]{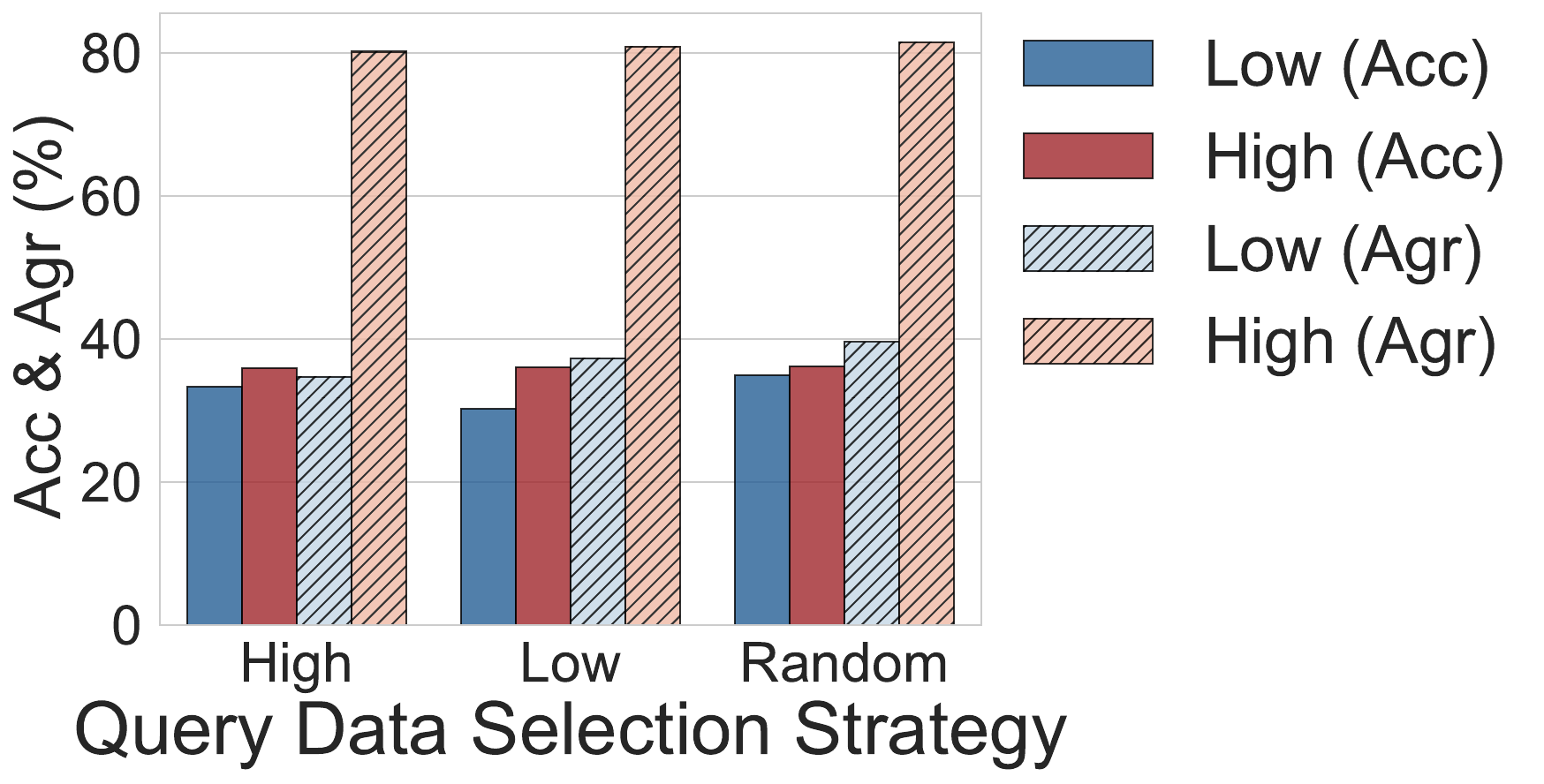}
\caption{CIFAR-10 (swin)}
\end{subfigure}
\begin{subfigure}{0.245\columnwidth}
\includegraphics[width=\columnwidth]{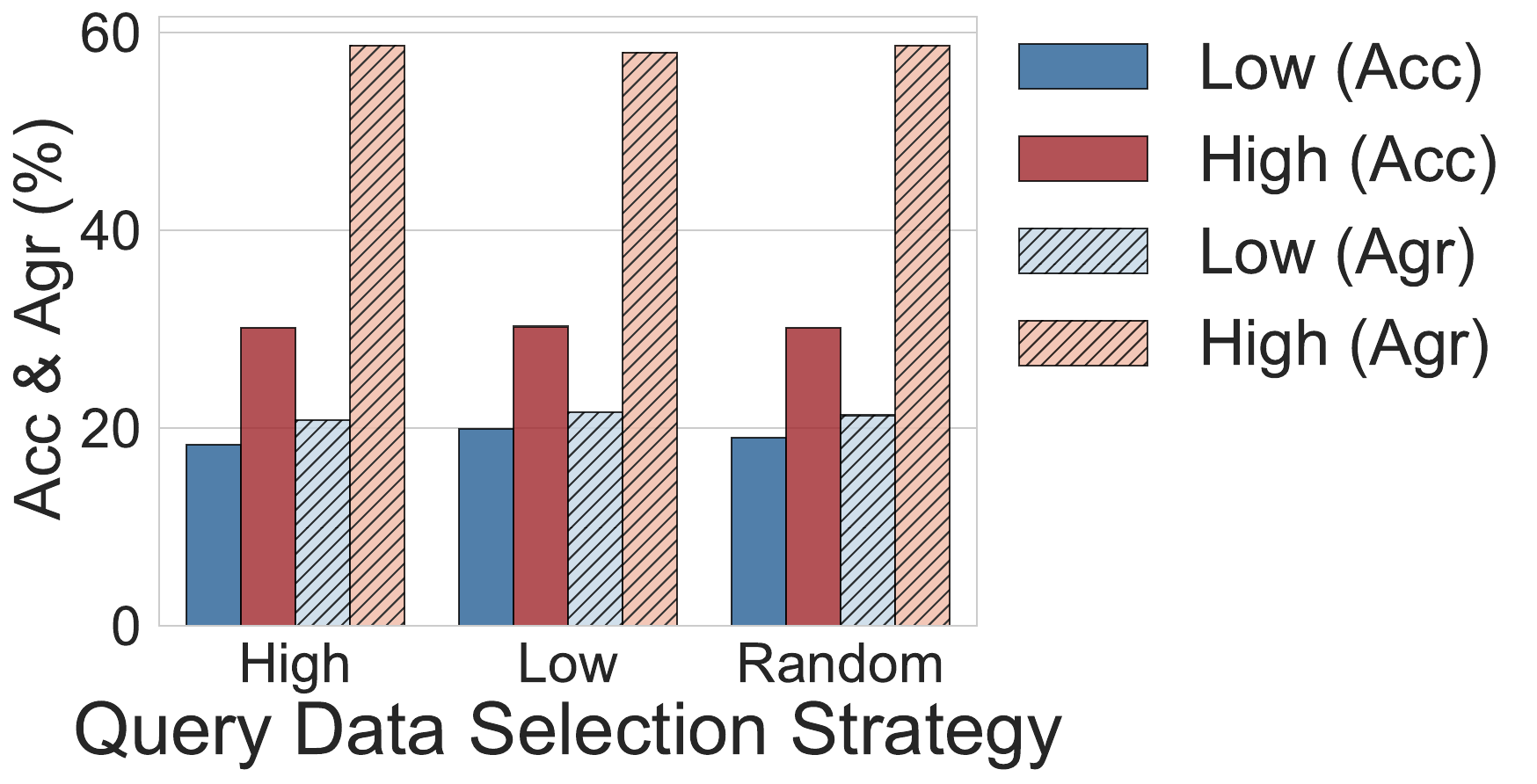}
\caption{TinyImageNet}
\end{subfigure}
\begin{subfigure}{0.245\columnwidth}
\includegraphics[width=\columnwidth]{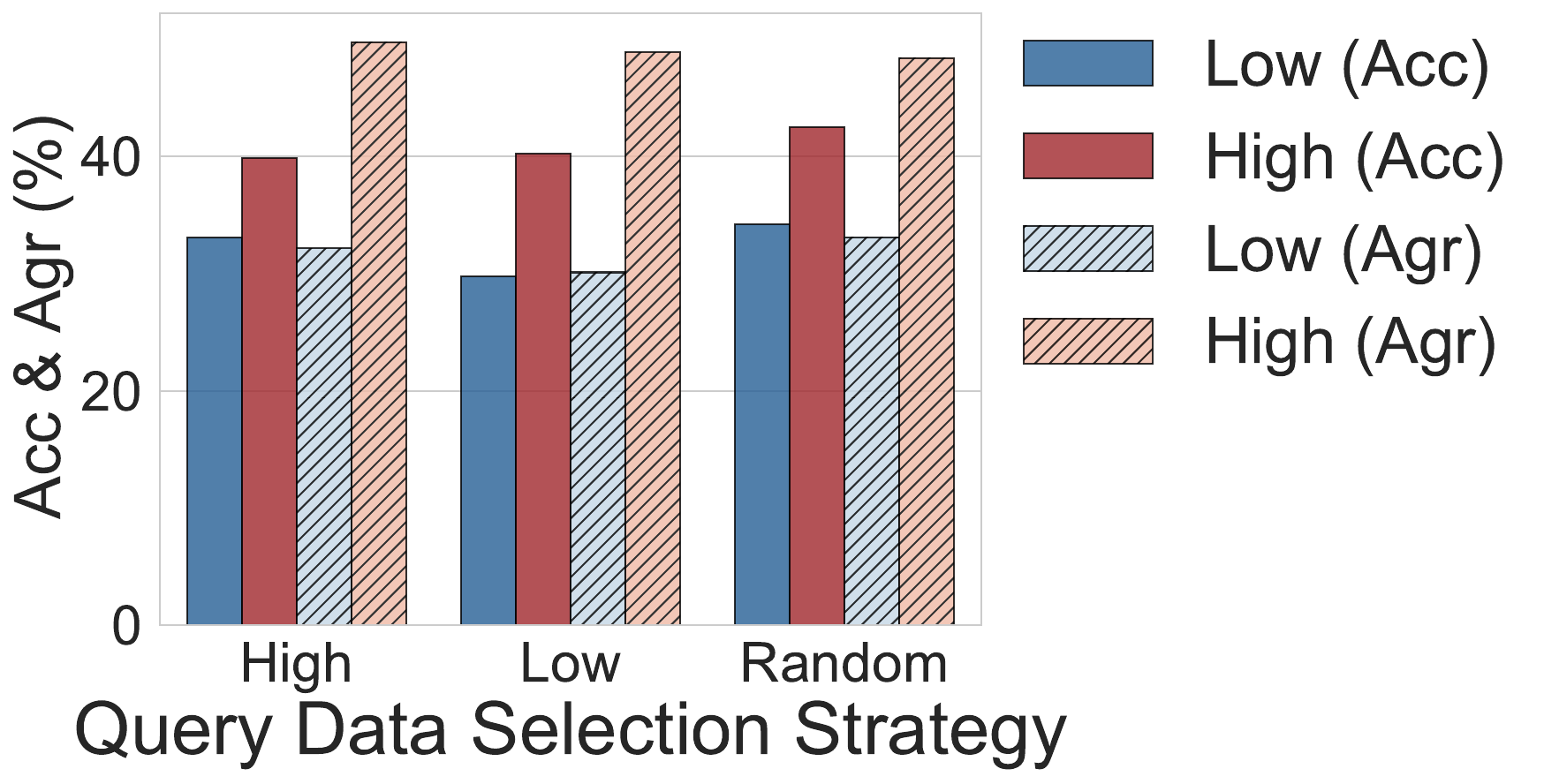}
\caption{PubFig83}
\end{subfigure}
\caption{Model stealing under Shapley-based data selection. ``Acc'' denotes stolen-model test accuracy, ``Agr'' the fraction of test samples on which target and stolen models agree, and ``Random'' random query selection; higher Acc and Agr indicate more successful extraction. Query pools contain 1,000 distinct images (500 for PubFig83 owing to data limitations).}
\label{fig:ms_shapley}
\end{figure}

\begin{table*}[t]
\centering
\caption{Sharpness, Expected Calibration Error, and PCA Participation Ratio for vision models.}
\vspace{-2mm}
\setlength{\tabcolsep}{6.pt}
\scalebox{0.90}{
\begin{tabular}{ll|ccc|l|ccc}
\toprule
\rowcolor{white}
Dataset & Value & Sharpness & ECE & PCA PR & Paradigm & Sharpness & ECE & PCA PR \\
\midrule
\multirow{2}{*}{CIFAR-10} & Low & 0.0410 & 0.382 & 28.490 & Scratch & 0.0311 & 0.219 & 14.266 \\
 & High & 0.0591 & 0.430 & 10.793 & Pre-train & 0.1332 & 0.232 & 25.137 \\
\midrule
\multirow{2}{*}{CIFAR-10 (swin)} & Low & 0.0170 & 0.410 & 13.907 & Scratch & 0.0010 & 0.060 & 9.897 \\
& High & 0.0189 & 0.436 & 10.896 & Pre-train & 0.0373 & 0.241 & 26.786 \\
\midrule
\multirow{2}{*}{TinyImageNet} & Low & 0.0548 & 0.336 & 265.298 & Scratch & 0.0511 & 0.317 & 157.477 \\
 & High & 0.0600 & 0.384 & 74.960 & Pre-train & 0.1201 & 0.251 & 163.206 \\
\midrule
\multirow{2}{*}{PubFig83} & Low & 0.1234 & 0.208 & 48.395 & Scratch & 0.1349 & 0.350 & 31.940 \\
 & High & 0.1475 & 0.299 & 33.936 & Pre-train & 0.2551 & 0.384 & 39.009 \\
\bottomrule
\end{tabular}
}
\label{tab:sharp_ece_pca_CV}
\end{table*}

\subsection{Experimental Results}
\label{sec:cv_results}

Our controlled comparisons uncover a fundamental trade-off between data efficiency and security. Despite comparable generalization and overfitting levels (the gap between training and testing accuracy), High-Shapley Efficient Variants exhibit heightened MIA vulnerability (Table~\ref{tab:mia_shapley}). Relative to the Low-Shapley baselines, the gap is most pronounced with Swin Transformers and on challenging datasets such as TinyImageNet and PubFig83. Beyond this privacy exposure, High-Shapley variants remain correctly classified for fewer PGD iterates, revealing reduced adversarial robustness (Table~\ref{tab:distance}). The model-stealing results broaden this concern further: attacks against these targets produce stolen models with higher accuracy and agreement (Figure~\ref{fig:ms_shapley}). The consistency across these distinct attack surfaces establishes that importance-based data efficiency can preserve predictive utility while weakening the model's overall security posture.

The comparison of training paradigms further highlights the security cost of efficiency gains. At comparable utility, the pre-trained Efficient Variant exhibits greater MIA risk and lower adversarial robustness than the scratch-trained baseline (Tables~\ref{tab:mis_ms_pretrain} and~\ref{tab:distance}). Model stealing reveals a critical nuance: although pre-trained targets resist scratch-initialized surrogates, this advantage disappears when the attacker uses the same pre-trained checkpoint. Because this checkpoint is publicly available, the informed attack requires no privileged access. Under this realistic threat model, the higher stolen-model accuracy and agreement show that the reusable prior reducing task-specific training cost can also lower the empirical barrier to functional replication.

\begin{figure}[t]
\centering
\begin{subfigure}{0.245\columnwidth}
\includegraphics[width=\columnwidth]{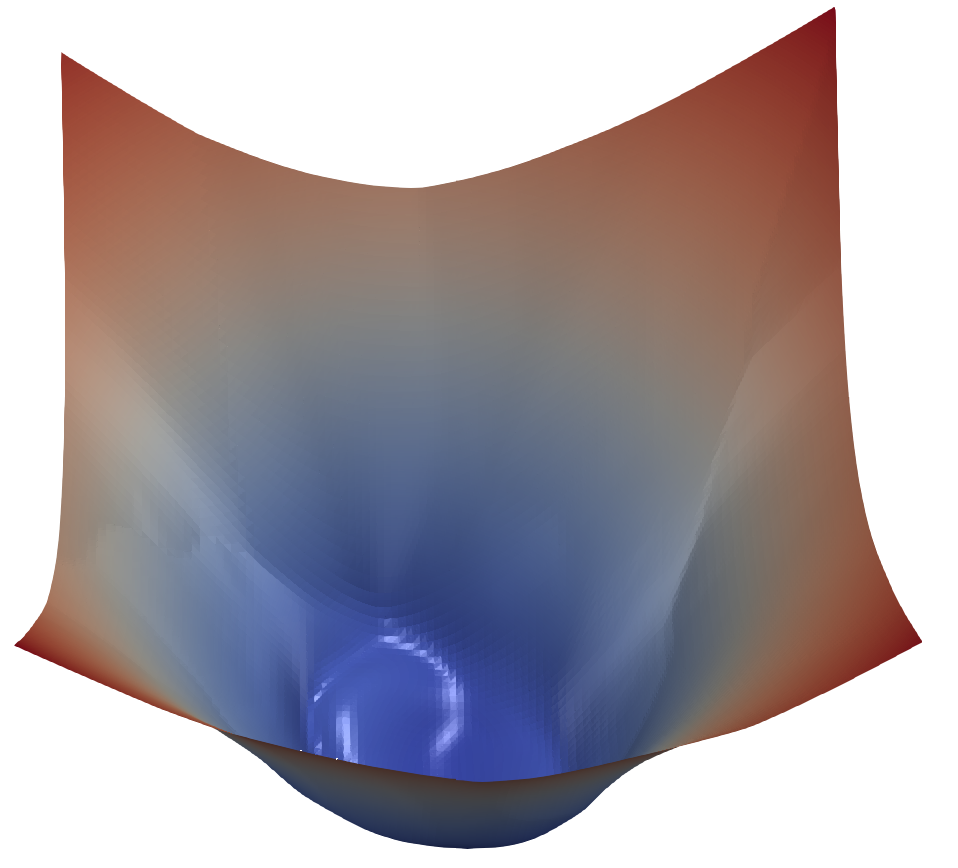}
\caption{High}
\end{subfigure}
\hfill
\begin{subfigure}{0.245\columnwidth}
\includegraphics[width=\columnwidth]{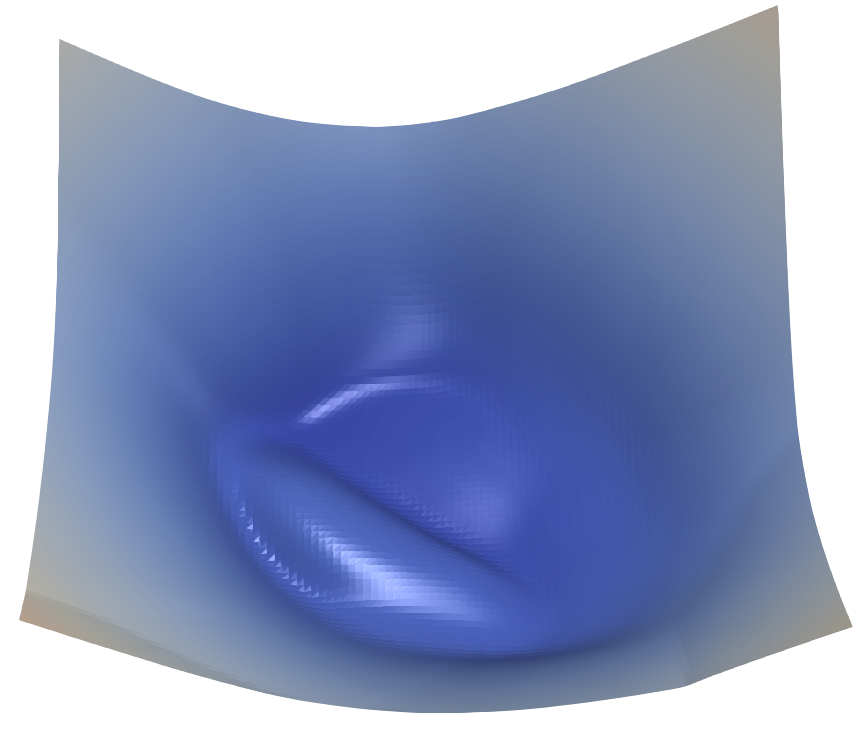}
\caption{Low}
\end{subfigure}
\hfill
\begin{subfigure}{0.245\columnwidth}
\includegraphics[width=\columnwidth]{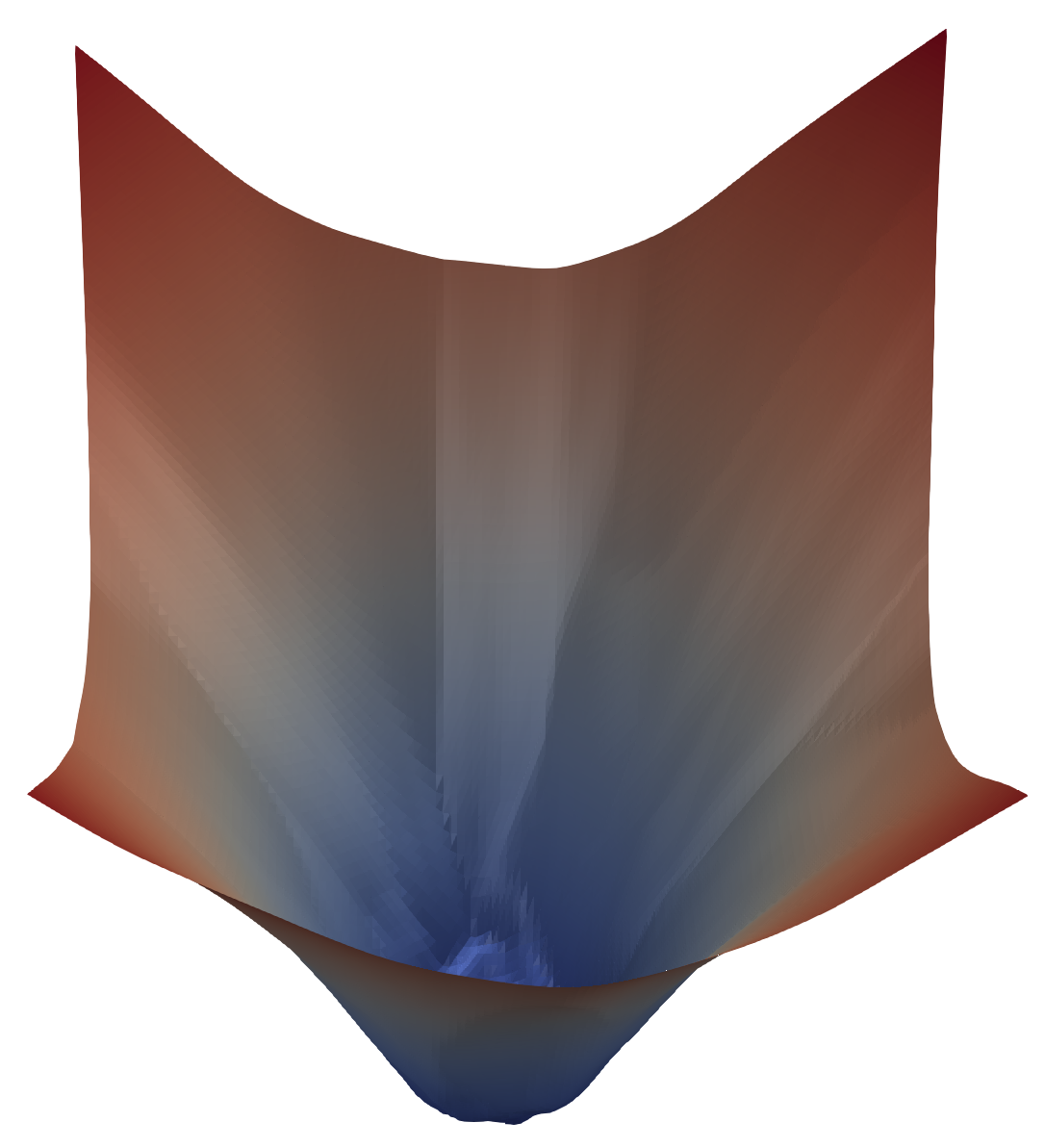}
\caption{Pre-train}
\end{subfigure}
\hfill
\begin{subfigure}{0.245\columnwidth}
\includegraphics[width=\columnwidth]{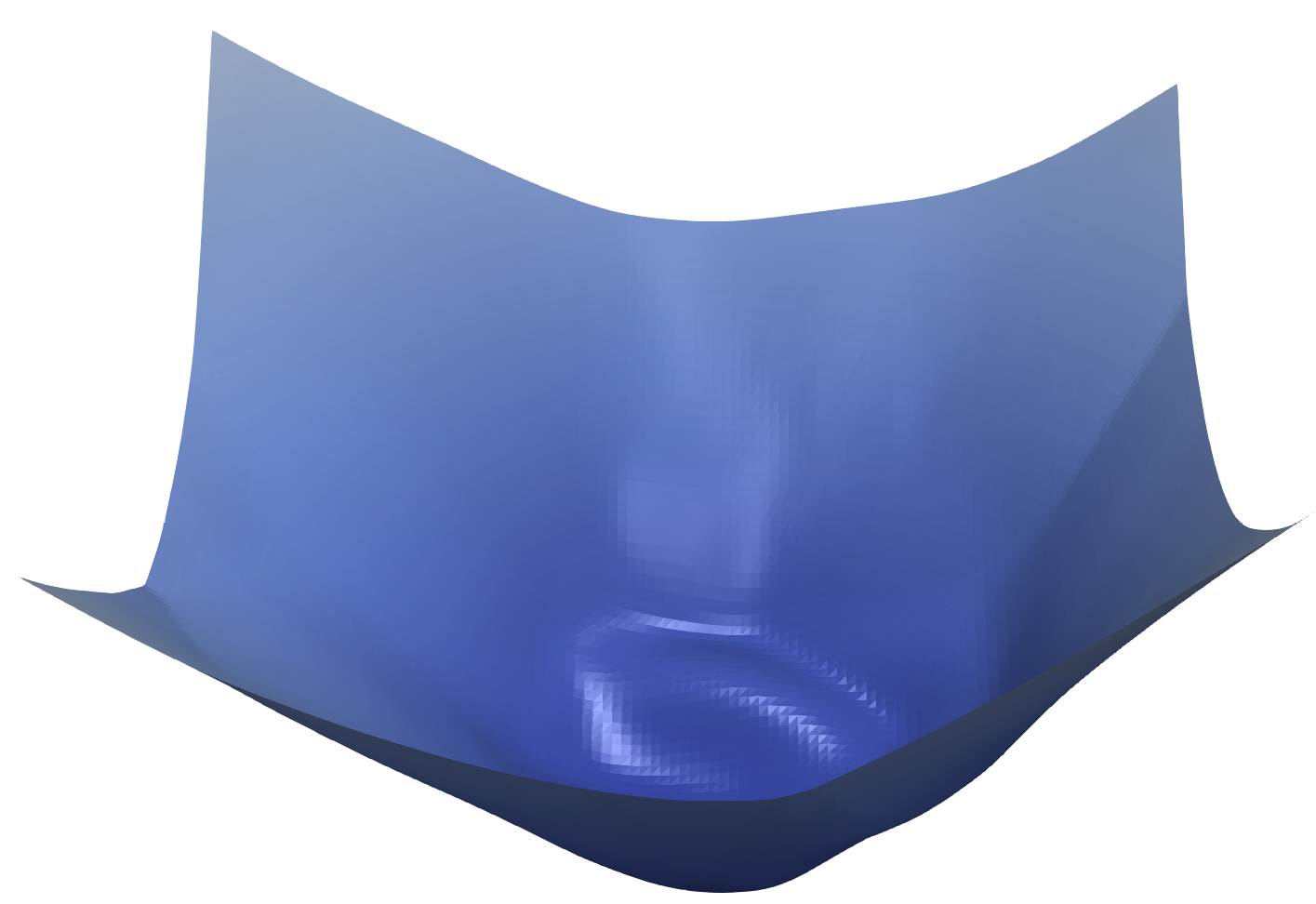}
\caption{Scratch}
\end{subfigure}
\caption{Loss landscape for CIFAR-10.}
\label{fig:landscape_cifar}
\end{figure}

\section{Mechanistic Analysis}
\label{sec:cv_analysis}
We analyze structural differences between Efficient Variants and baselines from two complementary perspectives: geometric stability and functional representation structure. These analyses identify concrete geometric and functional signatures of the evaluated Efficient Variants, providing mechanistic insight into their observed security degradation.

\subsection{Geometric Instability}
\label{sec:geo_instability}
We hypothesize that the evaluated efficiency strategies steer models toward geometrically less stable solutions. We characterize geometric instability directly through loss-landscape sharpness and examine its accompanying calibration behavior through Expected Calibration Error (ECE), which measures the alignment between predictive confidence and accuracy.

\textbf{Loss Landscape Sharpness.}
We measure sharpness through local loss sensitivity around the model parameters $\theta$. For a test batch $b$ with loss $\mathcal{L}_b$, let $\widehat\delta_b$ denote the bounded loss-increasing perturbation found by projected gradient ascent, and let $\mathcal{B}$ denote the collection of test batches. We report
\begin{equation}
\label{equation:sharpness}
\widehat S(\theta)=\frac{1}{|\mathcal{B}|}\sum_{b\in\mathcal{B}}
\left[\mathcal{L}_b(\theta+\widehat\delta_b)-\mathcal{L}_b(\theta)\right].
\end{equation}
A larger value indicates greater local loss sensitivity. A local Taylor expansion connects this quantity to directional gradients and Hessian curvature (Appendix~\ref{app:sharpness_theory}), motivating its use as a measure of geometric stability. This connection suggests that prioritizing highly influential samples can concentrate gradient contributions, while fine-tuning on limited data can reduce gradient diversity and favor interpolation regimes associated with sharper solutions~\cite{GZ19,YPLPRB18,F20,KRJML22}.

Across all evaluated comparisons, the High-Shapley and pre-trained variants exhibit higher sharpness than their respective baselines (Table~\ref{tab:sharp_ece_pca_CV}) and visibly sharper local loss landscapes (Figure~\ref{fig:landscape_cifar}). Despite operating at different training stages, both efficiency strategies therefore exhibit the same pattern of reduced geometric flatness.

\begin{figure}[t]
\centering
\begin{subfigure}{0.245\columnwidth}
\includegraphics[width=\columnwidth]{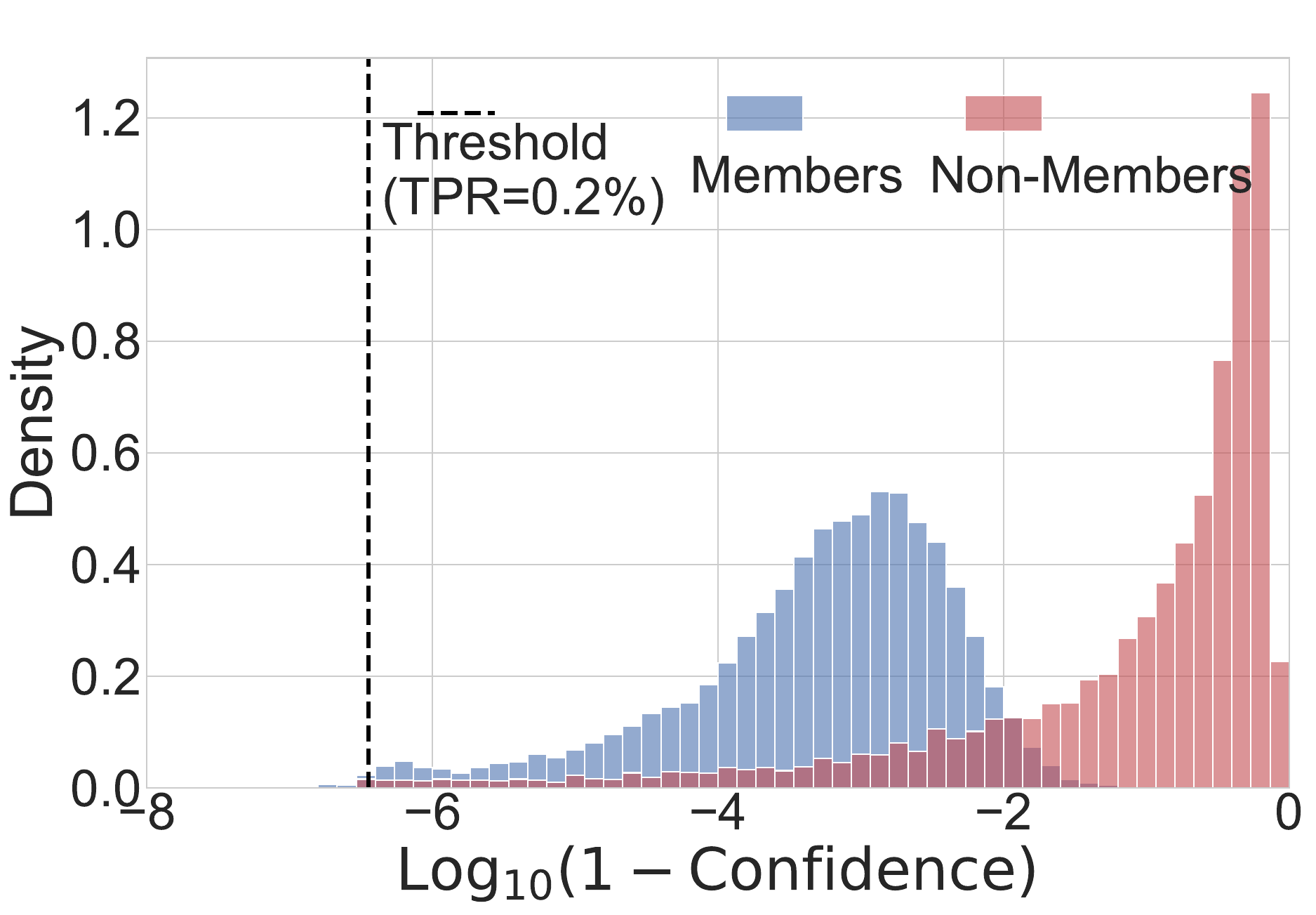}
\caption{High (confidence)}
\end{subfigure}
\begin{subfigure}{0.245\columnwidth}
\includegraphics[width=\columnwidth]{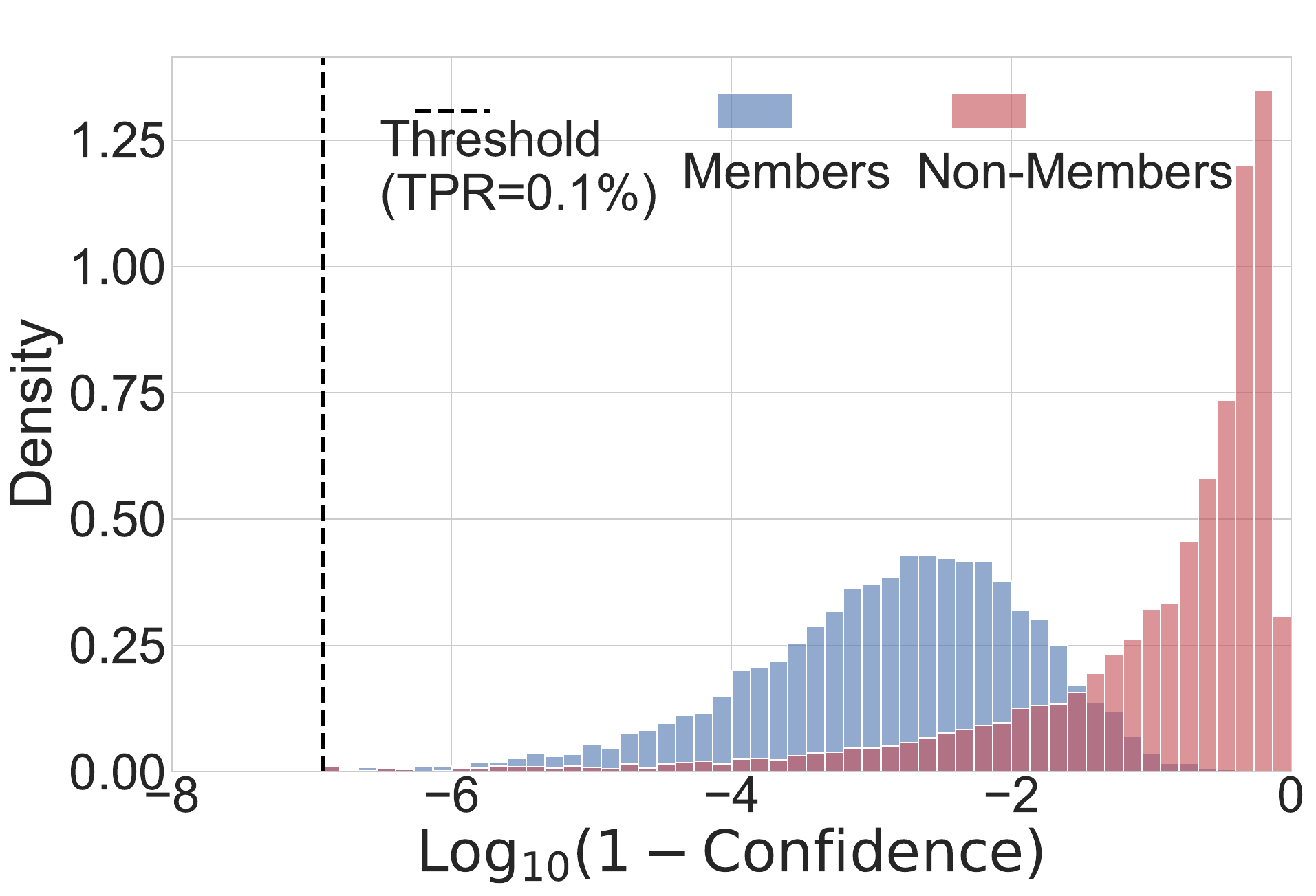}
\caption{Low (confidence)}
\end{subfigure}
\begin{subfigure}{0.245\columnwidth}
\includegraphics[width=\columnwidth]{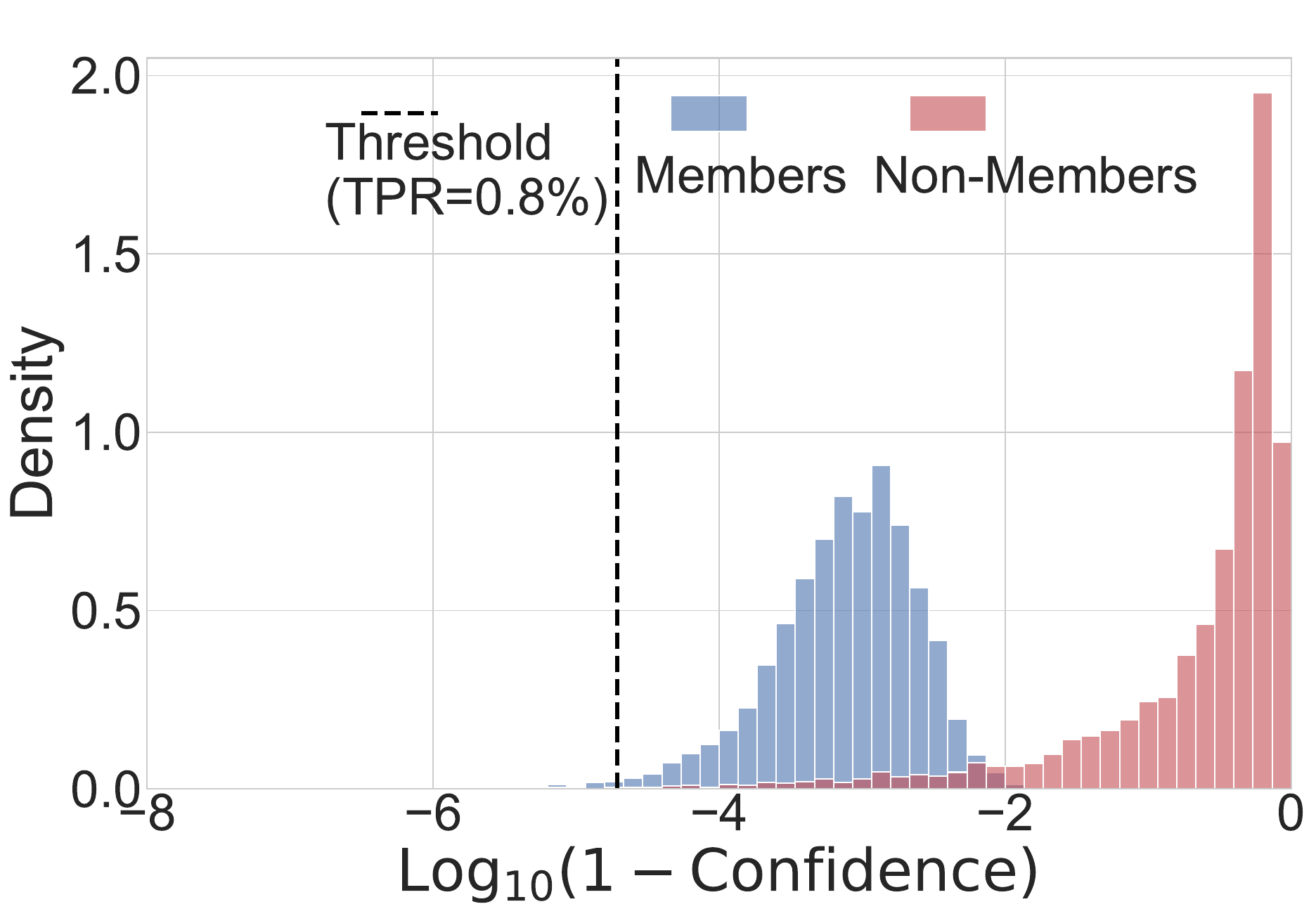}
\caption{Pre-train (confidence)}
\end{subfigure}
\begin{subfigure}{0.245\columnwidth}
\includegraphics[width=\columnwidth]{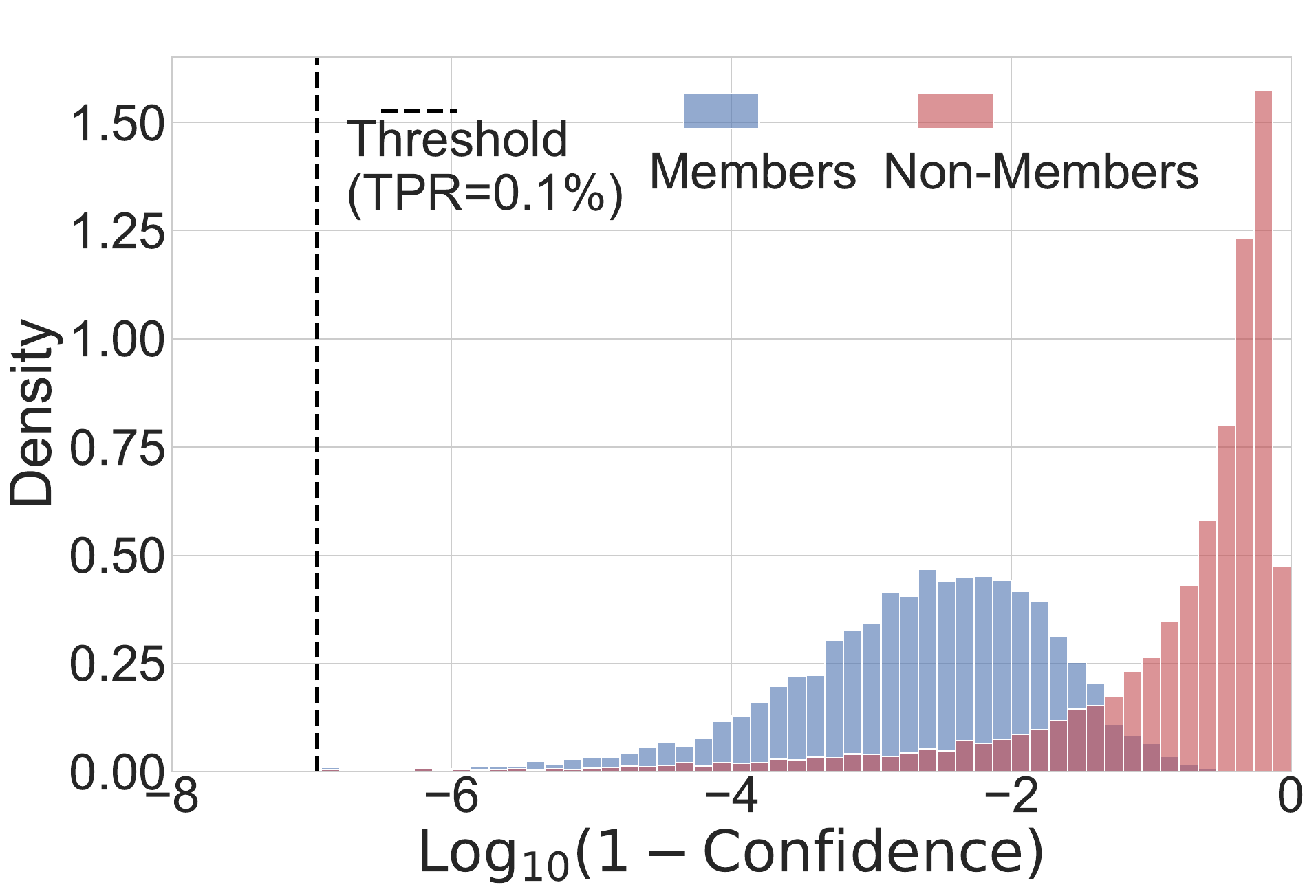}
\caption{Scratch (confidence)}
\end{subfigure}
\caption{Log-scale member/non-member confidence distributions on TinyImageNet.}
\label{fig:loss_confidence_Tiny}
\end{figure}

\textbf{Expected Calibration Error.}
We form $M$ equal-width confidence bins~\cite{GPSW17}:
\begin{equation}
\mathrm{ECE}=\sum_{m=1}^{M}\frac{|B_m|}{N}\left|\operatorname{acc}(B_m)-\operatorname{conf}(B_m)\right|,
\end{equation}
where $N$ is the total sample count, and $\operatorname{acc}(B_m)$ and $\operatorname{conf}(B_m)$ are the average empirical accuracy and confidence in bin $B_m$, respectively. Classical analyses show that, for separable data and positively homogeneous models, gradient-based optimization can increase weight norms and confidence after the training error vanishes~\cite{SHNS18,LL20}. Guided by this result, we interpret the observed behavior as follows: the evaluated efficiency strategies may accelerate this ``logit explosion'' without commensurate accuracy gains. Data selection can concentrate influential gradient contributions, while limited-data fine-tuning can reach separation earlier and devote more updates to margin growth.

Our analysis reveals generally elevated miscalibration among Efficient Variants (Table~\ref{tab:sharp_ece_pca_CV} and Figure~\ref{fig:loss_confidence_Tiny}), while the associated separation between member and non-member confidence supplies an exploitable signal for MIA. Interestingly, TinyImageNet exhibits an inverse relationship between calibration and privacy. Although pre-training improves calibration, it heightens MIA vulnerability. We interpret this contrast through the overlap between TinyImageNet and ImageNet: pre-training supplies a strong prior that drives extreme confidence on members, widening their separation from non-members. In contrast, the baseline suffers from generalized overconfidence, obscuring the distributional distinguishability required for inference.

\subsection{Functional Complexity and Redundancy}
\label{sec:functional_analysis}
We next examine functional complexity and redundancy in relation to feature encoding and Model Stealing. We characterize these properties using Principal Component Analysis (PCA) and Centered Kernel Alignment (CKA).

\begin{figure}[t]
\centering
\begin{subfigure}{0.245\columnwidth}
\includegraphics[width=\columnwidth]{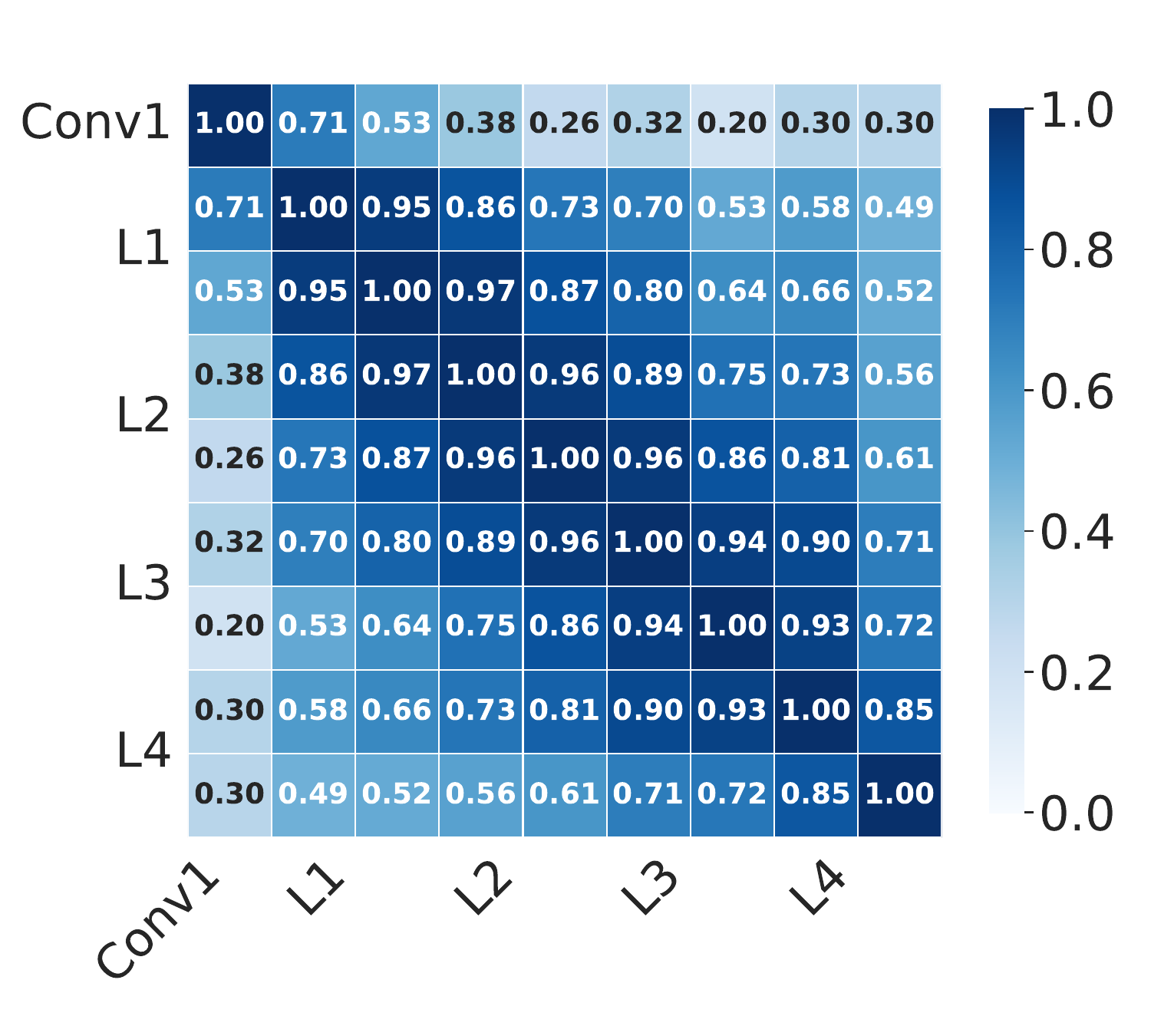}
\caption{High (CIFAR-10)}
\end{subfigure}
\begin{subfigure}{0.245\columnwidth}
\includegraphics[width=\columnwidth]{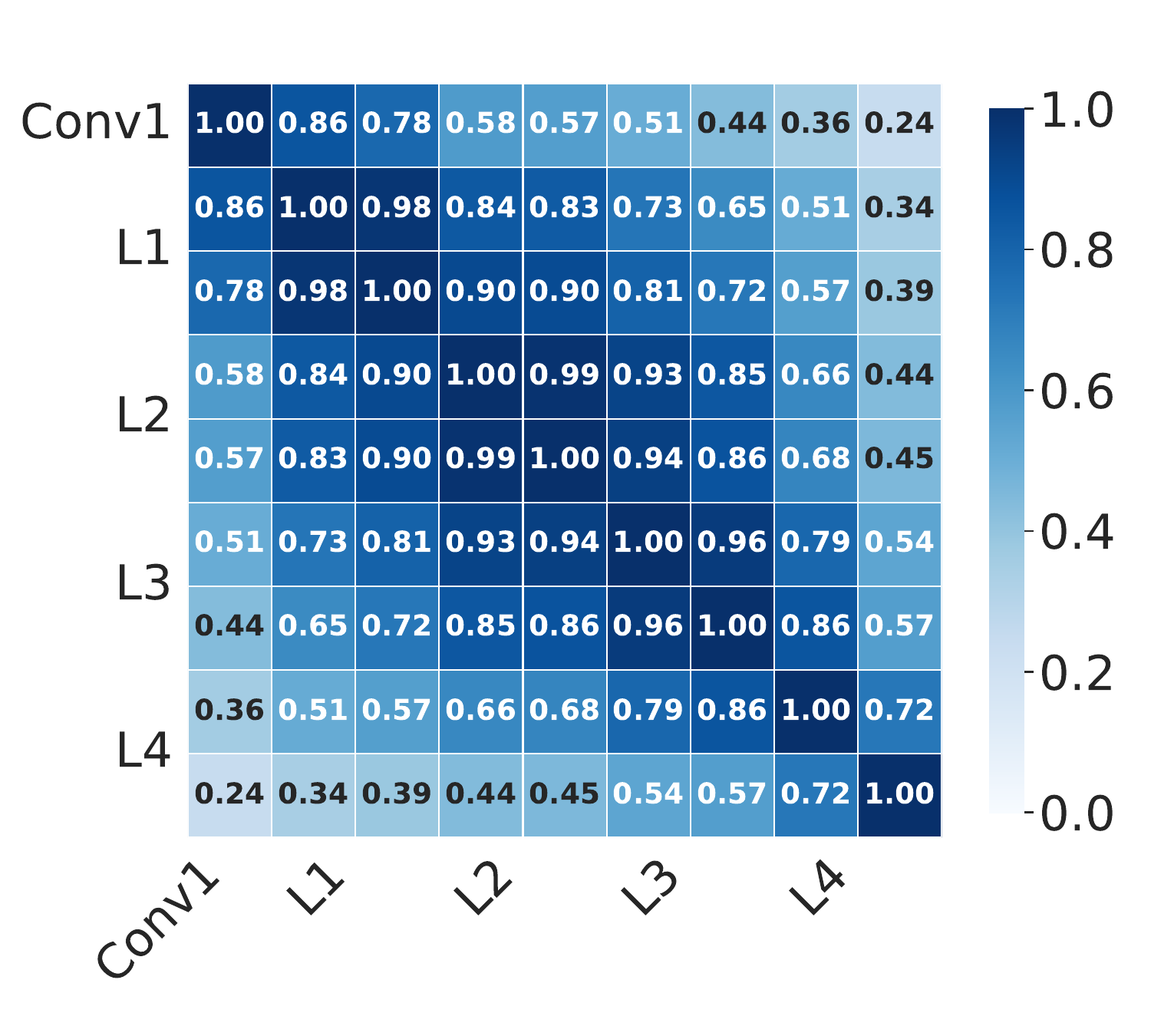}
\caption{Low (CIFAR-10)}
\end{subfigure}
\begin{subfigure}{0.245\columnwidth}
\includegraphics[width=\columnwidth]{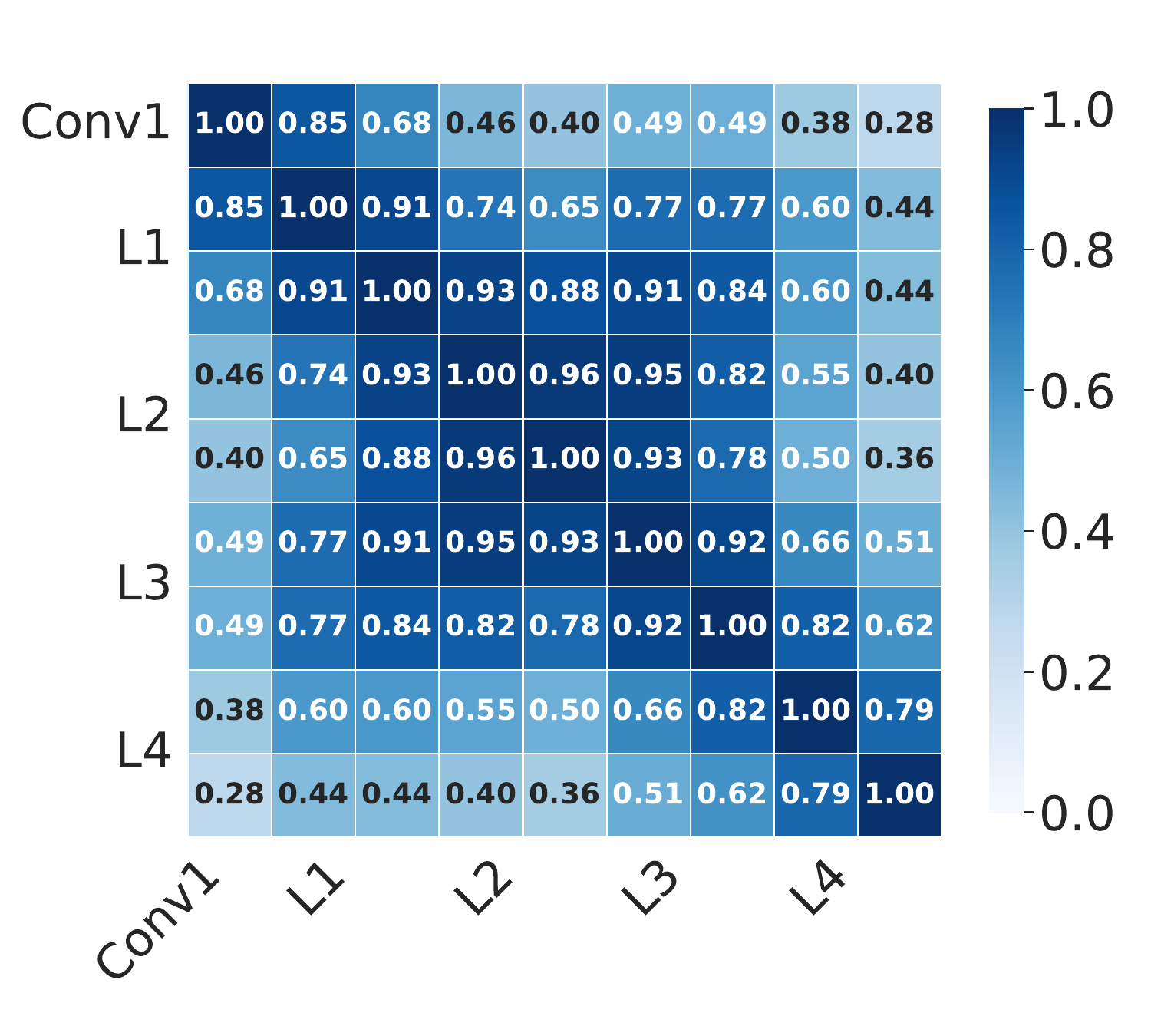}
\caption{High (TinyImageNet)}
\end{subfigure}
\begin{subfigure}{0.245\columnwidth}
\includegraphics[width=\columnwidth]{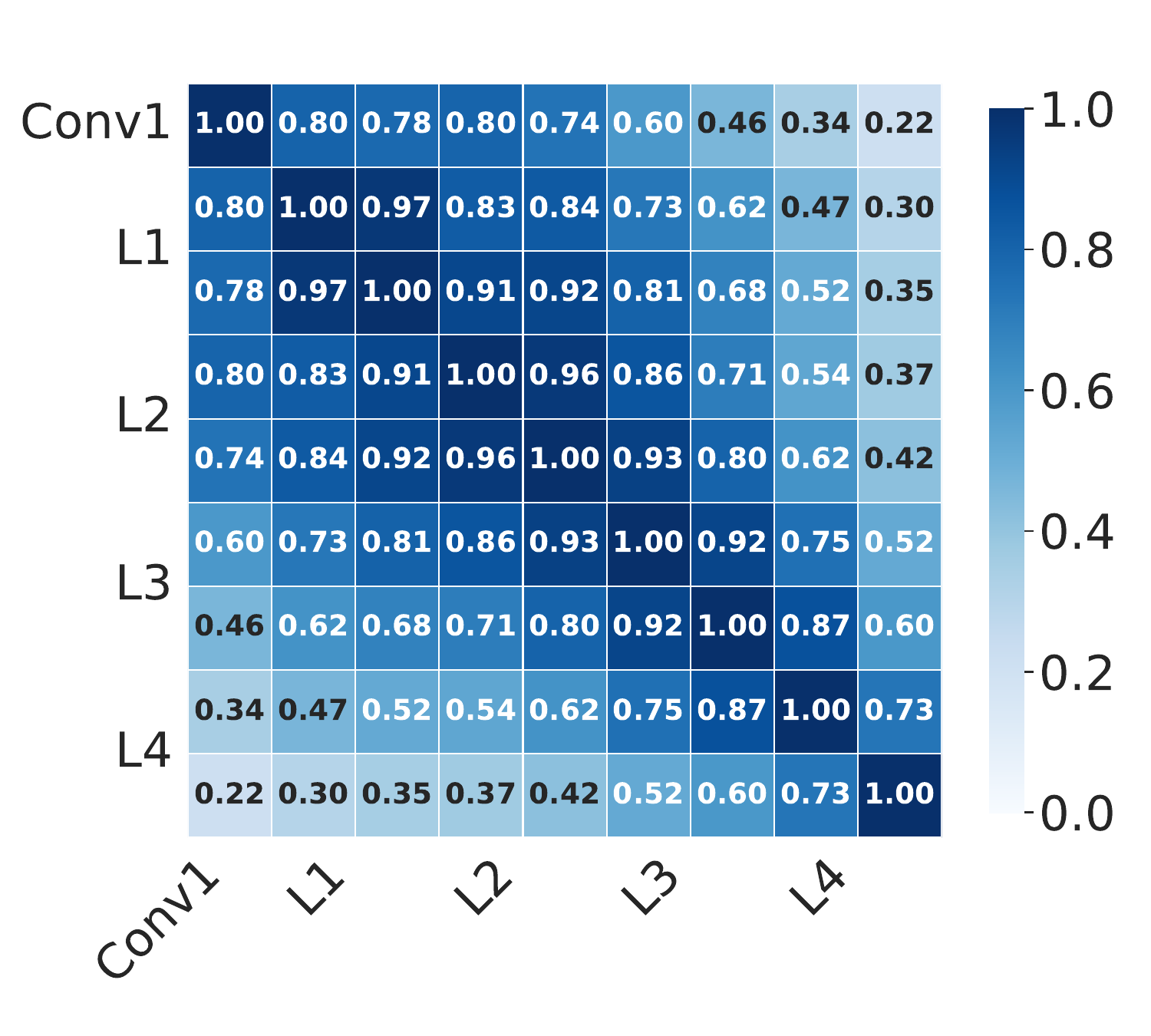}
\caption{Low (TinyImageNet)}
\end{subfigure}
\caption{Intra-CKA under Shapley-based data selection on CIFAR-10 and TinyImageNet.}
\label{fig:cka_cifar}
\end{figure}

\textbf{Feature Dimensionality (PCA).} 
We analyze Model Stealing through the effective feature dimension $d_{\text{eff}}$~\cite{MBW20,ALMZ19}, which we estimate using the PCA Participation Ratio (PR). Let $\lambda_i$ denote the eigenvalues of the empirical covariance $\mathbf{C}$ of centered penultimate-layer features. The participation ratio is $\operatorname{PR}(\mathbf{C})=(\sum_{i=1}^{d}\lambda_i)^2/\sum_{i=1}^{d}\lambda_i^2$. A lower PR indicates that feature variance is concentrated in fewer dominant directions.

This spectral interpretation helps explain the data-selection results. High-Shapley selection retains easy, representative samples that preserve task utility while filtering out harder intra-class variations. The learned representation can therefore preserve the information needed for the task while concentrating feature variance into fewer directions, reducing the effective complexity faced by a naive (scratch-initialized) attacker. Consistent with this mechanism, High-Shapley variants exhibit lower PR across all evaluated settings (Table~\ref{tab:sharp_ece_pca_CV}) and greater vulnerability to naive Model Stealing (Figure~\ref{fig:ms_shapley}).

Pre-training follows a different pathway. Rather than reducing feature dimensionality, pre-trained variants maintain higher PR and resist scratch-initialized surrogates. However, an informed attacker initialized from the same public checkpoint inherits their shared representation prior and primarily needs to recover the task-specific adaptation (Table~\ref{tab:mis_ms_pretrain}). These results identify compact feature geometry and shared initialization as distinct routes through which the evaluated efficiency strategies lower the empirical barrier to extraction.

\textbf{Representation Similarity (CKA).}
To quantify changes in the feature hierarchy, we employ linear CKA~\cite{KNLH19}, a standard representation-similarity metric. Let $X\in\mathbb{R}^{n\times d_1}$ and $Y\in\mathbb{R}^{n\times d_2}$ denote the activation matrices of two layers, with Gram matrices $K=XX^\top$ and $L=YY^\top$. Using the Hilbert--Schmidt Independence Criterion (HSIC), it is defined as
\begin{equation}
\operatorname{CKA}(K,L)
=
\frac{\operatorname{HSIC}(K,L)}
{\sqrt{\operatorname{HSIC}(K,K)\operatorname{HSIC}(L,L)}}.
\end{equation}

We interpret the High-Shapley pattern through simplicity bias~\cite{AJBKBKMFCBL17,PCL19}. By emphasizing easy, representative samples, data selection reduces the pressure for later layers to encode challenging intra-class variations. Accordingly, the Intra-CKA heatmaps in Figure~\ref{fig:cka_cifar} reveal more extensive late-layer self-similarity in High-Shapley models than in their Low-Shapley baselines, indicating that successive layers preserve similar representational geometry and contribute limited additional feature refinement. The lower PCA PR values in Table~\ref{tab:sharp_ece_pca_CV} reinforce this picture of a lower-dimensional, more homogeneous feature hierarchy and help explain the higher Model Stealing success observed in our experiments.

\section{Analysis on Large Language Models}
\label{sec:exp_llm}
We now extend our evaluation to math-reasoning LLMs to examine whether the vulnerability pattern observed in vision also appears in another model domain. We focus on recent ``Zero RL'' or ``Simple RL'' approaches, which reduce alignment overhead by starting directly from capable base models without preliminary SFT and simplifying conventional alignment components. Our analysis examines whether models produced by these efficiency-oriented pipelines exhibit analogous security and structural signatures.

\noindent
\begin{minipage}[t]{0.48\textwidth}
\vspace{0pt}
\centering
\begin{minipage}[t]{0.49\linewidth}
\centering
\includegraphics[width=\linewidth]{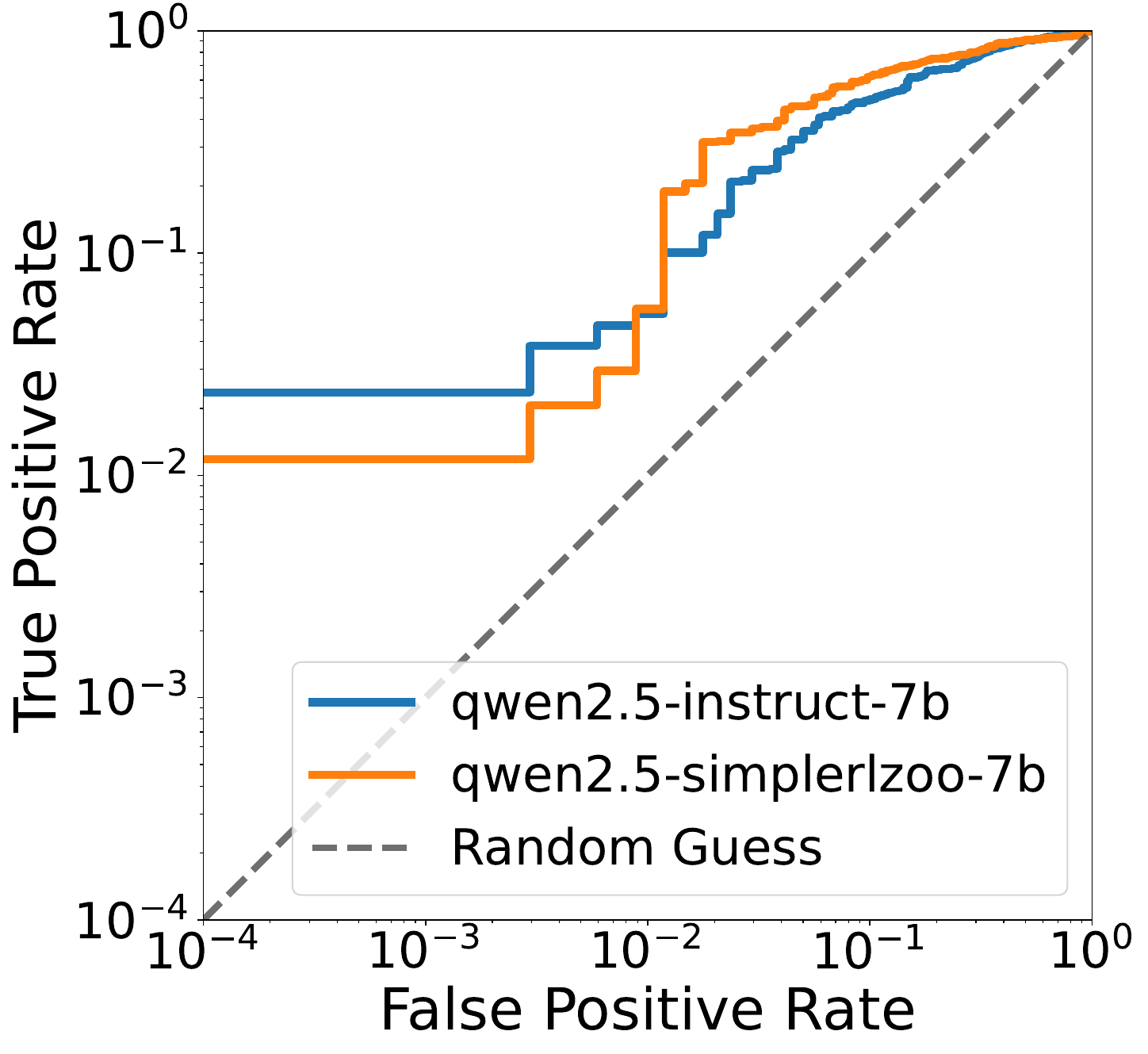}
\par\smallskip
\small (a) Answer likelihood
\end{minipage}\hfill
\begin{minipage}[t]{0.49\linewidth}
\centering
\includegraphics[width=\linewidth]{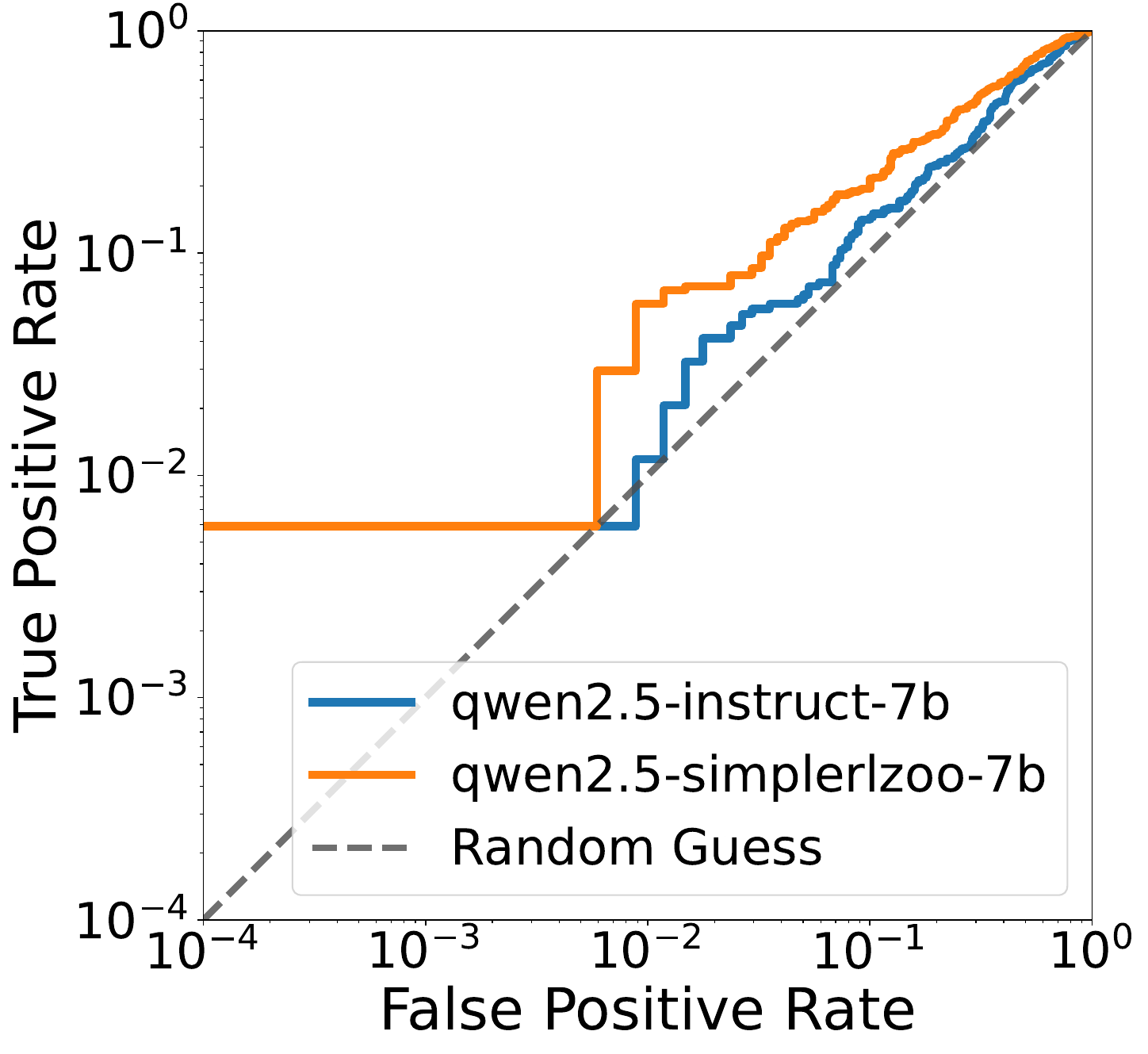}
\par\smallskip
\small (b) Confidence
\end{minipage}
\captionof{figure}{MIA for 7B math reasoning LLMs.}
\label{fig:mia_llm}
\end{minipage}\hfill
\begin{minipage}[t]{0.50\textwidth}
\vspace{0pt}
\centering
\small
\captionof{table}{Robustness and AutoDAN jailbreak results for 7B math reasoning LLMs. Higher accuracy indicates greater robustness; fewer steps and higher ASR indicate greater vulnerability.}
\vspace{-1mm}
\setlength{\tabcolsep}{5pt}
\begin{tabular}{l|cc|cc}
\toprule
\rowcolor{white}
\multirow{2}{*}{Model} & \multicolumn{2}{c|}{Robustness} & \multicolumn{2}{c}{AutoDAN} \\
\cmidrule(l{3pt}r{3pt}){2-3}\cmidrule(l{3pt}r{0pt}){4-5}
& Base & Math & Steps & ASR \\
\midrule
Instruct & 99.50 & 89.30 & 4.32 & 91.92\% \\
SimpleRL & 98.70 & 82.00 & 0.53 & 100.00\% \\
PRIME & 98.20 & 82.30 & 0.41 & 99.04\% \\
\bottomrule
\end{tabular}
\label{tab:adv_llm}
\end{minipage}
\par\medskip
To investigate this question, we conduct a comparative analysis between Qwen2.5-Math-7B-Instruct and two efficiency-oriented variants, Qwen2.5-7B-SimpleRL-Zoo and Eurus-2-7B-PRIME, obtained from Qwen2.5-Math-7B-Base through their respective zero-RL alignment pipelines. As shown in Figure~\ref{fig:mia_llm}, while the Instruct baseline exhibits greater leakage under answer likelihood at stringent low-FPR thresholds, this metric is sensitive to exact ground-truth phrasing and can therefore mask the vulnerability of the RL model, whose generated trajectories need not follow the static reference text after reward optimization. Self-confidence reveals the broader risk: the efficient RL variant is more susceptible than the baseline and exhibits pronounced overconfidence in its own generated trajectories. This reversal demonstrates that the efficient RL model is more vulnerable overall, with its leakage manifested primarily through overconfidence in generated reasoning rather than verbatim memorization.
Security differences also emerge in model integrity, as shown in Table~\ref{tab:adv_llm}. Although all three models perform comparably on the clean RobustBase dataset, the zero-RL models lose more accuracy on the adversarial RobustMath benchmark. AutoDAN on AdvBench provides a complementary jailbreak evaluation: SimpleRL and PRIME achieve higher attack success rates and require substantially fewer attack steps than the Instruct model. The clearest disparity appears under out-of-distribution fine-tuning on IMDB (Table~\ref{tab:finetune_llm}). The Instruct model largely retains its mathematical-reasoning performance, whereas both zero-RL models lose more reasoning accuracy, with catastrophic forgetting most pronounced for PRIME. Its average score falls from 53.7 to 20.8, indicating that its specialized capabilities are more readily overwritten in this setting. We additionally evaluate privacy using the Min-$k$\% probability attack and extend the Instruct--SimpleRL comparison to 32B, observing similar patterns across membership inference, adversarial robustness, jailbreak susceptibility, and fine-tuning stability; complete results are reported in Appendix~\ref{sec:additional_llm_results}.

\begin{table}[t]
\centering
\caption{Fine-tuning stability attack on IMDB dataset.}
\vspace{-2mm}
\setlength{\tabcolsep}{1.5pt}
\scalebox{0.8}{
\begin{tabular}{l|l|cccccccc}
\toprule
\rowcolor{white}
\multicolumn{2}{c}{Dataset} & GSM8K & mawps & Minerva Math & MATH 500 & Olympiad Bench & AIME24 & AMC23 & Avg. \\
\midrule
\multirow{2}{*}{Qwen2.5-Math-7B-Instruct} & Before & 95.38 & 98.60 & 31.62 & 77.20 & 33.93 & 3.33 & 52.50 & 56.08 \\
& After & 94.77 & 98.30 & 27.21 & 78.00 & 32.89 & 10.00 & 45.00 & 55.17 \\
\midrule
\multirow{2}{*}{Qwen2.5-7B-SimpleRL-Zoo} & Before & 93.40 & 97.09 & 25.74 & 75.40 & 30.37 & 10.00 & 45.00 & 53.86 \\
& After & 81.58 & 84.99 & 18.75 & 68.80 & 29.33 & 6.67 & 42.50 & 47.52 \\
\midrule
\multirow{2}{*}{Eurus-2-7B-PRIME} & Before & 89.69 & 96.22 & 25.37 & 70.60 & 32.00 & 6.67 & 55.00 & 53.65 \\
& After & 41.77 & 33.22 & 12.87 & 24.40 & 8.15 & 0.00 & 25.00 & 20.77 \\
\bottomrule
\end{tabular}
}
\label{tab:finetune_llm}
\end{table}

\begin{figure}[t]
\centering
\begin{subfigure}{0.25\columnwidth}
\includegraphics[width=\columnwidth]{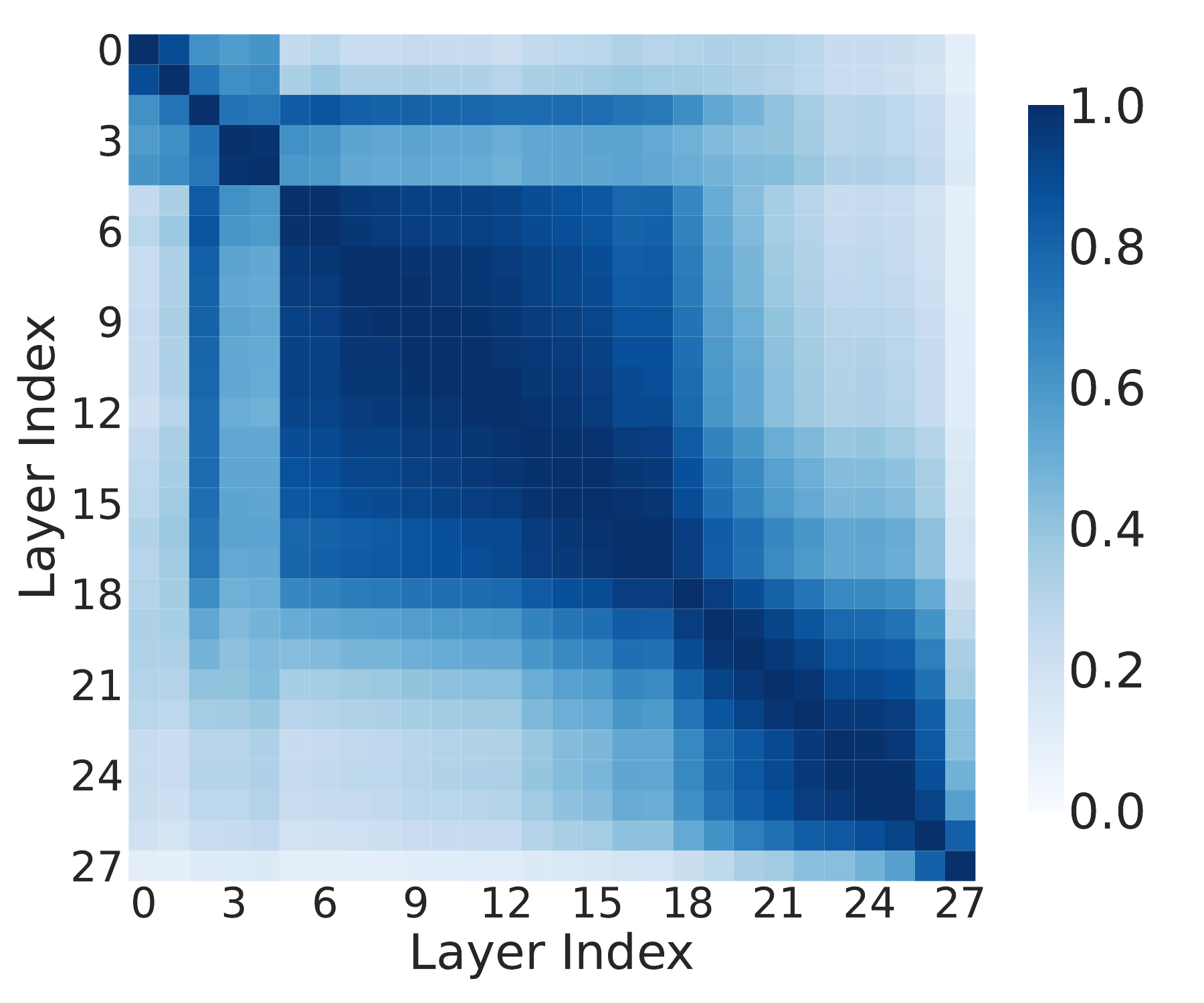}
\caption{Instruct}
\end{subfigure}
\begin{subfigure}{0.25\columnwidth}
\includegraphics[width=\columnwidth]{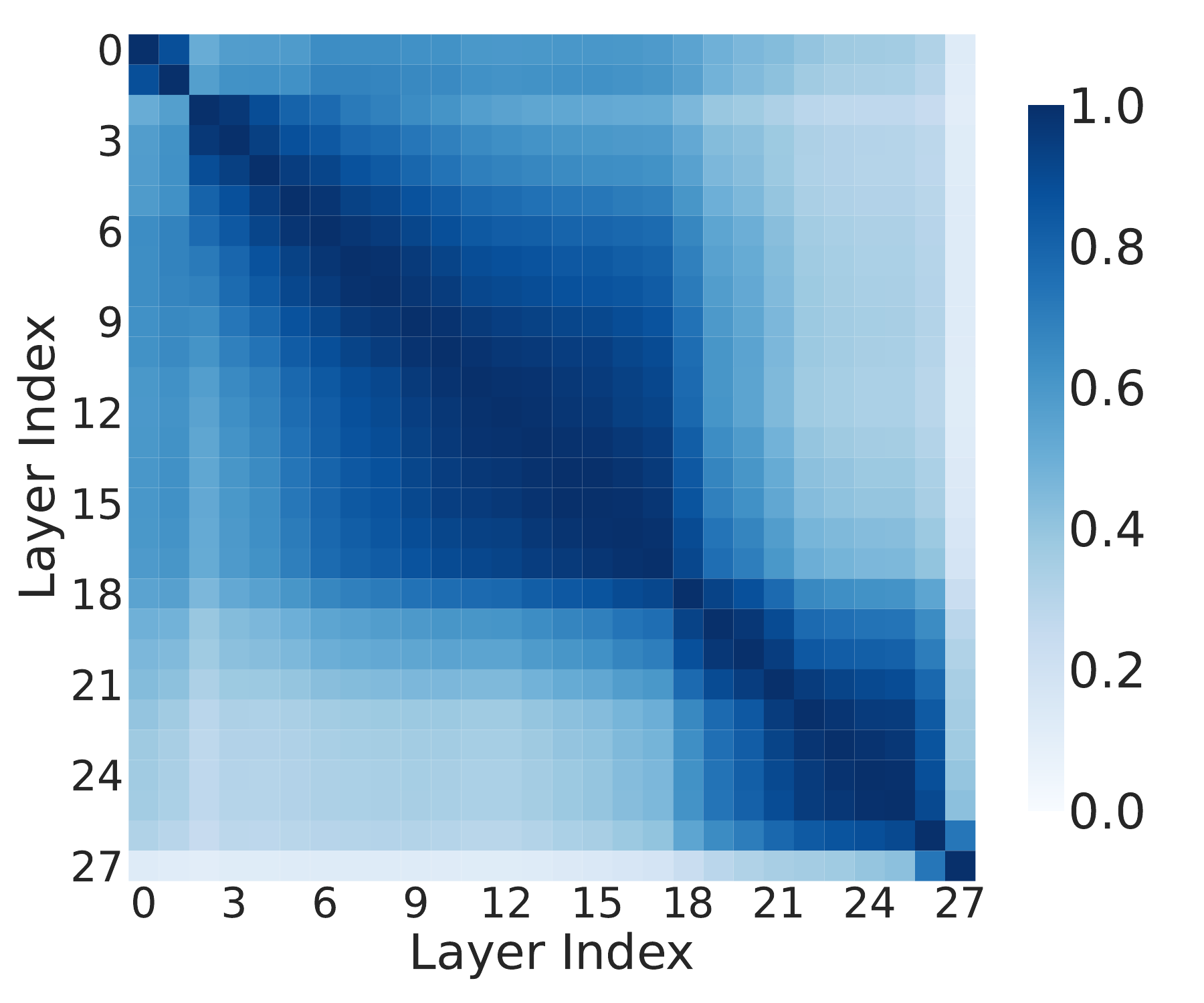}
\caption{SimpleRL}
\end{subfigure}
\begin{subfigure}{0.25\columnwidth}
\includegraphics[width=\columnwidth]{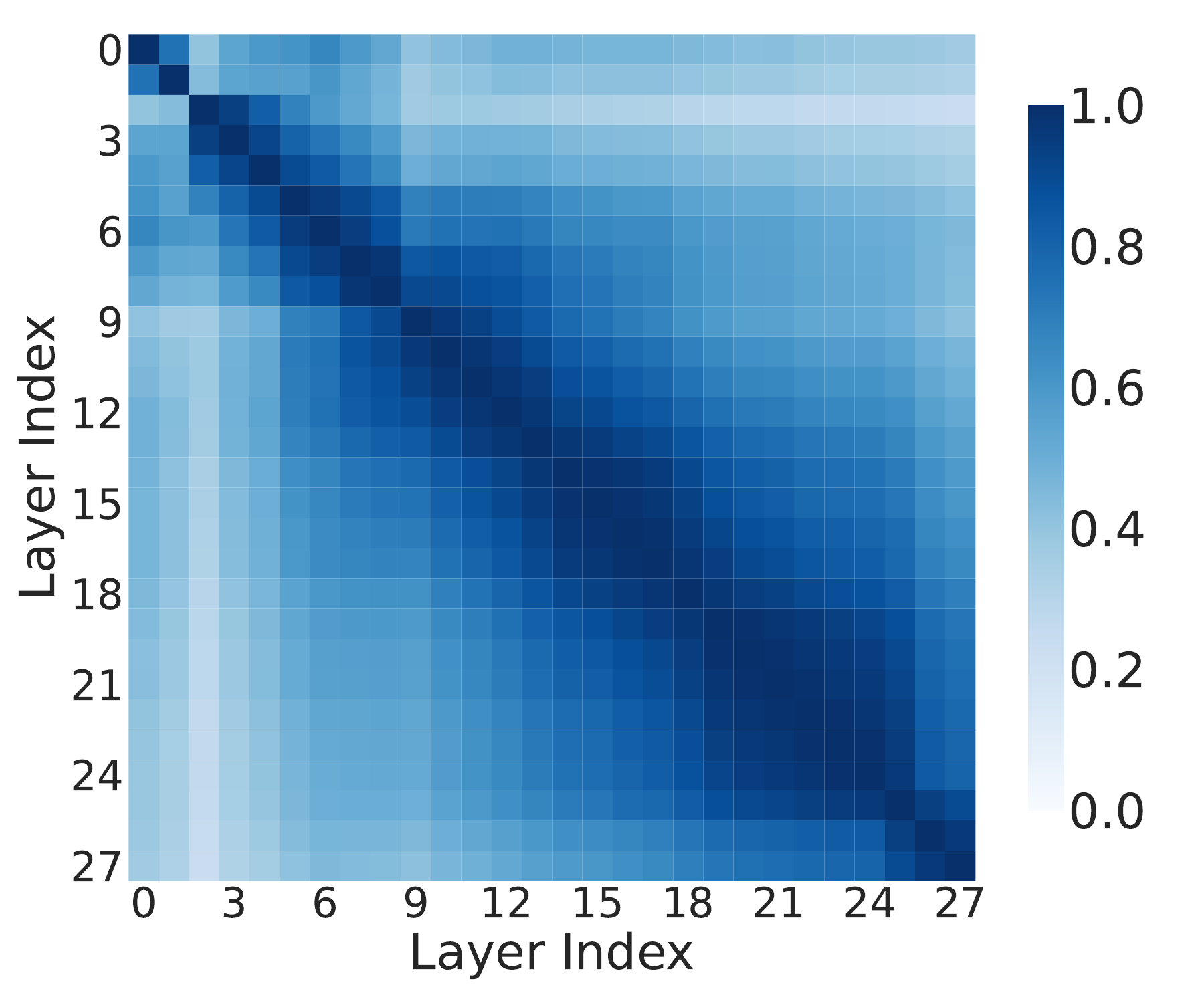}
\caption{Prime}
\end{subfigure}
\caption{Intra-CKA on GSM8K.}
\label{fig:cka_gsm8k}
\end{figure}

\begin{wraptable}{r}{0.5\textwidth}
\centering
\small
\vspace{-4.5mm}
\caption{Sharpness and ECE for 7B math reasoning LLMs.}
\vspace{-2mm}
\setlength{\tabcolsep}{1pt}
\begin{tabular}{l|cccc}
\toprule
\rowcolor{white} \multirow{2}{*}{Model} & \multicolumn{2}{c}{Sharpness} & \multicolumn{2}{c}{ECE} \\
\cmidrule(l{3pt}r{3pt}){2-3}\cmidrule(l{3pt}r{0pt}){4-5}
& GSM8K & MATH 500 & GSM8K & MATH 500 \\
\midrule
Instruct & 0.0615 & 0.0398 & 0.0375 & 0.4006 \\
SimpleRL & 0.9120 & 0.5183 & 0.0899 & 0.4453 \\
PRIME & 0.5565 & 0.5032 & 0.1073 & 0.4492 \\
\bottomrule
\end{tabular}
\label{tab:sharp_ece_llm}
\end{wraptable}
To elucidate the structural basis of the observed vulnerabilities, we shift from behavioral outputs to the models' internal geometry, calibration, and functional representations. Table~\ref{tab:sharp_ece_llm} reveals a pronounced geometric divergence: the Instruct model exhibits substantially flatter local geometry, whereas the zero-RL models have markedly higher sharpness. Explicit KL regularization ordinarily constrains policy deviation from a reference model; its absence from the evaluated zero-RL pipelines therefore offers a plausible optimization-level explanation for this sharper geometry. The efficient models also exhibit consistently higher ECE, indicating greater miscalibration and overconfidence. Functionally, the Intra-CKA analysis in Figure~\ref{fig:cka_gsm8k} reveals stronger adjacent-layer similarity earlier in the zero-RL models, suggesting a more compressed and less differentiated progression of representation processing. The same pattern appears on 500 examples sampled from the MATH test set in Appendix Figure~\ref{fig:cka_math}. These findings demonstrate that efficiency gains in the evaluated LLM alignment pipelines are accompanied by broader security vulnerability, with sharper local geometry and less differentiated representations providing a consistent structural explanation.

\section{Conclusion}
\label{sec:conclusion}
In this work, we presented the first systematic evaluation of the hidden security costs associated with efficiency-driven foundation model development. Across the evaluated vision and language settings, we show that strategies used to improve training efficiency, including importance-based data selection, pre-training, and simplified alignment, can compromise robustness and privacy even when task utility is preserved.

Our results reveal a recurring ``no free lunch'' trade-off: across the strategies studied, lower training cost is repeatedly accompanied by greater exposure to privacy leakage and adversarial attacks. The mechanistic analyses connect this pattern to sharper local loss geometry, greater miscalibration, and systematic changes in functional representations. These findings establish that efficiency gains can carry consequential security costs. As models continue to scale, we advocate for ``Robust Efficiency'', a paradigm shift toward training methods that integrate stability constraints to ensure models are both cheap and secure.

\bibliography{normal_generated_py3}
\bibliographystyle{iclr2027_conference}

\newpage
\appendix
\onecolumn

\section{Details of Theory}
\subsection{Loss Landscape Sharpness}
\label{app:sharpness_theory}
Let $\mathcal{U}_\rho=\{\delta:\|\delta_j\|_2\leq\rho\ \text{for every parameter tensor }j\}$ denote the tensor-wise perturbation set used by the procedure in Appendix~\ref{app:experimental_settings}. For batch $b$, define $g_b=\nabla_\theta\mathcal{L}_b(\theta)$ and $\mathcal{H}_b=\nabla_\theta^2\mathcal{L}_b(\theta)$. A local Taylor expansion gives
\begin{equation}
\mathcal{L}_b(\theta+\delta)-\mathcal{L}_b(\theta)
=g_b^\top\delta+\frac{1}{2}\delta^\top \mathcal{H}_b\delta+o(\|\delta\|_2^2).
\end{equation}
Thus, the measured loss increase reflects both the local gradient and curvature. In the stationary local-minimum case, where $g_b=0$ and $\mathcal{H}_b\succeq0$, maximizing over a single global $L_2$ ball yields $\frac{1}{2}\rho^2\lambda_{\max}(\mathcal{H}_b)+o(\rho^2)$; the tensor-wise constraint used in our experiments defines a different, implementation-aligned neighborhood. Projected gradient ascent approximates rather than exactly solves this maximization, so $\widehat S$ is an empirical measure of local loss sensitivity rather than an exact Hessian-eigenvalue estimate.

\subsection{Expected Calibration Error (ECE)}
For vision classification, we provide an idealized margin-based interpretation of the ECE behavior rather than a guarantee that an efficiency strategy must increase miscalibration. For a sample $x_i$, let $z_{ik}$ be the logit for class $k$ and $\hat y_i=\arg\max_k z_{ik}$ the predicted class. Defining the pairwise margins $\Delta_{ij}=z_{i\hat y_i}-z_{ij}$ for $j\neq\hat y_i$, the corresponding softmax confidence is
\begin{equation}
c_i
=
\frac{1}{1+\sum_{j\neq\hat y_i}\exp(-\Delta_{ij})}.
\end{equation}
As long as the predicted class is unchanged, confidence increases monotonically with every margin because
\begin{equation}
\frac{\partial c_i}{\partial \Delta_{ij}}
=
c_i^2\exp(-\Delta_{ij})>0.
\end{equation}
For separable data and positively homogeneous models, gradient-based optimization of exponential-tail losses can continue increasing parameter norms and classification margins even after the training error vanishes~\cite{SHNS18,LL20}. This asymptotic result provides theoretical intuition for increasingly extreme confidence, but does not by itself imply poorer calibration.

The connection to ECE becomes precise when confidence grows without a corresponding improvement in correctness. For an overconfident bin $B_m$ whose membership and accuracy remain fixed locally, its ECE contribution is
\begin{equation}
\mathrm{ECE}_m
=
\frac{|B_m|}{N}
\left(
\operatorname{conf}(B_m)-\operatorname{acc}(B_m)
\right),
\qquad
\frac{\partial\mathrm{ECE}_m}
{\partial\operatorname{conf}(B_m)}
=
\frac{|B_m|}{N}>0.
\end{equation}
Thus, margin growth enlarges the contribution of an already overconfident bin when its accuracy does not improve. This local argument does not require ECE to increase globally: predictions may move between bins, and greater confidence can reduce the gap in underconfident bins. It instead motivates our hypothesis that high-influence data selection and limited-data fine-tuning can aggravate miscalibration when their confidence shifts outpace changes in correctness, as assessed empirically in Table~\ref{tab:sharp_ece_pca_CV}. For LLMs, we retain the same bin-based ECE definition but use the sequence-level confidence specified in Appendix~\ref{app:experimental_settings}; the classification-margin derivation above is not applied to that confidence measure.

\subsection{Feature Dimensionality (PCA)}
Let $f(x_i)\in\mathbb{R}^d$ denote the penultimate-layer feature of sample $x_i$ and $\boldsymbol{\mu}=\frac{1}{N}\sum_{i=1}^{N}f(x_i)$ their empirical mean. We compute the covariance $\mathbf{C}=\frac{1}{N-1}\sum_{i=1}^{N}(f(x_i)-\boldsymbol{\mu})(f(x_i)-\boldsymbol{\mu})^\top$. For its eigenvalues $\lambda_i$, define the normalized spectrum $\widetilde{\lambda}_i=\lambda_i/\sum_{j=1}^{d}\lambda_j$. The PR defined in Section~\ref{sec:functional_analysis} is then equivalently $\operatorname{PR}(\mathbf{C})=(\sum_{i=1}^{d}\widetilde{\lambda}_i^2)^{-1}$. We use PR as a continuous estimate of the effective rank of the feature covariance: if variance is evenly distributed across $k$ directions, $d_{\mathrm{eff}}\approx k$; if a single direction dominates, $d_{\mathrm{eff}}\approx1$.

For a naive (scratch-initialized) attacker without a shared representation prior, the target's effective feature dimension characterizes how many independent directions must be approximated. High-Shapley selection preferentially retains easy, representative samples sufficient for task utility while omitting harder intra-class variations. If these omitted variations contribute primarily to the tail of the covariance spectrum, their removal concentrates variance in the leading eigenvalues and lowers PR. From an information-bottleneck perspective, the resulting representation encodes less within-class variation while preserving the information needed to match measured utility. Its lower effective dimension therefore provides a natural explanation for the higher stolen-model accuracy and agreement observed under naive extraction. This is an effective-complexity account of the empirical pattern rather than a universal linear query-complexity law.

The pre-training setting produces the opposite spectral pattern and requires a different explanation. Let $\theta_0$ denote the public pre-trained initialization and write the fine-tuned target as $\theta_T=\theta_0+\Delta_T$. A scratch-initialized attacker must learn both a useful representation and the target's task-specific behavior, whereas an informed surrogate initialized at $\theta_0$ already inherits the shared prior and primarily needs to recover the adaptation $\Delta_T$. This does not reduce extraction to learning only a linear head, but it narrows the empirical gap between the surrogate and target. The higher PR of the pre-trained variants is therefore consistent with their resistance to scratch-initialized attacks, while the shared initialization explains why this protection weakens under informed extraction.

\subsection{Representation Similarity (CKA)}
The CKA metric relies on the Hilbert--Schmidt Independence Criterion (HSIC) to measure statistical dependence between layer representations. Given activation matrices $X\in\mathbb{R}^{n\times d_1}$ and $Y\in\mathbb{R}^{n\times d_2}$, we compute the linear-kernel Gram matrices $K=XX^\top$ and $L=YY^\top$. Let $H=I_n-\frac{1}{n}\mathbf{1}\mathbf{1}^\top$ be the centering matrix and define $K_c=HKH$ and $L_c=HLH$. The empirical HSIC is
\begin{equation}
\operatorname{HSIC}(K,L)
=\frac{1}{(n-1)^2}\operatorname{Tr}(K_cL_c)
=\frac{1}{(n-1)^2}\operatorname{Tr}(KHLH).
\end{equation}
This value quantifies alignment between the inter-sample similarity structures of the two layers. For linear kernels, a value near 0 indicates little alignment between the centered Gram matrices, while higher values indicate stronger structural similarity. Normalizing HSIC yields a scalar in $[0,1]$ that is invariant to orthogonal transformations and isotropic scaling. Consequently, high CKA does not establish that one layer is a literal identity mapping of another; it shows that the layers preserve similar representational geometry.

We interpret extensive late-layer similarity through simplicity bias, the tendency of neural networks to rely on the simplest predictive features available~\cite{AJBKBKMFCBL17,PCL19}. High-Shapley selection emphasizes easy, representative samples while omitting harder intra-class variations. Once early features are sufficient to separate this selected distribution, later layers face less pressure to reorganize inter-sample geometry, producing the more homogeneous hierarchy observed in Figure~\ref{fig:cka_cifar}. Thus, high late-layer CKA provides evidence of limited additional feature refinement without requiring the stronger claim that successive layers implement exact identity mappings.

\section{More Results}
\subsection{Vision Models}
\label{sec:additional_vision_results}
\shortsection{Three-Seed Vision Results}
\label{sec:three_seed_vision_results}
This section reports the complete mean $\pm$ standard deviation results corresponding to the mean values presented in the main-text vision tables. Specifically, Appendix Table~\ref{tab:mia_shapley_three_seeds} expands the data-selection accuracy and membership-inference results in Table~\ref{tab:mia_shapley}; Appendix Table~\ref{tab:distance_three_seeds} expands the adversarial-robustness results in Table~\ref{tab:distance}; Appendix Table~\ref{tab:mis_ms_pretrain_three_seeds} expands the pre-trained fine-tuning results in Table~\ref{tab:mis_ms_pretrain}; and Appendix Table~\ref{tab:sharp_ece_pca_three_seeds} expands the sharpness, ECE, and PCA PR results in Table~\ref{tab:sharp_ece_pca_CV}.

\begin{table}[p]
\centering
\caption{Accuracy (training accuracy in brackets) and TPR at 0.1\% FPR of membership inference for data selection with Shapley values (mean $\pm$ standard deviation over three seeds).}
\setlength{\tabcolsep}{5pt}
\resizebox{\linewidth}{!}{
\begin{tabular}{ll|c|ccc}
\toprule
Dataset & Value & Acc & MIA (conf) & MIA (entr) & MIA (modi) \\
\midrule
\multirow{2}{*}{CIFAR-10} & Low & $48.07 \pm 0.47$ ($99.97 \pm 0.06$) & $0.12\% \pm 0.01\%$ & $0.05\% \pm 0.01\%$ & $0.11\% \pm 0.01\%$ \\
& High & $47.97 \pm 0.58$ ($99.97 \pm 0.05$) & $0.16\% \pm 0.02\%$ & $0.09\% \pm 0.01\%$ & $0.16\% \pm 0.01\%$ \\
\midrule
\multirow{2}{*}{CIFAR-10 (swin)} & Low & $37.01 \pm 0.29$ ($99.95 \pm 0.08$) & $0.08\% \pm 0.01\%$ & $0.05\% \pm 0.01\%$ & $0.09\% \pm 0.01\%$ \\
& High & $36.94 \pm 0.26$ ($99.93 \pm 0.07$) & $0.29\% \pm 0.02\%$ & $0.30\% \pm 0.02\%$ & $0.30\% \pm 0.01\%$ \\
\midrule
\multirow{2}{*}{TinyImageNet} & Low & $36.09 \pm 0.58$ ($100.00 \pm 0.00$) & $0.08\% \pm 0.02\%$ & $0.06\% \pm 0.01\%$ & $0.10\% \pm 0.02\%$ \\
& High & $36.17 \pm 0.46$ ($100.00 \pm 0.00$) & $0.22\% \pm 0.01\%$ & $0.35\% \pm 0.02\%$ & $0.27\% \pm 0.02\%$ \\
\midrule
\multirow{2}{*}{PubFig83} & Low & $44.01 \pm 0.55$ ($100.00 \pm 0.00$) & $0.32\% \pm 0.03\%$ & $0.32\% \pm 0.03\%$ & $0.32\% \pm 0.03\%$ \\
& High & $44.26 \pm 0.57$ ($100.00 \pm 0.00$) & $1.51\% \pm 0.06\%$ & $1.43\% \pm 0.07\%$ & $1.51\% \pm 0.06\%$ \\
\bottomrule
\end{tabular}
}
\label{tab:mia_shapley_three_seeds}
\end{table}

\begin{table}[p]
\centering
\caption{Adversarial robustness for vision models (mean $\pm$ standard deviation over three seeds).}
\setlength{\tabcolsep}{6pt}
\resizebox{0.82\linewidth}{!}{
\begin{tabular}{ll|c|l|c}
\toprule
Dataset & Value & Robustness & Paradigm & Robustness \\
\midrule
\multirow{2}{*}{CIFAR-10} & Low & $3.85 \pm 0.21$ & Scratch & $3.58 \pm 0.58$ \\
& High & $1.33 \pm 0.07$ & Pre-train & $1.00 \pm 0.05$ \\
\midrule
\multirow{2}{*}{CIFAR-10 (swin)} & Low & $0.70 \pm 0.13$ & Scratch & $3.22 \pm 0.11$ \\
& High & $0.18 \pm 0.03$ & Pre-train & $0.11 \pm 0.03$ \\
\midrule
\multirow{2}{*}{TinyImageNet} & Low & $20.89 \pm 2.00$ & Scratch & $14.03 \pm 1.70$ \\
& High & $10.46 \pm 1.21$ & Pre-train & $3.64 \pm 0.13$ \\
\midrule
\multirow{2}{*}{PubFig83} & Low & $2.34 \pm 0.25$ & Scratch & $1.83 \pm 0.11$ \\
& High & $1.30 \pm 0.08$ & Pre-train & $0.41 \pm 0.07$ \\
\bottomrule
\end{tabular}
}
\label{tab:distance_three_seeds}
\end{table}

\begin{table}[p]
\centering
\caption{Accuracy (training accuracy in brackets), membership inference, and model stealing for pre-trained fine-tuning and training from scratch (mean $\pm$ standard deviation over three seeds). Values in parentheses under MS denote agreement.}
\setlength{\tabcolsep}{3pt}
\resizebox{\linewidth}{!}{
\begin{tabular}{ll|c|ccc|cc}
\toprule
\multirow{2}{*}{Dataset} & \multirow{2}{*}{Paradigm} & \multirow{2}{*}{Acc} & \multicolumn{3}{c|}{MIA} & \multicolumn{2}{c}{MS} \\
& & & conf & entr & modi & 1000 & 1000 (Pre-train) \\
\midrule
\multirow{2}{*}{CIFAR-10} & Scratch & $67.71 \pm 0.36$ ($99.91 \pm 0.09$) & $0.05\% \pm 0.03\%$ & $0.04\% \pm 0.01\%$ & $0.05\% \pm 0.02\%$ & $62.60 \pm 0.27$ ($0.748 \pm 0.005$) & $65.22 \pm 0.20$ ($0.758 \pm 0.004$) \\
& Pre-train & $67.88 \pm 0.24$ ($100.00 \pm 0.00$) & $0.21\% \pm 0.01\%$ & $0.16\% \pm 0.01\%$ & $0.21\% \pm 0.00\%$ & $58.05 \pm 0.16$ ($0.669 \pm 0.007$) & $67.25 \pm 0.14$ ($0.785 \pm 0.003$) \\
\midrule
\multirow{2}{*}{CIFAR-10 (swin)} & Scratch & $73.85 \pm 0.35$ ($100.00 \pm 0.00$) & $0.12\% \pm 0.01\%$ & $0.12\% \pm 0.01\%$ & $0.11\% \pm 0.01\%$ & $42.41 \pm 0.42$ ($0.446 \pm 0.006$) & $57.70 \pm 0.52$ ($0.536 \pm 0.009$) \\
& Pre-train & $74.01 \pm 0.14$ ($100.00 \pm 0.00$) & $0.54\% \pm 0.02\%$ & $0.54\% \pm 0.02\%$ & $0.54\% \pm 0.02\%$ & $39.01 \pm 0.24$ ($0.437 \pm 0.010$) & $60.06 \pm 0.19$ ($0.651 \pm 0.006$) \\
\midrule
\multirow{2}{*}{TinyImageNet} & Scratch & $34.91 \pm 0.75$ ($99.98 \pm 0.02$) & $0.13\% \pm 0.01\%$ & $0.07\% \pm 0.03\%$ & $0.14\% \pm 0.03\%$ & $24.65 \pm 0.45$ ($0.381 \pm 0.004$) & $27.53 \pm 0.43$ ($0.379 \pm 0.008$) \\
& Pre-train & $34.93 \pm 0.38$ ($100.00 \pm 0.00$) & $0.75\% \pm 0.01\%$ & $0.75\% \pm 0.01\%$ & $0.67\% \pm 0.01\%$ & $17.65 \pm 0.43$ ($0.305 \pm 0.006$) & $29.82 \pm 0.73$ ($0.529 \pm 0.010$) \\
\midrule
\multirow{2}{*}{PubFig83} & Scratch & $35.19 \pm 0.25$ ($100.00 \pm 0.00$) & $0.01\% \pm 0.01\%$ & $0.01\% \pm 0.01\%$ & $0.01\% \pm 0.01\%$ & $17.81 \pm 0.73$ ($0.358 \pm 0.008$) & $18.59 \pm 0.40$ ($0.299 \pm 0.001$) \\
& Pre-train & $34.79 \pm 0.33$ ($100.00 \pm 0.00$) & $0.36\% \pm 0.01\%$ & $0.46\% \pm 0.02\%$ & $0.37\% \pm 0.01\%$ & $7.89 \pm 0.50$ ($0.183 \pm 0.006$) & $25.26 \pm 0.16$ ($0.533 \pm 0.010$) \\
\bottomrule
\end{tabular}
}
\label{tab:mis_ms_pretrain_three_seeds}
\end{table}

\begin{table}[p]
\centering
\caption{Sharpness, Expected Calibration Error (ECE), and PCA Participation Ratio (PCA PR) for vision models (mean $\pm$ standard deviation over three seeds).}
\setlength{\tabcolsep}{4pt}
\resizebox{\linewidth}{!}{
\begin{tabular}{ll|ccc|l|ccc}
\toprule
Dataset & Value & Sharpness & ECE & PCA PR & Paradigm & Sharpness & ECE & PCA PR \\
\midrule
\multirow{2}{*}{CIFAR-10} & Low & $0.0410 \pm 0.0146$ & $0.382 \pm 0.012$ & $28.490 \pm 7.284$ & Scratch & $0.0311 \pm 0.0116$ & $0.219 \pm 0.005$ & $14.266 \pm 0.160$ \\
& High & $0.0591 \pm 0.0141$ & $0.430 \pm 0.014$ & $10.793 \pm 0.727$ & Pre-train & $0.1332 \pm 0.0088$ & $0.232 \pm 0.001$ & $25.137 \pm 0.962$ \\
\midrule
\multirow{2}{*}{CIFAR-10 (swin)} & Low & $0.0170 \pm 0.0005$ & $0.410 \pm 0.009$ & $13.907 \pm 0.861$ & Scratch & $0.0010 \pm 0.0002$ & $0.060 \pm 0.004$ & $9.897 \pm 1.008$ \\
& High & $0.0189 \pm 0.0015$ & $0.436 \pm 0.008$ & $10.896 \pm 1.633$ & Pre-train & $0.0373 \pm 0.0056$ & $0.241 \pm 0.036$ & $26.786 \pm 0.668$ \\
\midrule
\multirow{2}{*}{TinyImageNet} & Low & $0.0548 \pm 0.0025$ & $0.336 \pm 0.021$ & $265.298 \pm 13.393$ & Scratch & $0.0511 \pm 0.0017$ & $0.317 \pm 0.002$ & $157.477 \pm 0.655$ \\
& High & $0.0600 \pm 0.0028$ & $0.384 \pm 0.019$ & $74.960 \pm 3.202$ & Pre-train & $0.1201 \pm 0.0033$ & $0.251 \pm 0.008$ & $163.206 \pm 1.082$ \\
\midrule
\multirow{2}{*}{PubFig83} & Low & $0.1234 \pm 0.0054$ & $0.208 \pm 0.017$ & $48.395 \pm 1.200$ & Scratch & $0.1349 \pm 0.0052$ & $0.350 \pm 0.009$ & $31.940 \pm 0.914$ \\
& High & $0.1475 \pm 0.0075$ & $0.299 \pm 0.014$ & $33.936 \pm 2.311$ & Pre-train & $0.2551 \pm 0.0061$ & $0.384 \pm 0.008$ & $39.009 \pm 0.107$ \\
\bottomrule
\end{tabular}
}
\label{tab:sharp_ece_pca_three_seeds}
\end{table}

\begin{table}[p]
\centering
\caption{Accuracy (training accuracy in brackets) and membership inference for Full-dataset training and High-Shapley subsets (mean $\pm$ standard deviation over three seeds). MIA is reported as TPR at 0.1\% FPR; higher values indicate greater privacy leakage.}
\setlength{\tabcolsep}{5pt}
\resizebox{\linewidth}{!}{
\begin{tabular}{ll|c|ccc}
\toprule
Dataset & Value & Acc & MIA (conf) & MIA (entr) & MIA (modi) \\
\midrule
\multirow{3}{*}{CIFAR-10} & Full & $94.84 \pm 0.32$ ($100.00 \pm 0.00$) & $0.02\% \pm 0.00\%$ & $0.01\% \pm 0.00\%$ & $0.01\% \pm 0.00\%$ \\
& High (20,000) & $88.54 \pm 0.25$ ($100.00 \pm 0.00$) & $0.07\% \pm 0.00\%$ & $0.06\% \pm 0.00\%$ & $0.07\% \pm 0.01\%$ \\
& High (40,000) & $93.38 \pm 0.27$ ($100.00 \pm 0.00$) & $0.04\% \pm 0.01\%$ & $0.02\% \pm 0.00\%$ & $0.04\% \pm 0.00\%$ \\
\midrule
\multirow{3}{*}{CIFAR-10 (swin)} & Full & $79.40 \pm 0.31$ ($100.00 \pm 0.00$) & $0.05\% \pm 0.00\%$ & $0.00\% \pm 0.00\%$ & $0.05\% \pm 0.00\%$ \\
& High (20,000) & $69.86 \pm 0.29$ ($100.00 \pm 0.00$) & $0.20\% \pm 0.01\%$ & $0.28\% \pm 0.01\%$ & $0.20\% \pm 0.01\%$ \\
& High (40,000) & $77.91 \pm 0.33$ ($100.00 \pm 0.00$) & $0.18\% \pm 0.01\%$ & $0.22\% \pm 0.00\%$ & $0.18\% \pm 0.01\%$ \\
\midrule
\multirow{3}{*}{TinyImageNet} & Full & $59.82 \pm 0.25$ ($100.00 \pm 0.00$) & $0.00\% \pm 0.00\%$ & $0.00\% \pm 0.00\%$ & $0.00\% \pm 0.00\%$ \\
& High (60,000) & $58.01 \pm 0.20$ ($100.00 \pm 0.00$) & $0.21\% \pm 0.02\%$ & $0.12\% \pm 0.02\%$ & $0.21\% \pm 0.02\%$ \\
& High (80,000) & $60.19 \pm 0.22$ ($100.00 \pm 0.00$) & $0.22\% \pm 0.01\%$ & $0.12\% \pm 0.01\%$ & $0.21\% \pm 0.01\%$ \\
\midrule
\multirow{3}{*}{PubFig83} & Full & $72.41 \pm 0.42$ ($100.00 \pm 0.00$) & $0.00\% \pm 0.00\%$ & $0.00\% \pm 0.00\%$ & $0.00\% \pm 0.00\%$ \\
& High (3,000) & $61.45 \pm 0.44$ ($100.00 \pm 0.00$) & $0.12\% \pm 0.02\%$ & $0.12\% \pm 0.02\%$ & $0.12\% \pm 0.02\%$ \\
& High (5,000) & $65.66 \pm 0.45$ ($100.00 \pm 0.00$) & $0.73\% \pm 0.05\%$ & $0.82\% \pm 0.06\%$ & $1.90\% \pm 0.09\%$ \\
\bottomrule
\end{tabular}
}
\label{tab:full_data_mia}
\end{table}
\begin{table}[p]
\centering
\caption{Adversarial robustness and model stealing using 1,000 distinct query images (500 for PubFig83) for Full-dataset training and High-Shapley subsets (mean $\pm$ standard deviation over three seeds). Higher robustness is better. MS reports stolen-model accuracy, with agreement in parentheses; higher values indicate a more successful stealing attack.}
\setlength{\tabcolsep}{7pt}
\resizebox{0.82\linewidth}{!}{
\begin{tabular}{ll|c|c}
\toprule
Dataset & Value & Robustness & MS \\
\midrule
\multirow{3}{*}{CIFAR-10} & Full & $18.60 \pm 0.92$ & $59.25 \pm 0.23$ ($0.599 \pm 0.002$) \\
& High (20,000) & $10.03 \pm 0.88$ & $67.05 \pm 0.29$ ($0.712 \pm 0.003$) \\
& High (40,000) & $17.60 \pm 0.82$ & $62.01 \pm 0.26$ ($0.634 \pm 0.002$) \\
\midrule
\multirow{3}{*}{CIFAR-10 (swin)} & Full & $5.68 \pm 0.22$ & $41.20 \pm 0.20$ ($0.429 \pm 0.003$) \\
& High (20,000) & $2.57 \pm 0.11$ & $45.56 \pm 0.21$ ($0.499 \pm 0.004$) \\
& High (40,000) & $4.85 \pm 0.19$ & $43.10 \pm 0.19$ ($0.458 \pm 0.003$) \\
\midrule
\multirow{3}{*}{TinyImageNet} & Full & $36.93 \pm 1.62$ & $19.39 \pm 0.11$ ($0.215 \pm 0.006$) \\
& High (60,000) & $29.77 \pm 2.11$ & $28.66 \pm 0.12$ ($0.362 \pm 0.009$) \\
& High (80,000) & $34.85 \pm 1.73$ & $22.93 \pm 0.18$ ($0.267 \pm 0.007$) \\
\midrule
\multirow{3}{*}{PubFig83} & Full & $3.32 \pm 0.11$ & $37.71 \pm 0.39$ ($0.396 \pm 0.009$) \\
& High (3,000) & $1.27 \pm 0.09$ & $44.70 \pm 0.35$ ($0.547 \pm 0.010$) \\
& High (5,000) & $1.67 \pm 0.13$ & $42.17 \pm 0.41$ ($0.520 \pm 0.012$) \\
\bottomrule
\end{tabular}
}
\label{tab:full_data_adv_ms}
\end{table}
\begin{table}[p]
\centering
\caption{Sharpness, Expected Calibration Error (ECE), and PCA Participation Ratio (PCA PR) for Full-dataset training and High-Shapley subsets (mean $\pm$ standard deviation over three seeds).}
\setlength{\tabcolsep}{7pt}
\resizebox{0.88\linewidth}{!}{
\begin{tabular}{ll|ccc}
\toprule
Dataset & Value & Sharpness & ECE & PCA PR \\
\midrule
\multirow{3}{*}{CIFAR-10} & Full & $0.0253 \pm 0.0078$ & $0.033 \pm 0.003$ & $10.544 \pm 0.231$ \\
& High (20,000) & $0.0313 \pm 0.0081$ & $0.076 \pm 0.004$ & $9.586 \pm 0.292$ \\
& High (40,000) & $0.0265 \pm 0.0077$ & $0.042 \pm 0.003$ & $9.841 \pm 0.221$ \\
\midrule
\multirow{3}{*}{CIFAR-10 (swin)} & Full & $0.0128 \pm 0.0004$ & $0.160 \pm 0.002$ & $9.048 \pm 0.201$ \\
& High (20,000) & $0.0152 \pm 0.0006$ & $0.245 \pm 0.003$ & $8.424 \pm 0.102$ \\
& High (40,000) & $0.0142 \pm 0.0003$ & $0.185 \pm 0.002$ & $8.316 \pm 0.162$ \\
\midrule
\multirow{3}{*}{TinyImageNet} & Full & $0.0440 \pm 0.0020$ & $0.200 \pm 0.002$ & $221.403 \pm 1.982$ \\
& High (60,000) & $0.0482 \pm 0.0024$ & $0.220 \pm 0.005$ & $158.711 \pm 2.109$ \\
& High (80,000) & $0.0458 \pm 0.0032$ & $0.209 \pm 0.004$ & $193.468 \pm 2.187$ \\
\midrule
\multirow{3}{*}{PubFig83} & Full & $0.1550 \pm 0.0043$ & $0.038 \pm 0.009$ & $41.872 \pm 1.021$ \\
& High (3,000) & $0.1296 \pm 0.0038$ & $0.116 \pm 0.008$ & $41.028 \pm 0.987$ \\
& High (5,000) & $0.1594 \pm 0.0041$ & $0.165 \pm 0.008$ & $49.366 \pm 1.223$ \\
\bottomrule
\end{tabular}
}
\label{tab:full_data_mechanistic}
\end{table}
\begin{table}[t]
\centering
\caption{Accuracy degradation under 50\% unstructured weight pruning (mean $\pm$ standard deviation over three seeds). ``Drop'' denotes the decrease from pre-pruning to post-pruning accuracy; higher values indicate greater pruning sensitivity.}
\setlength{\tabcolsep}{8pt}
\begin{tabular}{ll|ccc}
\toprule
Dataset & Value & Before & After & Drop \\
\midrule
\multirow{2}{*}{CIFAR-10} & Low & $48.07 \pm 0.47$ & $44.83 \pm 0.38$ & $3.24 \pm 0.60$ \\
& High & $47.97 \pm 0.58$ & $34.12 \pm 0.32$ & $13.85 \pm 0.66$ \\
\midrule
\multirow{2}{*}{TinyImageNet} & Low & $36.09 \pm 0.58$ & $34.54 \pm 0.44$ & $1.55 \pm 0.73$ \\
& High & $36.17 \pm 0.46$ & $18.87 \pm 0.61$ & $17.30 \pm 0.76$ \\
\bottomrule
\end{tabular}
\label{tab:pruning_sensitivity}
\end{table}
\begin{table}[t]
\centering
\caption{Membership inference for the 7B models using the Min-$k$\% probability attack with $k=15$. Higher TPR and AUC indicate greater membership leakage.}
\setlength{\tabcolsep}{9pt}
\begin{tabular}{l|cc}
\toprule
Model & TPR at 0.1\% FPR & AUC \\
\midrule
Qwen2.5-Math-7B-Instruct & 0.89\% & 0.823 \\
Qwen2.5-7B-SimpleRL-Zoo & 1.77\% & 0.857 \\
\bottomrule
\end{tabular}
\label{tab:min_k_mia}
\end{table}
\begin{table}[p]
\centering
\caption{Mathematical-reasoning performance before and after 4-bit NF4 quantization. Higher values are better.}
\setlength{\tabcolsep}{4pt}
\resizebox{\linewidth}{!}{
\begin{tabular}{ll|cccccccc}
\toprule
Model & State & GSM8K & MAWPS & Minerva & MATH 500 & Olympiad & AIME24 & AMC23 & Avg. \\
\midrule
\multirow{2}{*}{Instruct} & Before & 95.38 & 98.60 & 31.62 & 77.20 & 33.93 & 3.33 & 52.50 & 56.08 \\
& After & 95.00 & 98.60 & 31.25 & 79.20 & 33.48 & 6.67 & 47.50 & 55.96 \\
\midrule
\multirow{2}{*}{SimpleRL} & Before & 93.40 & 97.09 & 25.74 & 75.40 & 30.37 & 10.00 & 45.00 & 53.86 \\
& After & 90.30 & 97.29 & 26.10 & 71.20 & 30.96 & 6.67 & 40.00 & 51.78 \\
\midrule
\multirow{2}{*}{PRIME} & Before & 89.69 & 96.22 & 25.37 & 70.60 & 32.00 & 6.67 & 55.00 & 53.65 \\
& After & 80.44 & 77.87 & 24.26 & 66.00 & 31.56 & 6.67 & 47.50 & 47.76 \\
\bottomrule
\end{tabular}
}
\label{tab:nf4_quantization}
\end{table}
\shortsection{Comparison with Full-Dataset Training}
\label{sec:full_dataset_comparison}
Our primary data-selection study compares a small High-Shapley subset with a larger Low-Shapley subset calibrated to achieve similar clean accuracy. This utility-matched design evaluates vulnerability while controlling for predictive performance. Here, we provide a complementary comparison against conventional Full-dataset training. Unlike the primary study, these comparisons are not necessarily utility matched; they instead evaluate the practical security behavior of High-Shapley subsets at two different sizes relative to using all available training data.

Table~\ref{tab:full_data_mia} reports clean accuracy and membership-inference results. Across all evaluated datasets and architectures, the High-Shapley subsets exhibit higher membership-inference TPRs than Full-dataset training under confidence, entropy, and modified-entropy attacks. The differences remain present for the larger High-Shapley subsets, including settings in which their clean accuracy approaches that of Full-dataset training.

Table~\ref{tab:full_data_adv_ms} shows the same behavioral pattern for adversarial robustness and model stealing. Models trained on High-Shapley subsets remain correctly classified for fewer PGD iterates than their Full-data counterparts. Using the same number of distinct query images, the corresponding stolen models also attain consistently higher accuracy and agreement, indicating greater extraction susceptibility in these evaluated settings.

Table~\ref{tab:full_data_mechanistic} provides the corresponding structural measurements. ECE is higher for every High-Shapley subset than for Full-dataset training. Sharpness is generally higher for the High-Shapley models, although the difference becomes small for some larger subsets and reverses for PubFig83 with 3,000 samples. PCA PR is lower for the CIFAR-10 and TinyImageNet High-Shapley models, whereas the PubFig83 results are mixed. We therefore treat these measurements as complementary evidence rather than a uniform structural pattern across every dataset and subset size.

\shortsection{Sensitivity to Weight Pruning}
\label{sec:pruning_sensitivity}
We additionally evaluate sensitivity to 50\% unstructured weight pruning in two representative vision settings. As shown in Table~\ref{tab:pruning_sensitivity}, the High-Shapley models and their utility-matched Low-Shapley controls achieve comparable accuracy before pruning. After pruning, however, the High-Shapley models experience substantially larger performance degradation. This result provides supplementary evidence consistent with reduced functional redundancy in models trained on small, high-importance subsets.

\subsection{Large Language Models}
\label{sec:additional_llm_results}
\shortsection{Min-$k$\% Membership Inference on Math-Reasoning LLMs}
\label{sec:min_k_mia}
We complement the answer-likelihood and self-confidence attacks in the main text with the Min-$k$\% probability membership-inference attack, using $k=15$. For each sequence, this attack selects the 15\% of tokens assigned the lowest probabilities by the model and aggregates their log probabilities into a membership score. The attack is motivated by the tendency of models to assign higher probabilities to sequences encountered during training; here, we evaluate it using the source-defined groups specified in Appendix~\ref{app:experimental_settings}. We report the area under the receiver operating characteristic curve (AUC) and the true-positive rate (TPR) at a stringent false-positive rate (FPR) of 0.1\%.

As shown in Table~\ref{tab:min_k_mia}, the 7B SimpleRL checkpoint exhibits higher membership leakage than the Instruct checkpoint: the TPR increases from 0.89\% to 1.77\%, and the AUC increases from 0.823 to 0.857. This result is consistent with the privacy trend observed using self-confidence and provides additional evidence that the evaluated simplified-RL checkpoint is more susceptible to membership inference. Because this comparison uses existing checkpoints rather than a controlled component-level ablation, we interpret the result as an association with the simplified alignment recipe rather than as evidence that any single omitted training component directly causes the increased leakage.

\shortsection{Sensitivity to Low-Bit Quantization}
\label{sec:quantization_sensitivity}
We evaluate the sensitivity of the 7B math-reasoning models to 4-bit NormalFloat (NF4) quantization as a supplementary deployment-stage stress test. Table~\ref{tab:nf4_quantization} reports performance before and after quantization across seven reasoning benchmarks. Individual benchmark scores exhibit some variation in both directions, so we focus on the aggregate performance change rather than requiring uniform degradation on every task.

The Instruct model retains nearly all of its average performance after quantization, decreasing from 56.08 to 55.96. In comparison, SimpleRL decreases from 53.86 to 51.78, while PRIME decreases from 53.65 to 47.76. The larger aggregate degradation of the simplified-RL checkpoints provides supplementary evidence consistent with greater sensitivity to parameter perturbations. We treat this result as a quantization-sensitivity observation rather than as a direct security vulnerability or a general claim about all efficiently aligned models.

\shortsection{Evaluation at the 32B Scale}
\label{sec:llm_32b}
We extend the comparison to Qwen2.5-32B-Instruct and Qwen2.5-32B-SimpleRL-Zoo to examine whether the vulnerability pattern observed at 7B persists at a larger model scale. Table~\ref{tab:llm_32b_security} combines confidence-based and Min-$k$\% membership inference, adversarial robustness, and AutoDAN jailbreak results. SimpleRL exhibits higher membership leakage under both attacks. Both models achieve identical accuracy on RobustBase, while SimpleRL performs worse on RobustMath. Under AutoDAN, SimpleRL also attains a higher attack success rate with substantially fewer optimization steps. These results follow the same overall pattern as the 7B evaluation across complementary threat models.

\begin{table}[t]
\centering
\caption{Membership inference, adversarial robustness, and AutoDAN jailbreak susceptibility for 32B models. Confidence-based MIA reports TPR at 1.0\% FPR; Min-$k$\% MIA ($k=15$) reports TPR at 0.1\% FPR and AUC. Higher robustness accuracy is better; higher MIA TPR, AUC, and ASR and fewer AutoDAN steps indicate greater vulnerability.}
\setlength{\tabcolsep}{6pt}
\begin{tabular}{l|c|cc|cc|cc}
\toprule
\multirow{2}{*}{Model} & Conf. MIA & \multicolumn{2}{c|}{Min-$k$\% MIA} & \multicolumn{2}{c|}{Robustness} & \multicolumn{2}{c}{AutoDAN} \\
\cmidrule(l{3pt}r{3pt}){2-2}\cmidrule(l{3pt}r{3pt}){3-4}\cmidrule(l{3pt}r{3pt}){5-6}\cmidrule(l{3pt}r{0pt}){7-8}
& TPR & TPR & AUC & Base & Math & Steps & ASR \\
\midrule
Instruct & 0.59\% & 0.59\% & 0.802 & 99.50 & 97.00 & 1.32 & 98.08\% \\
SimpleRL & 0.88\% & 1.77\% & 0.833 & 99.50 & 93.70 & 0.23 & 100.00\% \\
\bottomrule
\end{tabular}
\label{tab:llm_32b_security}
\end{table}

We additionally evaluate stability under out-of-distribution fine-tuning on IMDB. As shown in Table~\ref{tab:llm_32b_finetuning}, the Instruct checkpoint retains most of its mathematical-reasoning performance after adaptation, whereas SimpleRL exhibits a substantially larger average decline across the seven benchmarks. This larger-scale result is consistent with the greater fine-tuning instability observed for the 7B simplified-RL checkpoints. Because the comparison uses existing checkpoints rather than a controlled training ablation, we interpret the result as an association with the evaluated alignment recipes.

\begin{table}[p]
\centering
\caption{Fine-tuning stability of Qwen2.5 32B models under IMDB adaptation. ``Before'' and ``After'' report mathematical-reasoning benchmark accuracy before and after fine-tuning.}
\setlength{\tabcolsep}{4pt}
\resizebox{\linewidth}{!}{
\begin{tabular}{ll|cccccccc}
\toprule
Model & State & GSM8K & MAWPS & Minerva & MATH 500 & Olympiad & AIME24 & AMC23 & Avg. \\
\midrule
\multirow{2}{*}{Instruct} & Before & 95.91 & 98.31 & 42.28 & 80.40 & 37.78 & 13.33 & 62.50 & 61.49 \\
& After & 95.07 & 98.31 & 37.87 & 77.60 & 36.30 & 10.00 & 52.50 & 58.24 \\
\midrule
\multirow{2}{*}{SimpleRL} & Before & 95.91 & 98.31 & 31.99 & 77.60 & 39.70 & 10.00 & 52.50 & 58.00 \\
& After & 85.37 & 96.66 & 21.69 & 50.20 & 17.93 & 3.33 & 30.00 & 43.60 \\
\bottomrule
\end{tabular}
}
\label{tab:llm_32b_finetuning}
\end{table}

\subsection{Additional Analysis}
\label{sec:additional_analysis}
\begin{figure}[t]
\centering
\begin{subfigure}{0.245\columnwidth}
\includegraphics[width=\columnwidth]{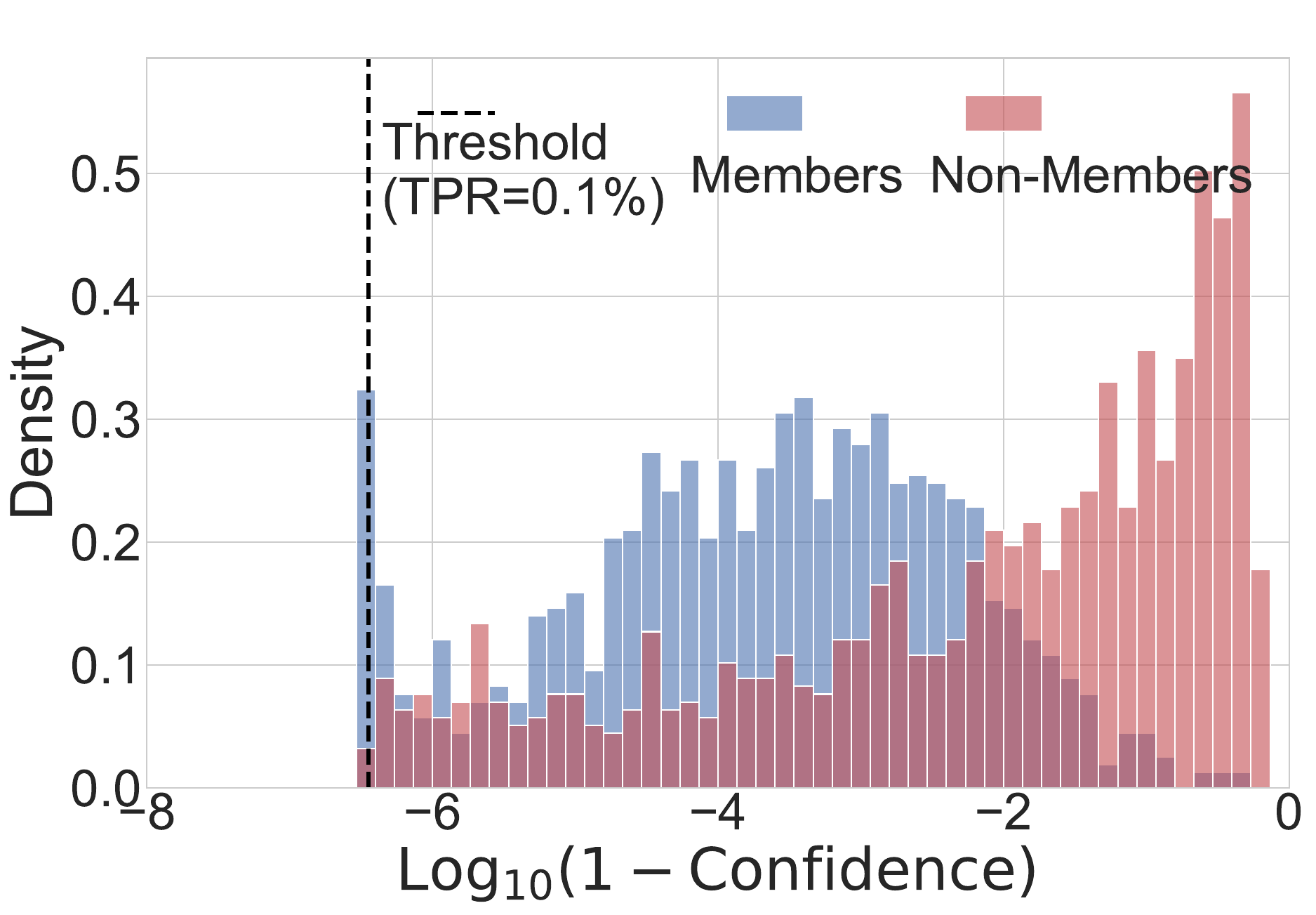}
\caption{High (confidence)}
\end{subfigure}
\begin{subfigure}{0.245\columnwidth}
\includegraphics[width=\columnwidth]{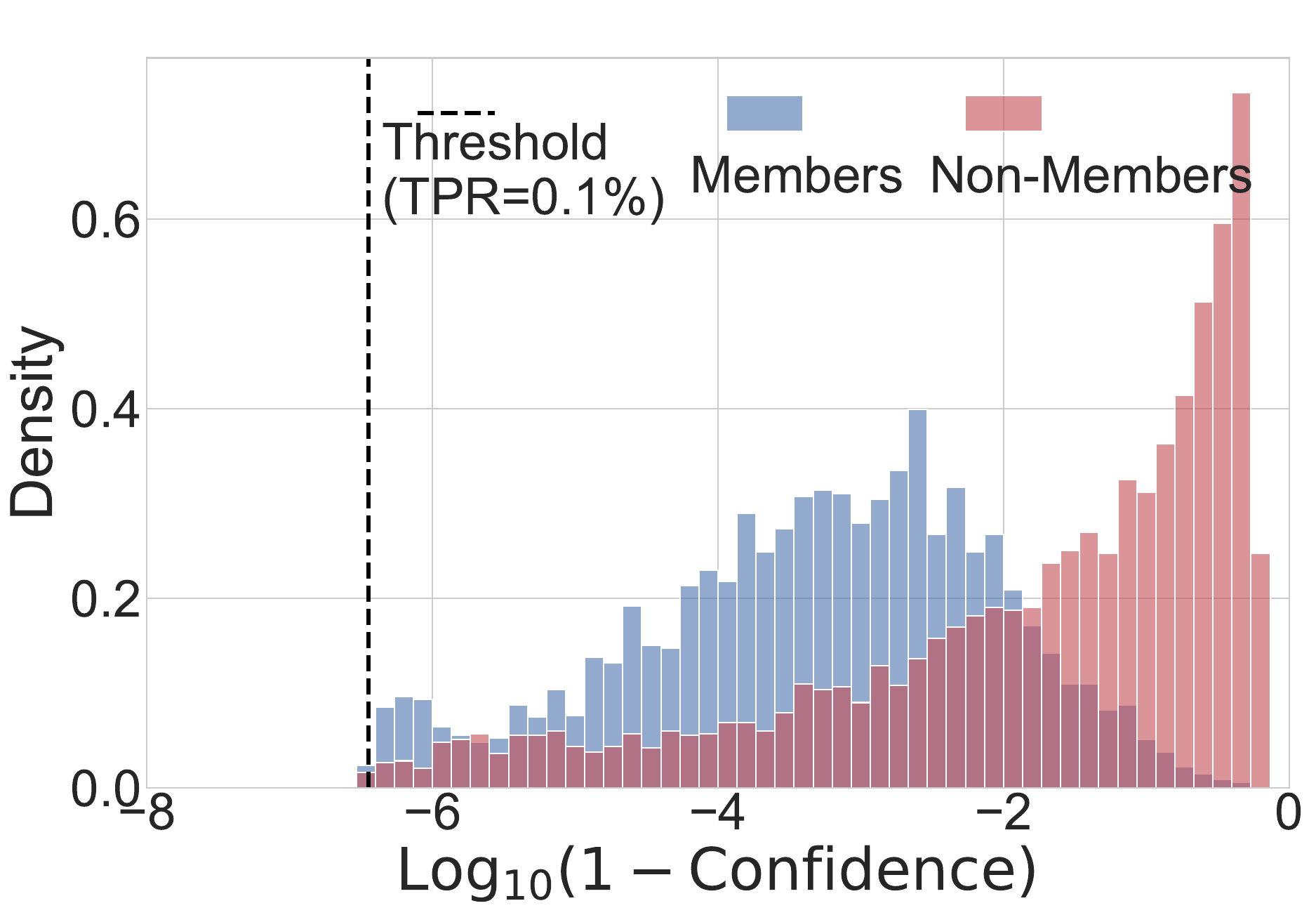}
\caption{Low (confidence)}
\end{subfigure}
\begin{subfigure}{0.245\columnwidth}
\includegraphics[width=\columnwidth]{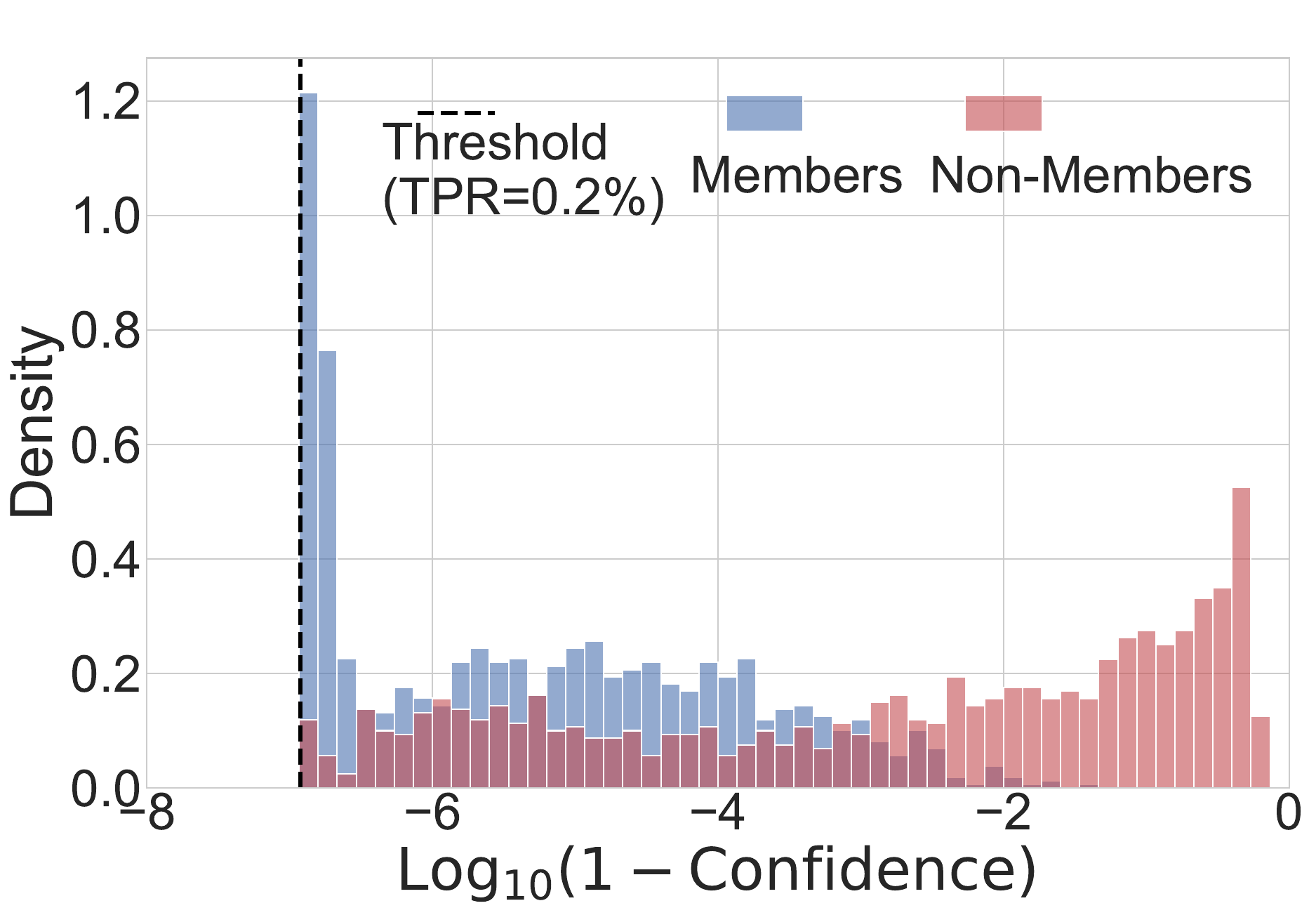}
\caption{Pre-train (confidence)}
\end{subfigure}
\begin{subfigure}{0.245\columnwidth}
\includegraphics[width=\columnwidth]{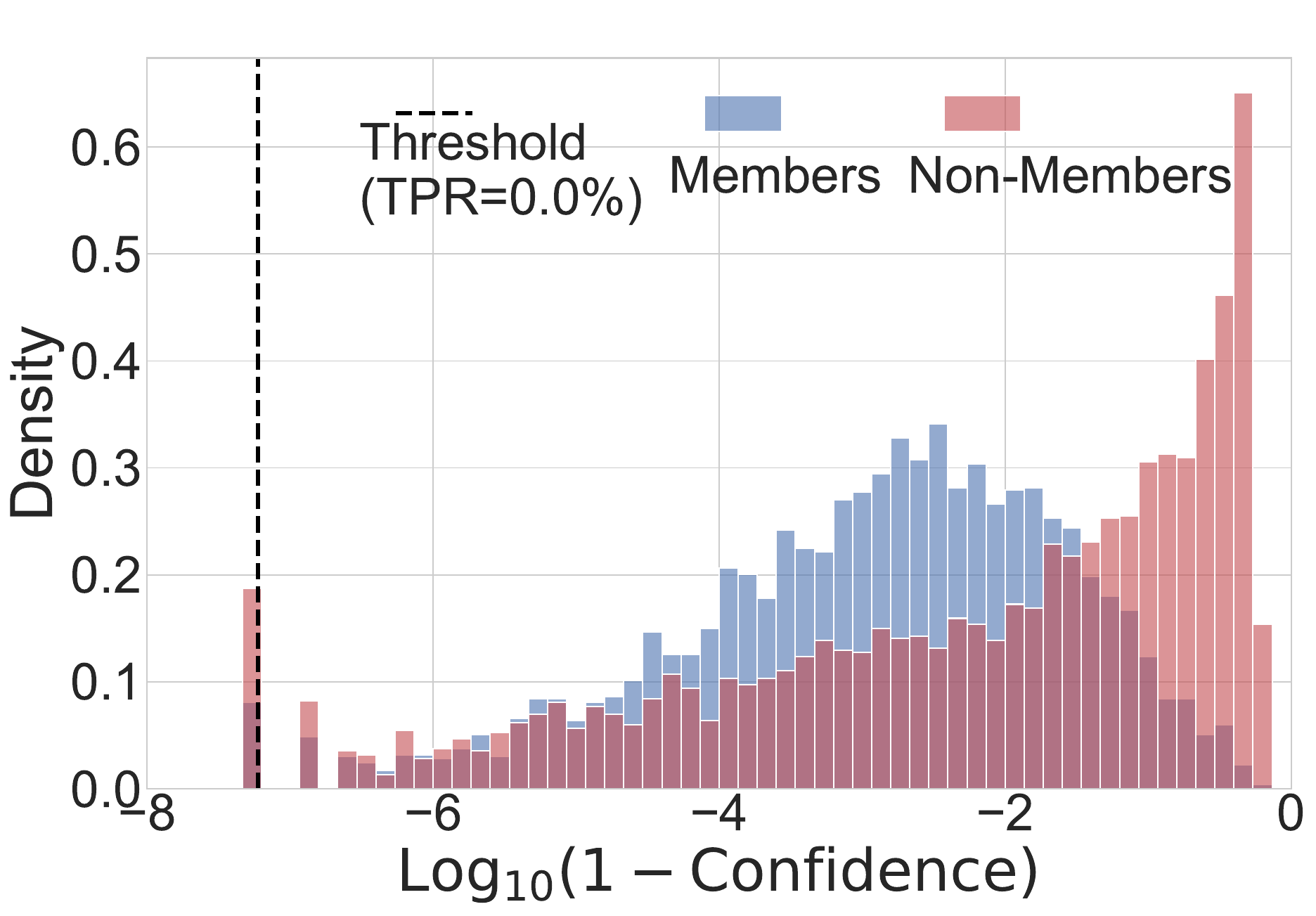}
\caption{Scratch (confidence)}
\end{subfigure}
\caption{Member and non-member confidence distributions on a logarithmic scale for CIFAR-10 with ResNet-18 under data selection and pre-training.}
\label{fig:loss_confidence_cifar}
\end{figure}

\begin{figure}[t]
\centering
\begin{subfigure}{0.245\columnwidth}
\includegraphics[width=\columnwidth]{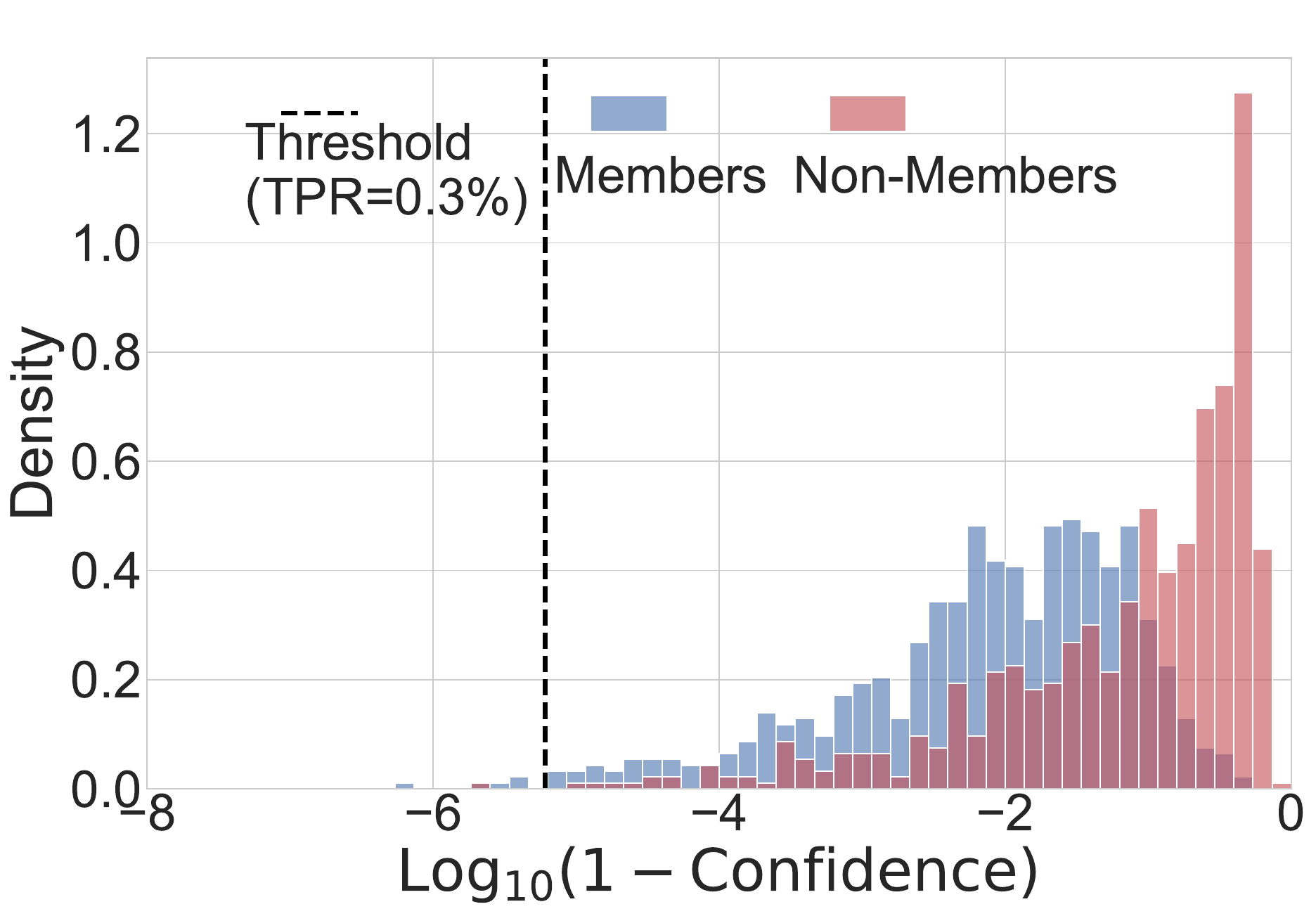}
\caption{High (confidence)}
\end{subfigure}
\begin{subfigure}{0.245\columnwidth}
\includegraphics[width=\columnwidth]{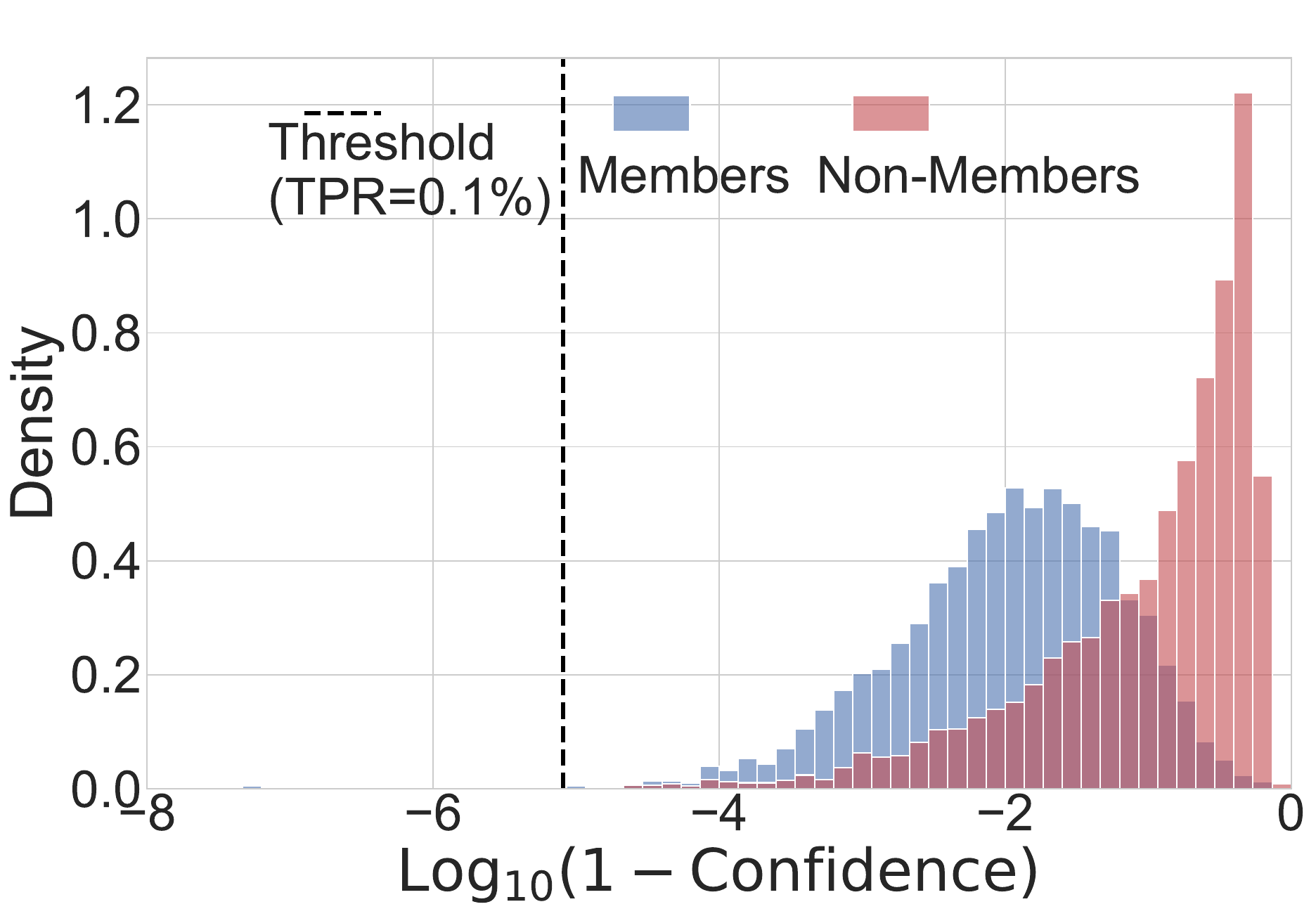}
\caption{Low (confidence)}
\end{subfigure}
\begin{subfigure}{0.245\columnwidth}
\includegraphics[width=\columnwidth]{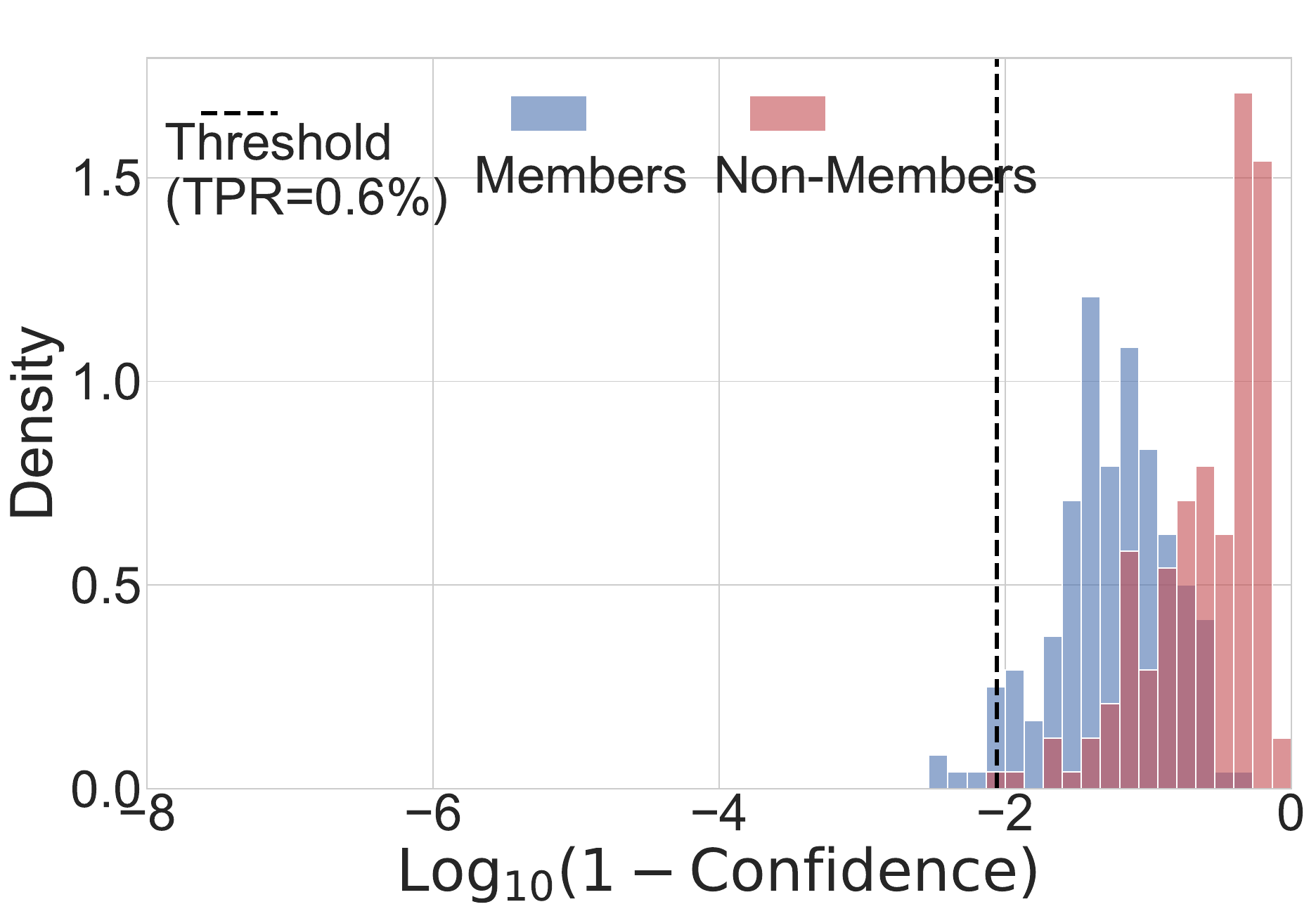}
\caption{Pre-train (confidence)}
\end{subfigure}
\begin{subfigure}{0.245\columnwidth}
\includegraphics[width=\columnwidth]{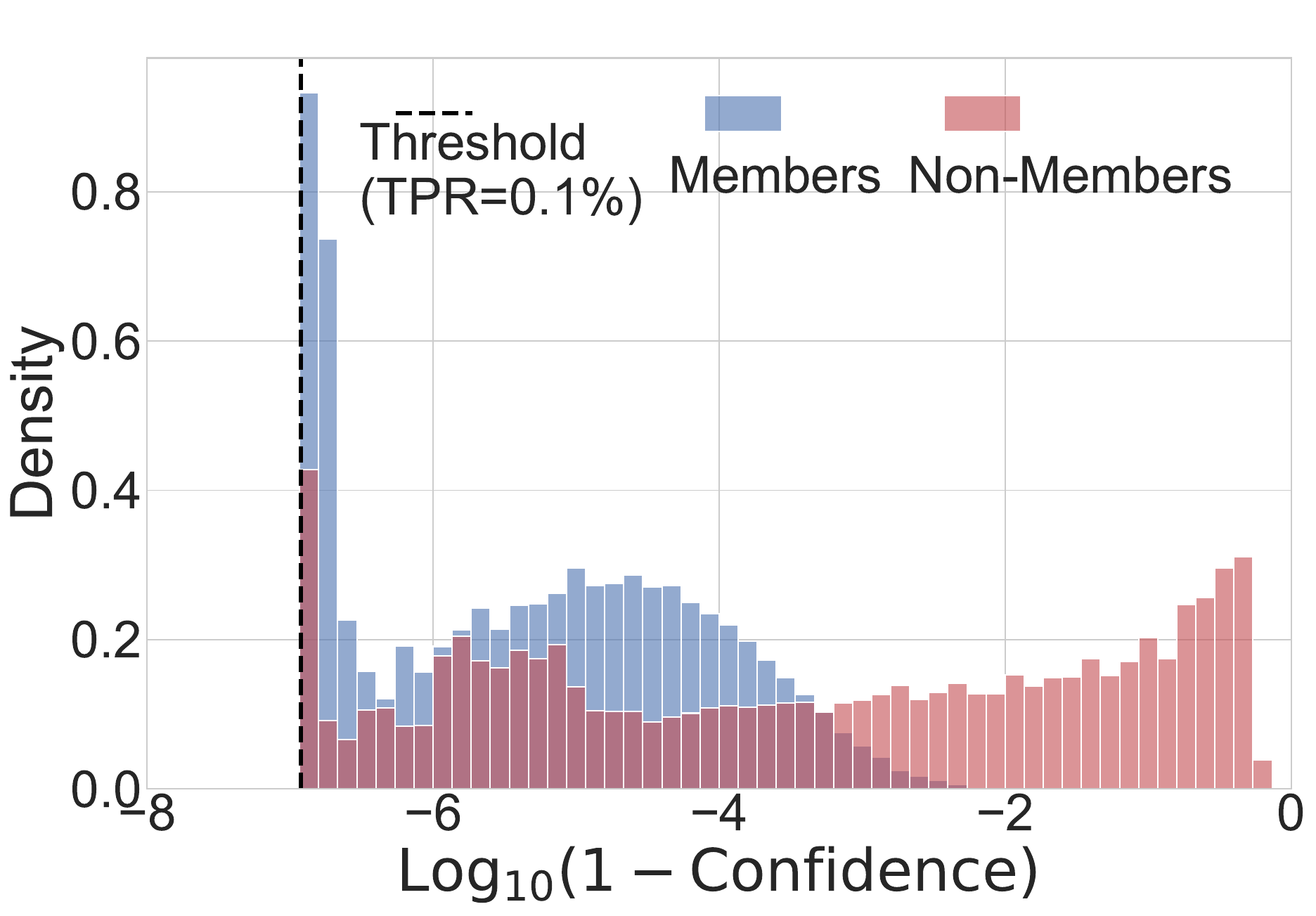}
\caption{Scratch (confidence)}
\end{subfigure}
\caption{Member and non-member confidence distributions on a logarithmic scale for CIFAR-10 with Swin under data selection and pre-training.}
\label{fig:loss_confidence_swin}
\end{figure}

\begin{figure}[t]
\centering
\begin{subfigure}{0.245\columnwidth}
\includegraphics[width=\columnwidth]{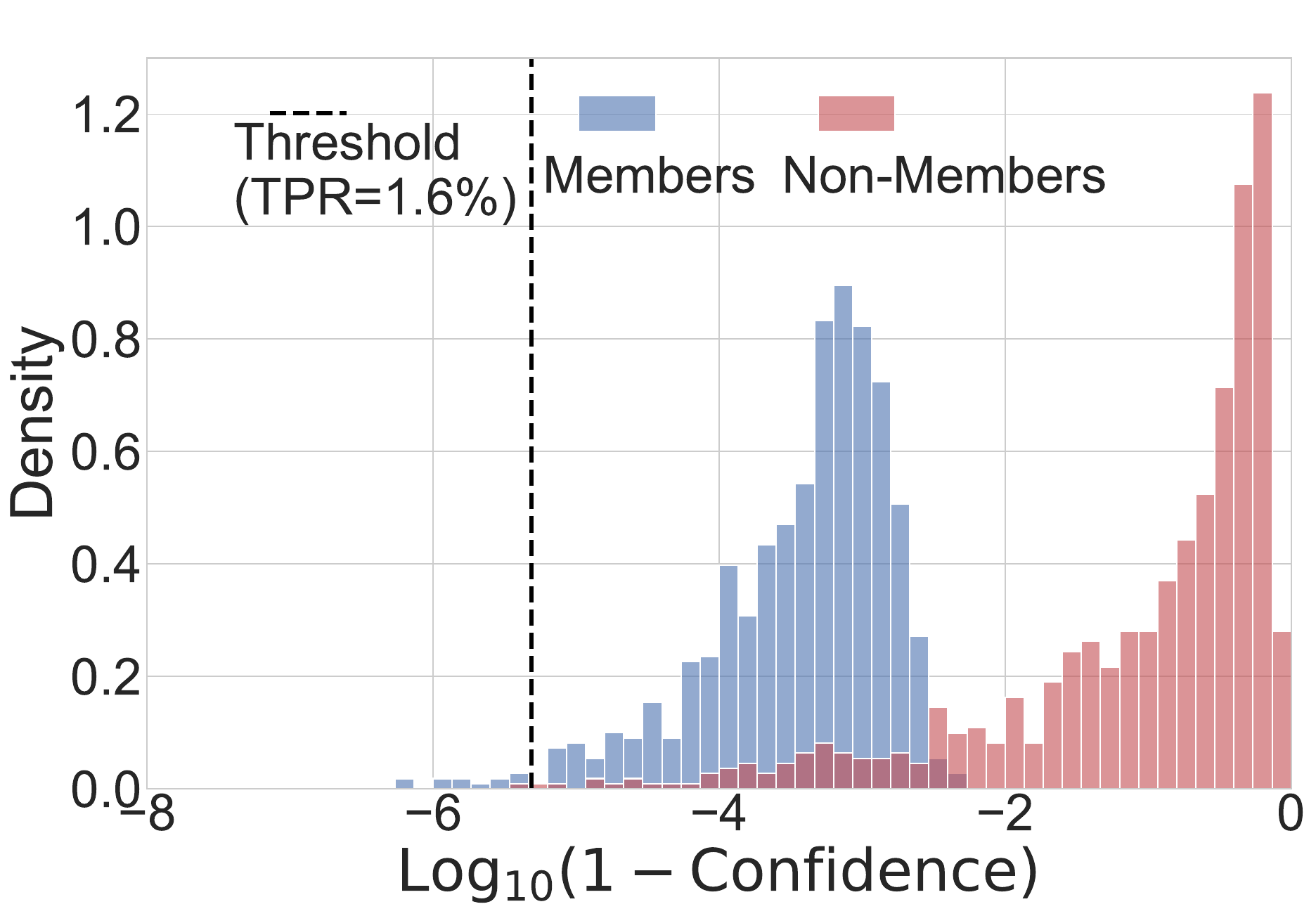}
\caption{High (confidence)}
\end{subfigure}
\begin{subfigure}{0.245\columnwidth}
\includegraphics[width=\columnwidth]{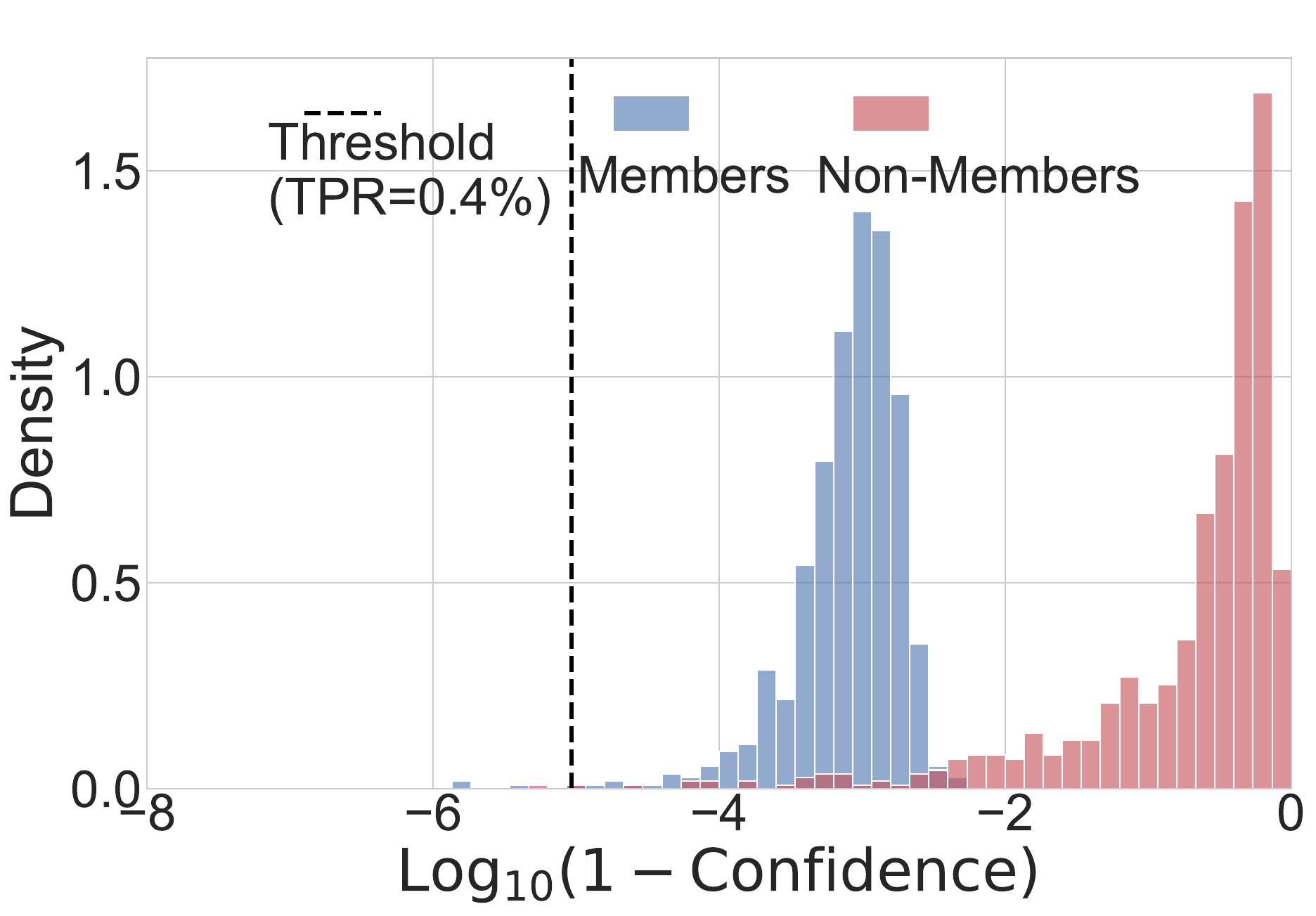}
\caption{Low (confidence)}
\end{subfigure}
\begin{subfigure}{0.245\columnwidth}
\includegraphics[width=\columnwidth]{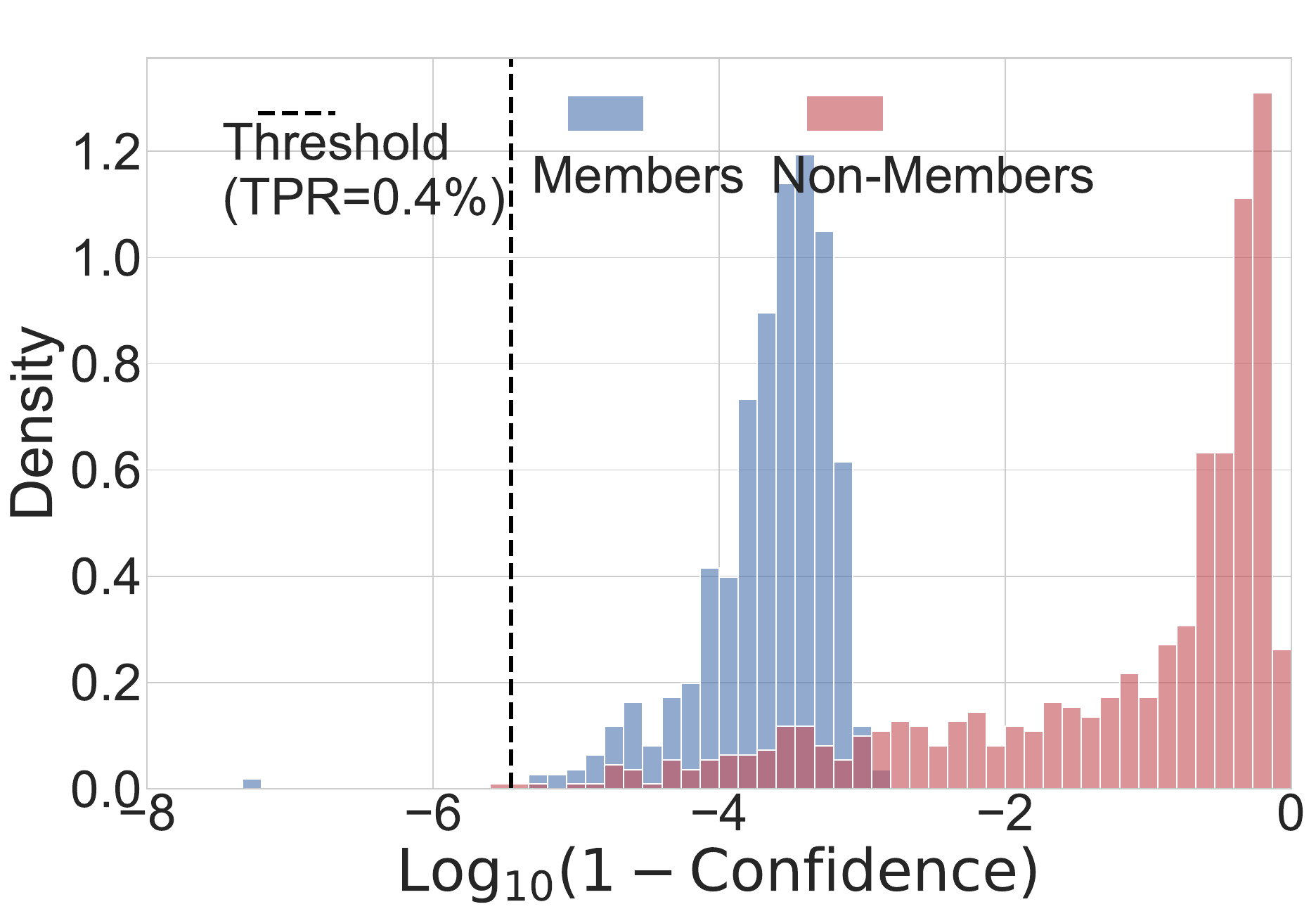}
\caption{Pre-train (confidence)}
\end{subfigure}
\begin{subfigure}{0.245\columnwidth}
\includegraphics[width=\columnwidth]{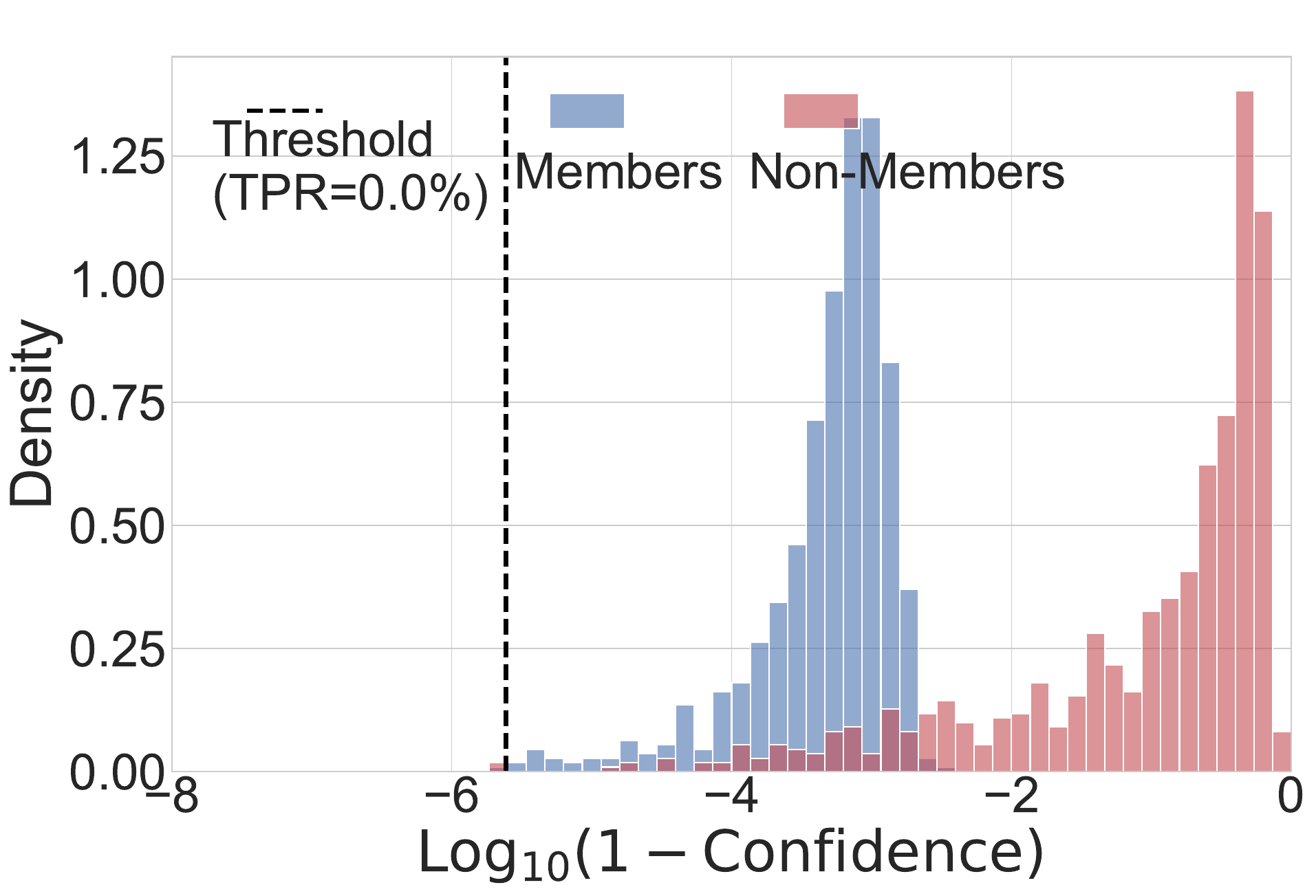}
\caption{Scratch (confidence)}
\end{subfigure}
\caption{Member and non-member confidence distributions on a logarithmic scale for PubFig83 under data selection and pre-training.}
\label{fig:loss_confidence_pubfig}
\end{figure}

\shortsection{Confidence Distributions}
Figures~\ref{fig:loss_confidence_cifar}, \ref{fig:loss_confidence_swin}, and \ref{fig:loss_confidence_pubfig} extend the confidence-distribution analysis in the main text to CIFAR-10 with ResNet-18, CIFAR-10 with Swin, and PubFig83, respectively. Each figure compares the utility-matched High- and Low-Shapley models as well as the pre-trained and scratch-trained models. Across these datasets and architectures, the efficient variants exhibit a stronger concentration toward extreme confidence values than their corresponding baselines. These distributional shifts are consistent with their generally higher ECE and confidence-based membership-inference leakage reported in Tables~\ref{tab:sharp_ece_pca_three_seeds}, \ref{tab:mia_shapley_three_seeds}, and \ref{tab:mis_ms_pretrain_three_seeds}. Together with the TinyImageNet visualization in Figure~\ref{fig:loss_confidence_Tiny}, these results show that the observed overconfidence is not restricted to a single dataset or architecture.

\begin{figure}[t]
\footnotesize
\centering
\begin{subfigure}{0.245\columnwidth}
\includegraphics[width=\columnwidth]{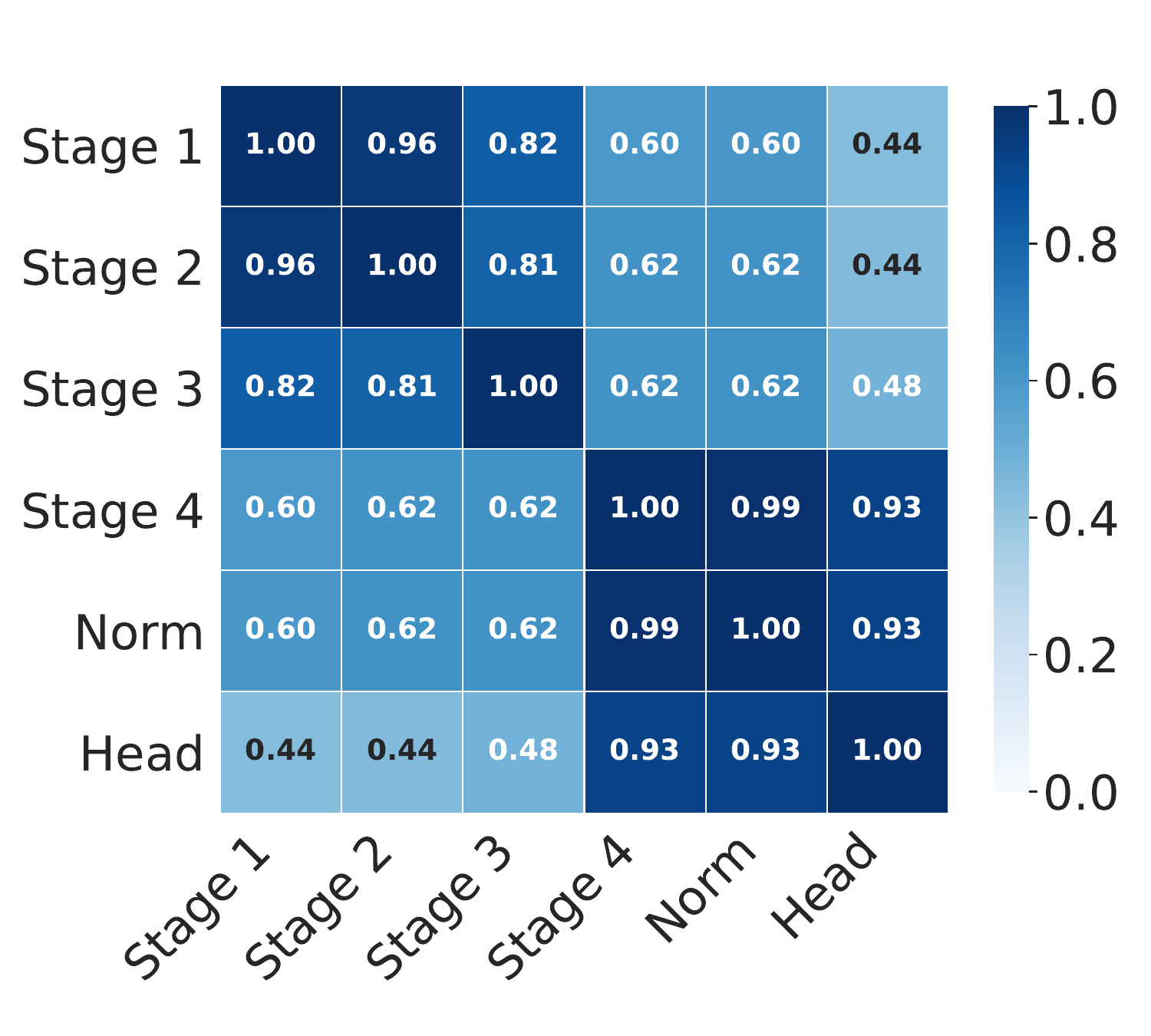}
\caption{{High (C-10 (swin))}}
\end{subfigure}
\begin{subfigure}{0.245\columnwidth}
\includegraphics[width=\columnwidth]{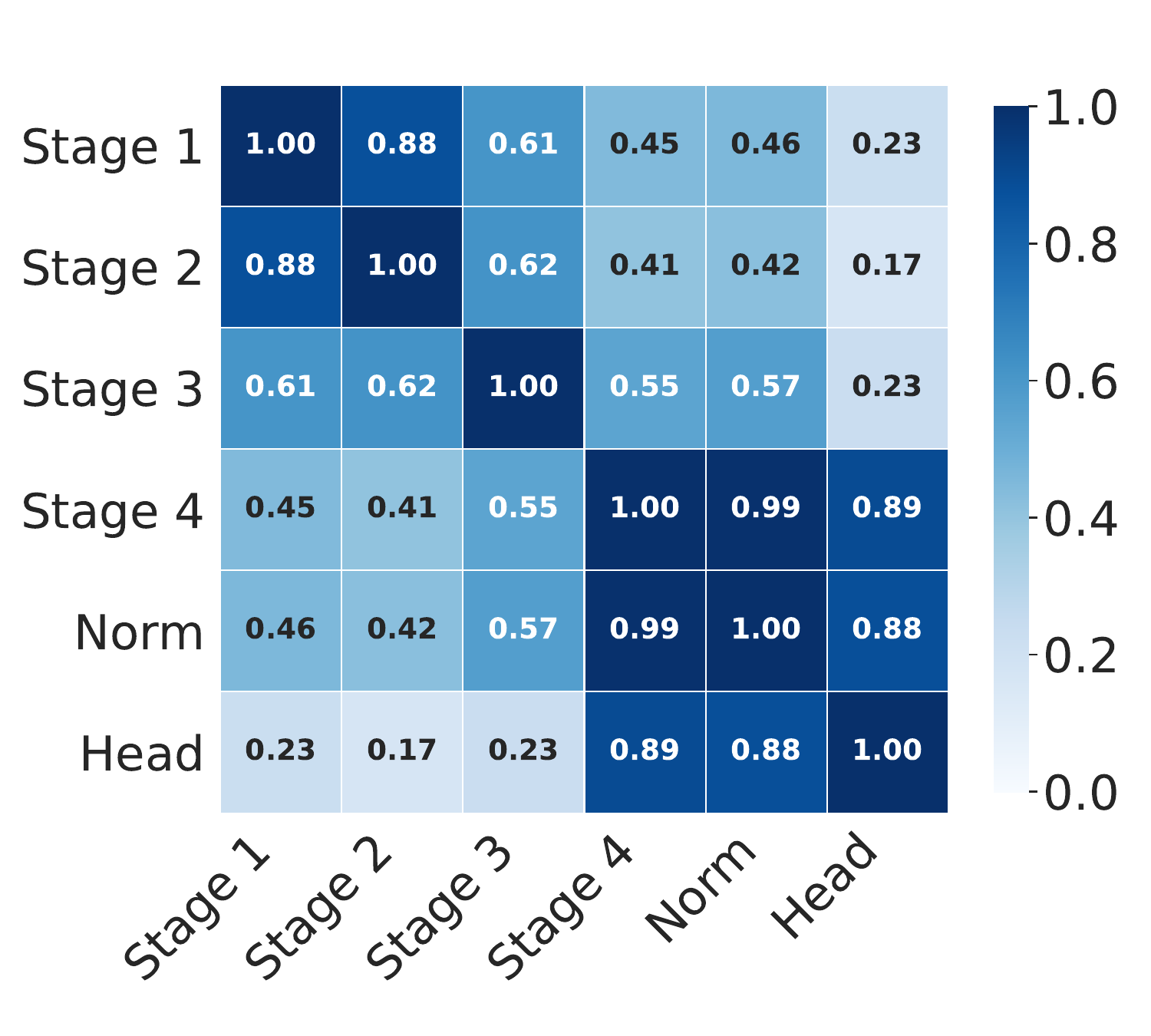}
\caption{Low (C-10 (swin))}
\end{subfigure}
\begin{subfigure}{0.245\columnwidth}
\includegraphics[width=\columnwidth]{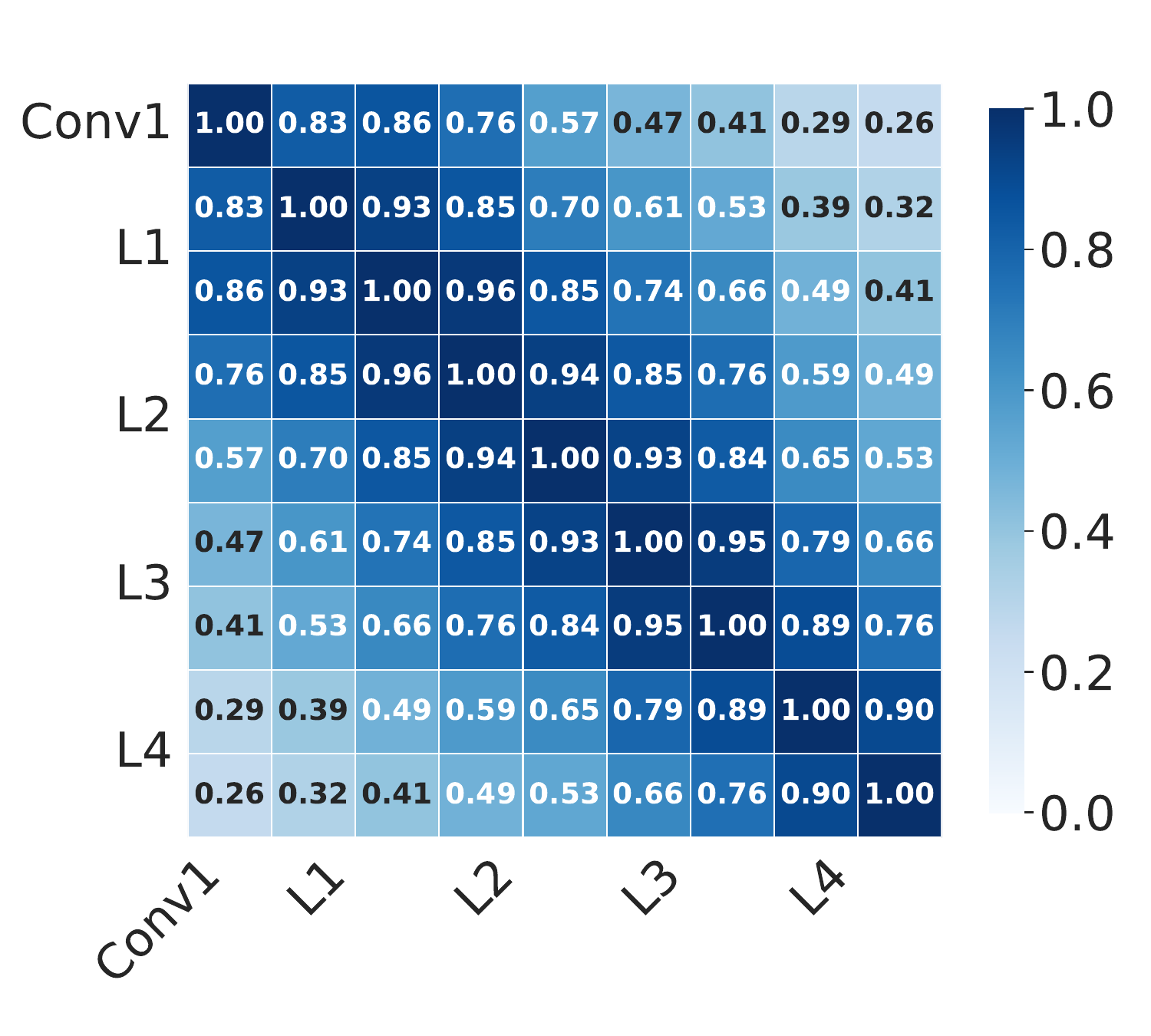}
\caption{High (PubFig83)}
\end{subfigure}
\begin{subfigure}{0.245\columnwidth}
\includegraphics[width=\columnwidth]{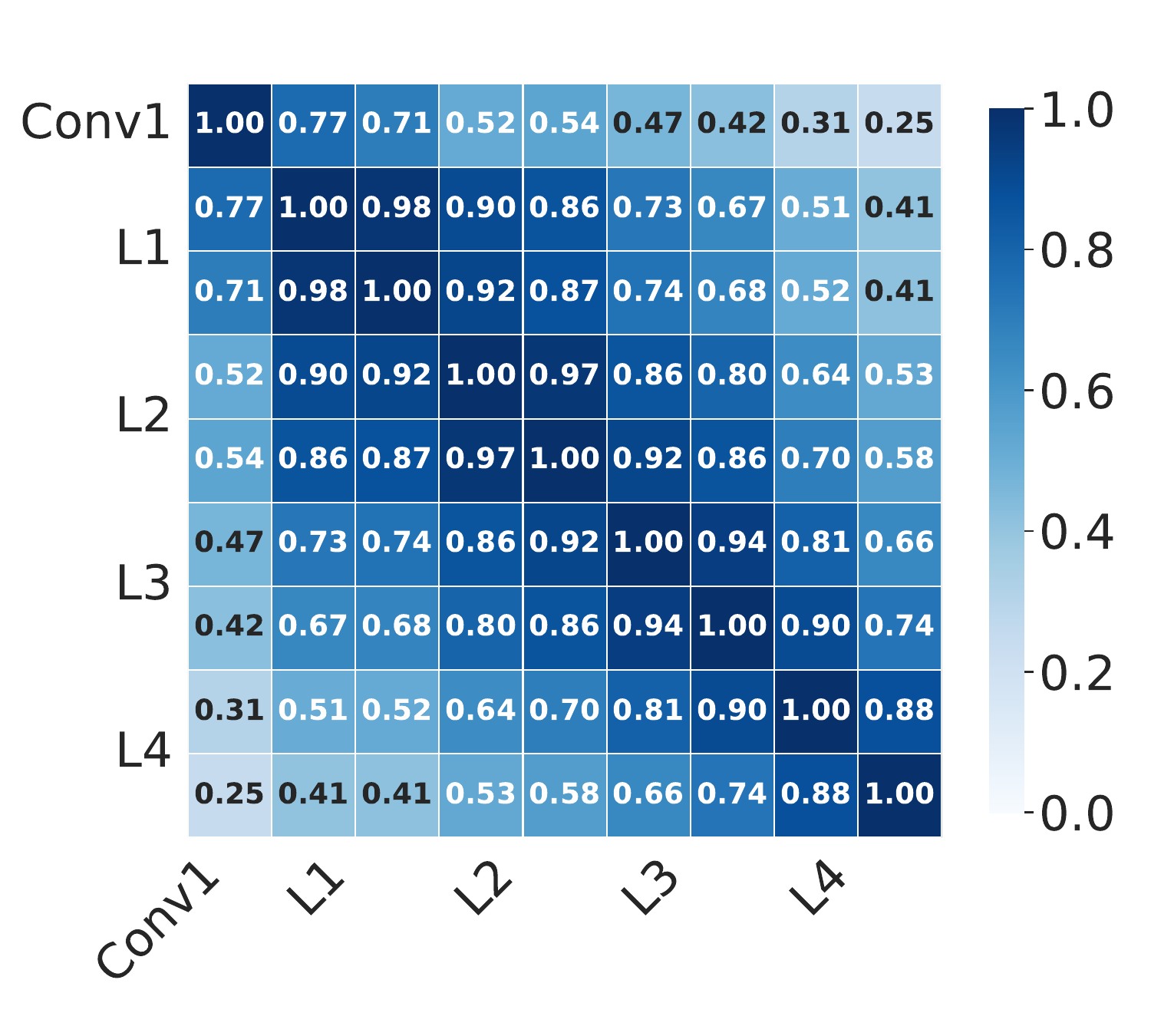}
\caption{Low (PubFig83)}
\end{subfigure}
\caption{Intra-CKA under Shapley-based data selection on CIFAR-10 (C-10) with Swin and PubFig83.}
\label{fig:cka_swin_pubfig}
\end{figure}

\shortsection{Representation Similarity}
Figure~\ref{fig:cka_swin_pubfig} presents additional Intra-CKA heatmaps for the High- and Low-Shapley models on CIFAR-10 with Swin and on PubFig83. Relative to the Low-Shapley controls, the High-Shapley models exhibit more extensive similarity across later layers, indicating that successive layers perform less differentiated transformations. This pattern complements the ResNet-18 and TinyImageNet results in Figure~\ref{fig:cka_cifar} and supports our interpretation that High-Shapley selection yields a less differentiated feature hierarchy.

\begin{figure}[t]
\centering
\begin{subfigure}{0.32\columnwidth}
\includegraphics[width=\columnwidth]{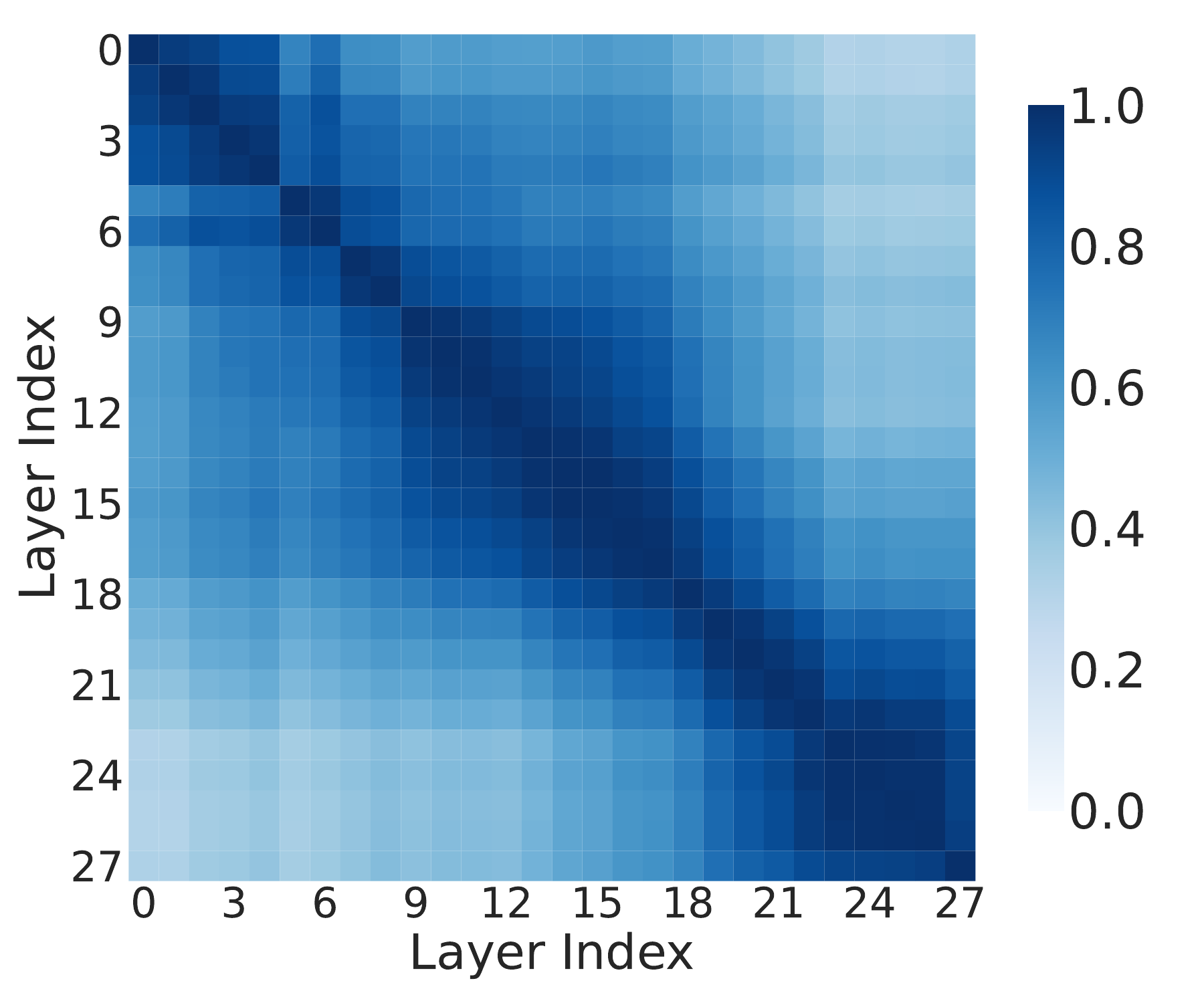}
\caption{Instruct}
\end{subfigure}
\begin{subfigure}{0.32\columnwidth}
\includegraphics[width=\columnwidth]{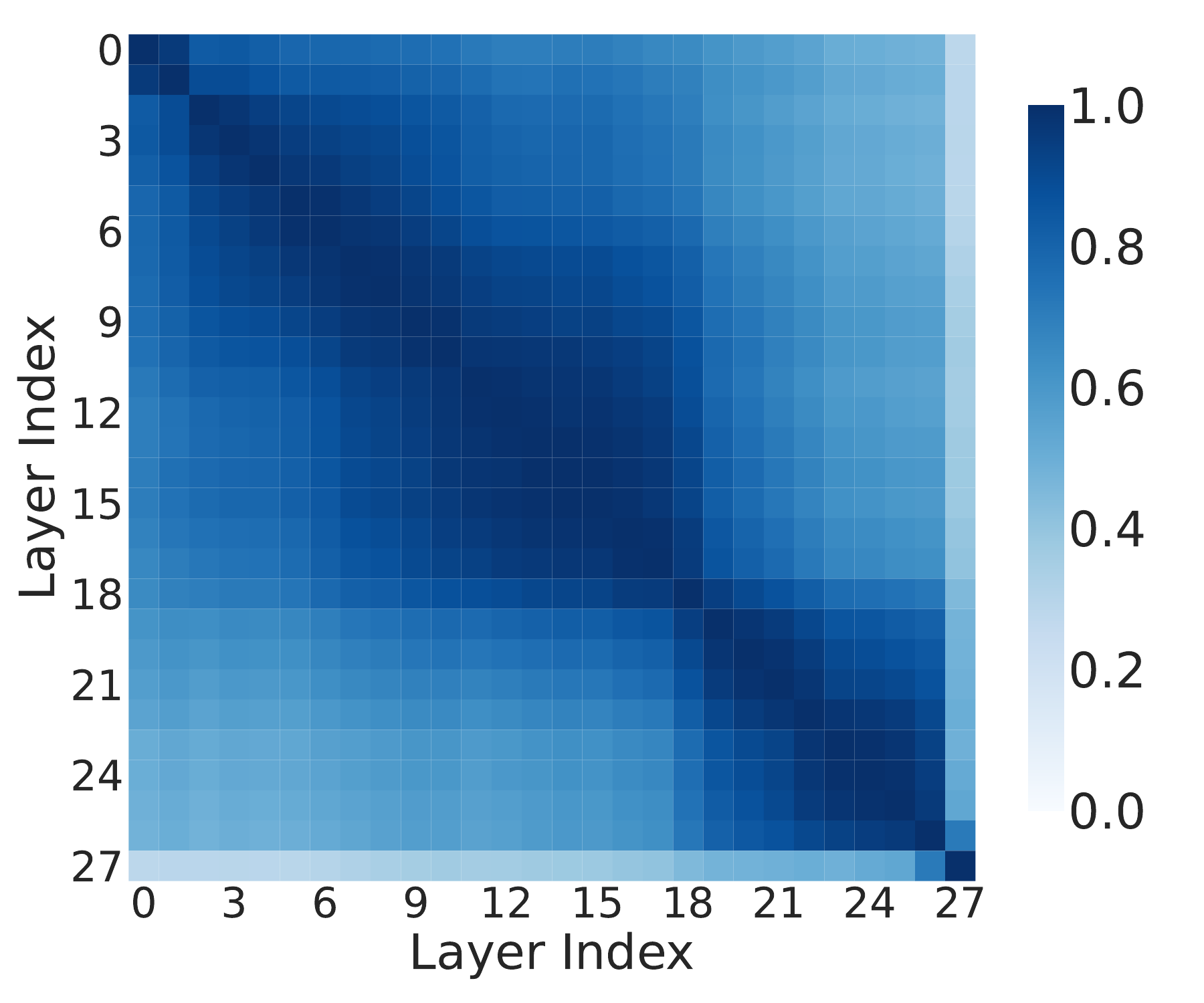}
\caption{SimpleRL}
\end{subfigure}
\begin{subfigure}{0.32\columnwidth}
\includegraphics[width=\columnwidth]{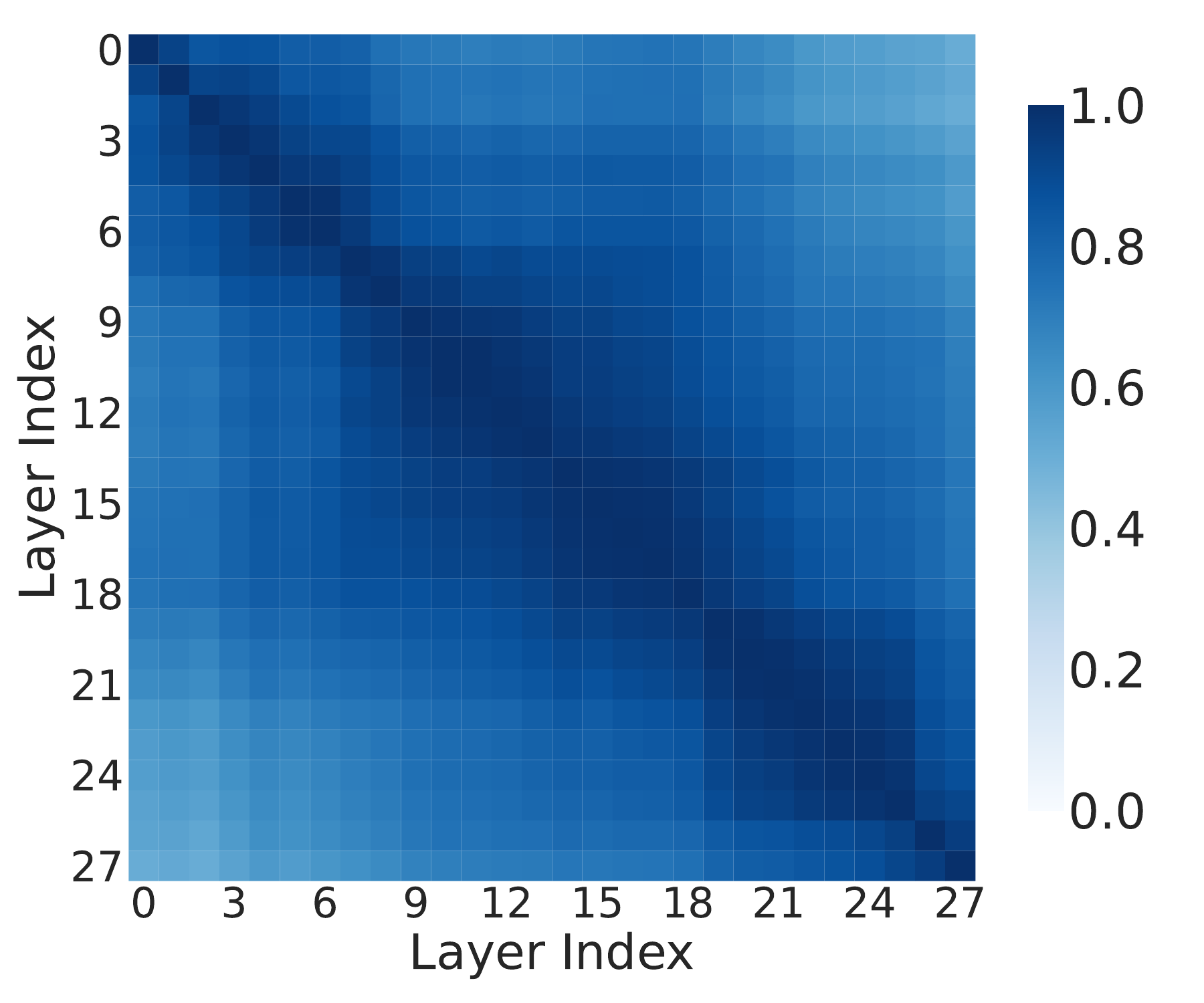}
\caption{Prime}
\end{subfigure}
\caption{Intra-CKA on 500 examples sampled from the MATH test set.}
\label{fig:cka_math}
\end{figure}

Figure~\ref{fig:cka_math} extends the language-model representation analysis from GSM8K in Figure~\ref{fig:cka_gsm8k} to 500 examples sampled from the full MATH test set. The SimpleRL and PRIME checkpoints again show stronger similarity between neighboring layers earlier in the network than the Instruct baseline. Observing the same qualitative structure on a second mathematical-reasoning benchmark provides additional evidence that the simplified-RL checkpoints develop more homogeneous internal representations rather than the effect being specific to GSM8K.

\shortsection{Activation Sparsity under Data Selection}
Table~\ref{tab:sparsity_shapley} reports the fraction of zero ReLU activations across five network stages for the utility-matched High- and Low-Shapley models. The High-Shapley models generally exhibit lower activation sparsity, particularly in the later stages, indicating that a larger proportion of their ReLU units remain active. This result shows that data selection changes the density of internal activations. Because activation density alone does not determine representational complexity or model-stealing difficulty, we treat it as a complementary descriptive observation rather than independent evidence of reduced functional redundancy.

\begin{table}[t]
\centering
\caption{Fraction of zero ReLU activations under data selection. Lower values indicate that a larger proportion of units are active.}
\setlength{\tabcolsep}{6pt}
\begin{tabular}{ll|ccccc}
\toprule
Dataset & Value & 0 & 1 & 2 & 3 & 4 \\
\midrule
\multirow{2}{*}{CIFAR-10} & Low & 0.512 & 0.509 & 0.561 & 0.677 & 0.781 \\
& High & 0.489 & 0.421 & 0.498 & 0.526 & 0.587 \\
\midrule
\multirow{2}{*}{TinyImageNet} & Low & 0.519 & 0.505 & 0.465 & 0.624 & 0.806 \\
& High & 0.531 & 0.440 & 0.467 & 0.506 & 0.684 \\
\bottomrule
\end{tabular}
\label{tab:sparsity_shapley}
\end{table}

\begin{figure}[t]
\centering
\begin{subfigure}{0.225\columnwidth}
\includegraphics[width=\columnwidth]{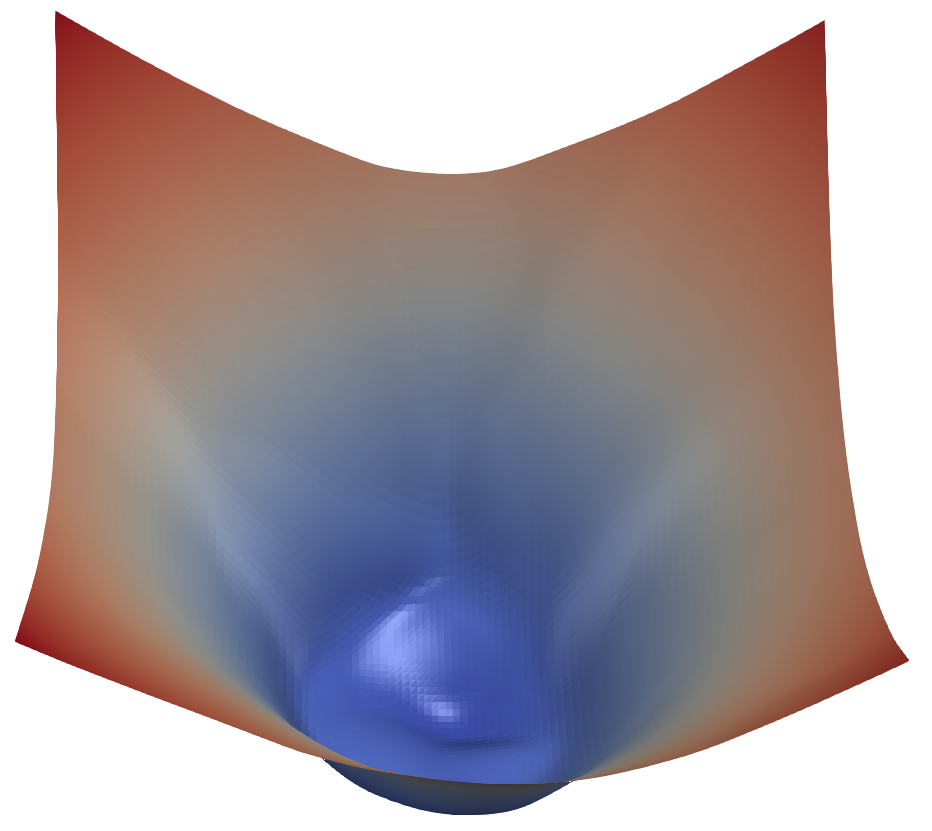}
\caption{High}
\end{subfigure}
\hfill
\begin{subfigure}{0.225\columnwidth}
\includegraphics[width=\columnwidth]{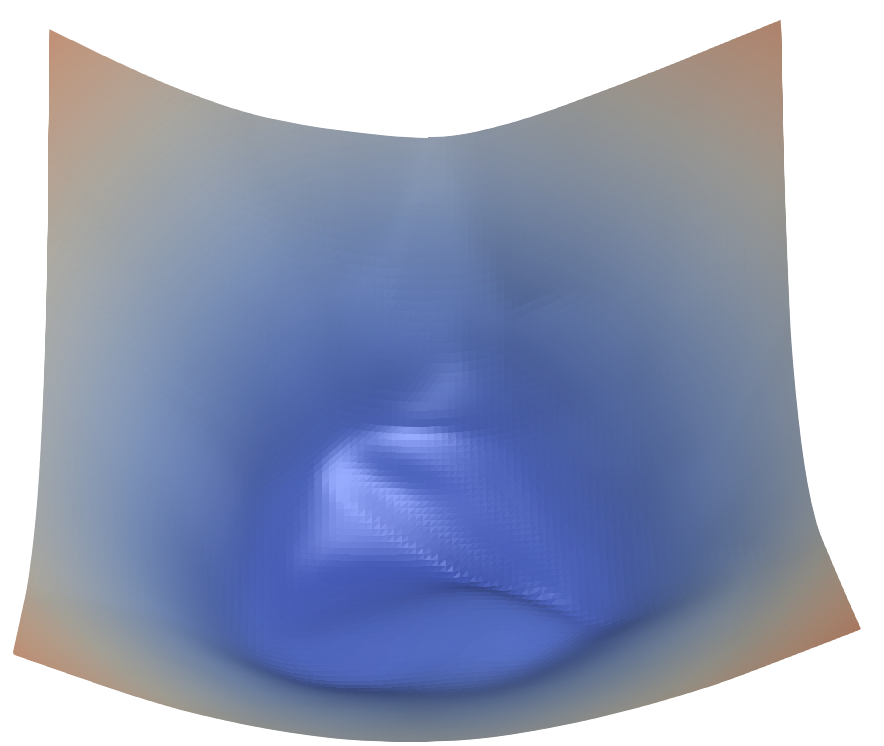}
\caption{Low}
\end{subfigure}
\hfill
\begin{subfigure}{0.225\columnwidth}
\includegraphics[width=\columnwidth]{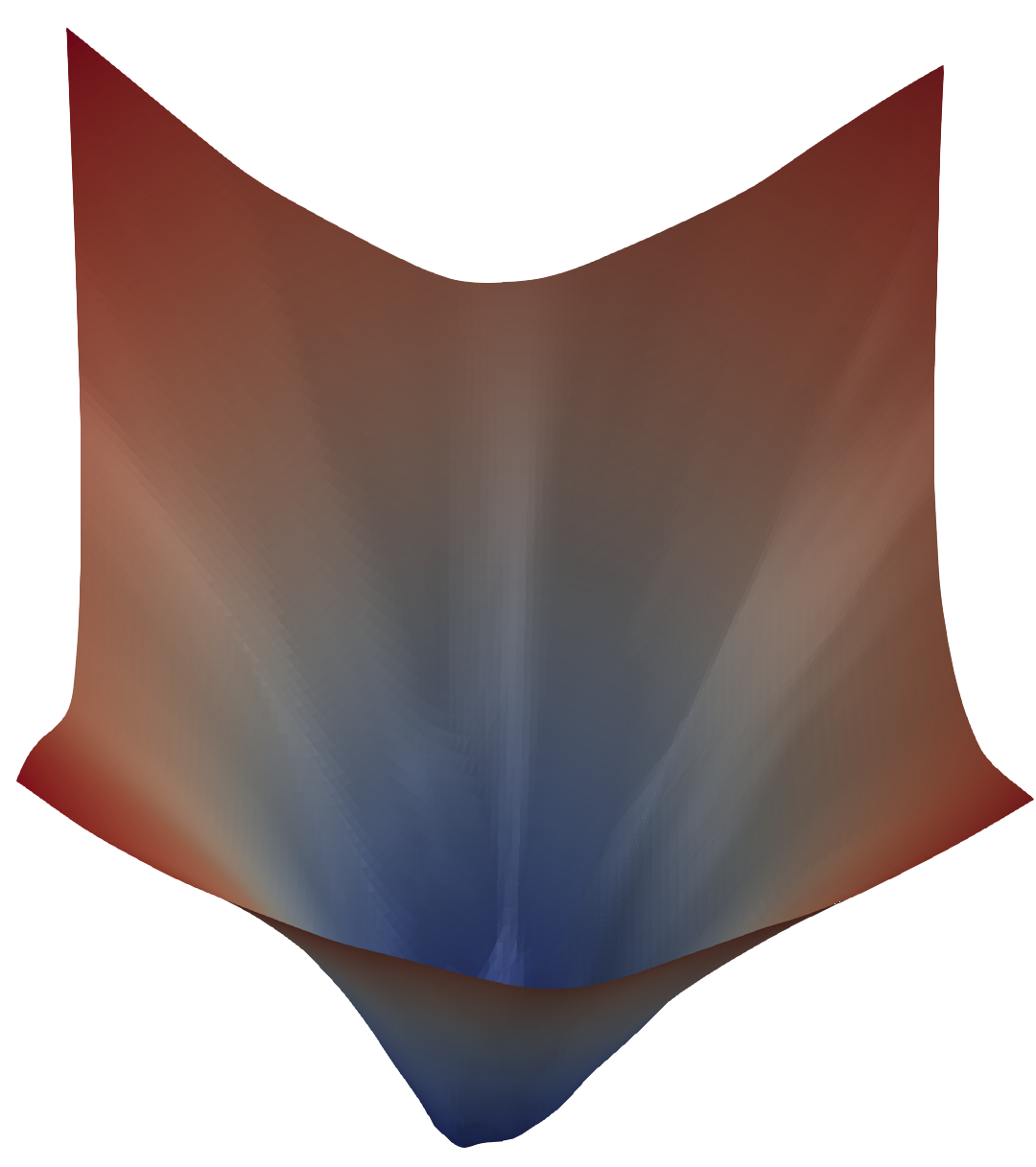}
\caption{Pre-train}
\end{subfigure}
\hfill
\begin{subfigure}{0.225\columnwidth}
\includegraphics[width=\columnwidth]{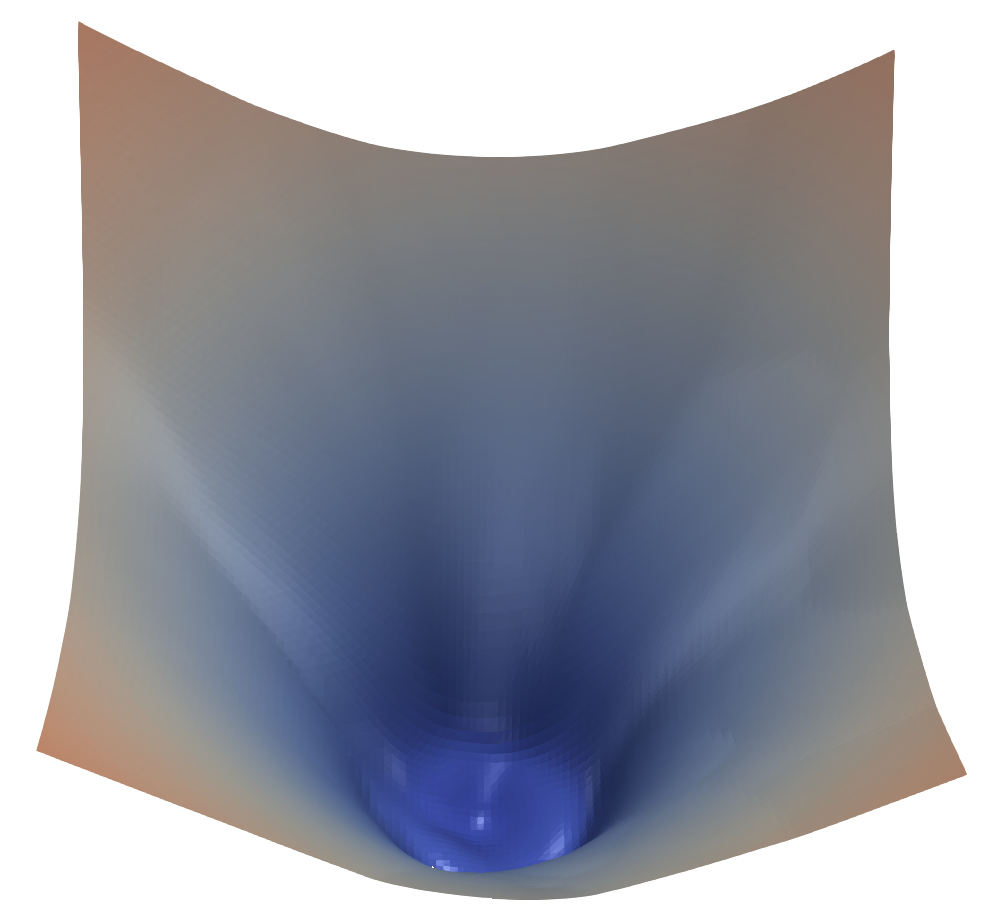}
\caption{Scratch}
\end{subfigure}
\caption{Loss landscape for TinyImageNet.}
\label{fig:landscape_tiny}
\end{figure}

\begin{figure}[t]
\centering
\begin{subfigure}{0.225\columnwidth}
\includegraphics[width=\columnwidth]{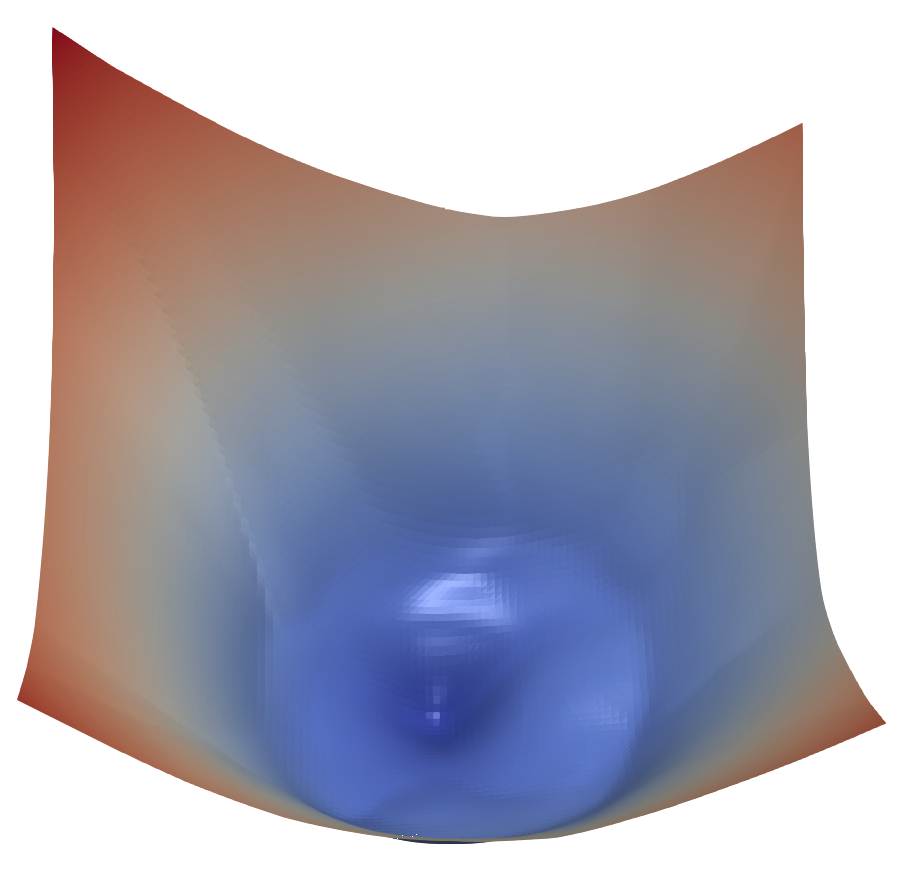}
\caption{High}
\end{subfigure}
\hfill
\begin{subfigure}{0.225\columnwidth}
\includegraphics[width=\columnwidth]{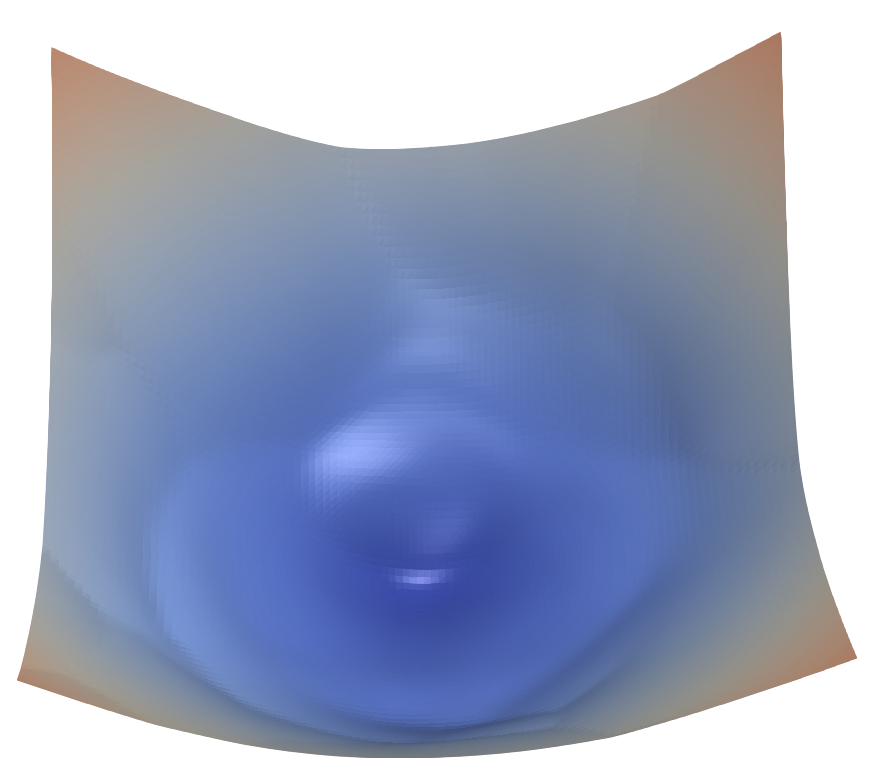}
\caption{Low}
\end{subfigure}
\hfill
\begin{subfigure}{0.225\columnwidth}
\includegraphics[width=\columnwidth]{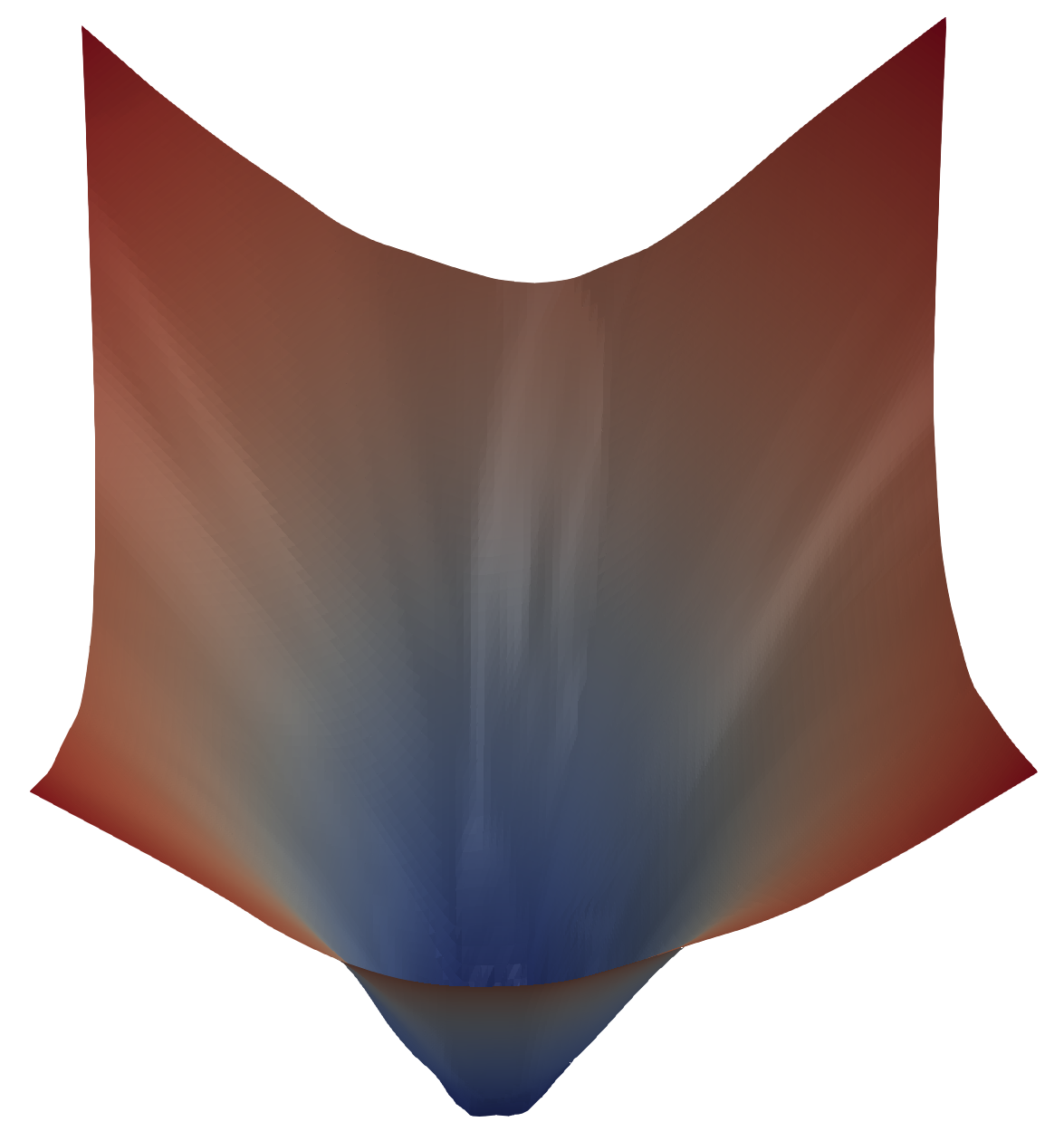}
\caption{Pre-train}
\end{subfigure}
\hfill
\begin{subfigure}{0.225\columnwidth}
\includegraphics[width=\columnwidth]{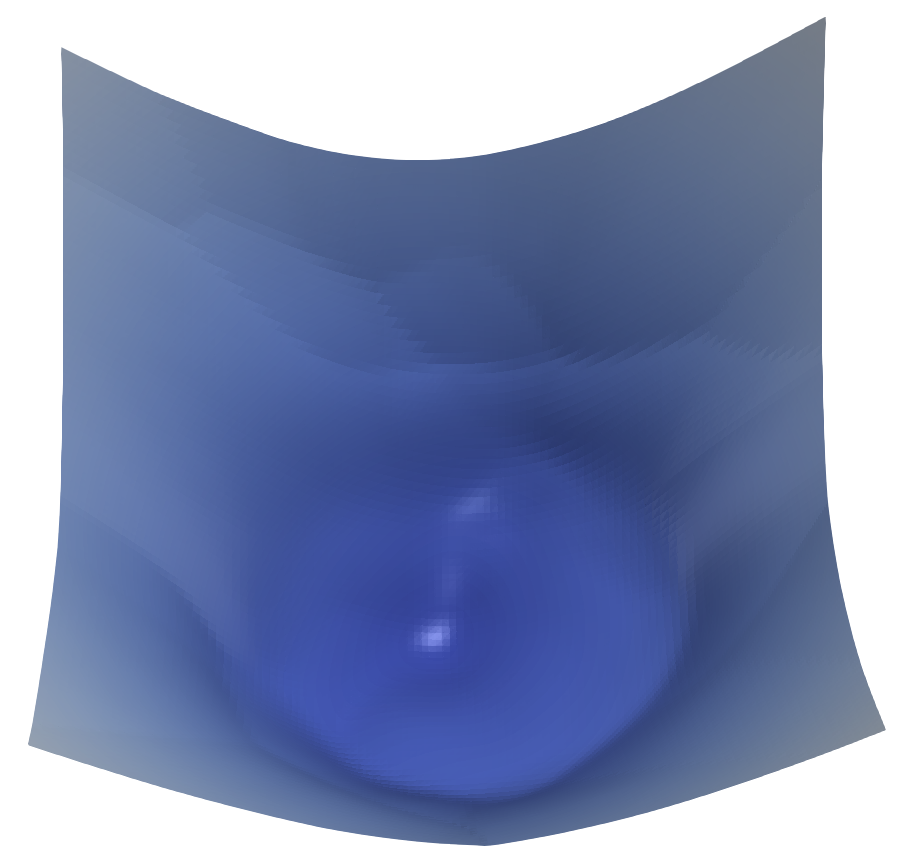}
\caption{Scratch}
\end{subfigure}
\caption{Loss landscape for PubFig83.}
\label{fig:landscape_pubfig}
\end{figure}

\shortsection{Loss Landscapes}
Figures~\ref{fig:landscape_tiny} and \ref{fig:landscape_pubfig} visualize the local loss geometry for TinyImageNet and PubFig83. For both data-selection comparisons, the High-Shapley models lie in sharper basins than their utility-matched Low-Shapley controls. Likewise, the pre-trained fine-tuned models exhibit sharper local geometry than the corresponding scratch-trained models. These visualizations agree with the quantitative sharpness measurements in Table~\ref{tab:sharp_ece_pca_three_seeds} and extend the CIFAR-10 landscapes in Figure~\ref{fig:landscape_cifar}. The consistent direction across datasets and training paradigms supports the conclusion that the evaluated efficiency mechanisms are associated with geometrically less stable solutions.

\section{Essential Experimental Settings}
\label{app:experimental_settings}
\begingroup
\small

\subsection{Vision Models}

\paragraph{Models and data.}
ResNet-18 is the default architecture for CIFAR-10, TinyImageNet, and PubFig83; we additionally evaluate Swin-Tiny on CIFAR-10. KNN-Shapley uses $K=6$. High- and Low-Shapley subsets contain the highest- and lowest-ranked examples, respectively, are disjoint, and have intentionally different sizes to match clean utility. In the pre-training comparison, the fine-tuned model uses public ImageNet-pretrained weights and a smaller task-specific subset, whereas the baseline is initialized randomly and trained from scratch on a larger subset. The backbone and classifier are optimized jointly. Table~\ref{tab:essential_vision_sizes} reports the training-set sizes after excluding reserved model-stealing queries. The PubFig83 test set contains 830 images (10 per class).

\begin{table}[h]
\centering
\caption{Training-set sizes in the vision experiments.}
\label{tab:essential_vision_sizes}
\setlength{\tabcolsep}{7pt}
\begin{tabular}{lrrrrr}
\toprule
Dataset / architecture & Full & High & Low & Fine-tuning & Scratch \\
\midrule
CIFAR-10 / ResNet-18 & 49,000 & 1,180 & 5,000 & 1,850 & 4,000 \\
CIFAR-10 / Swin-Tiny & 49,000 & 700 & 5,000 & 180 & 28,000 \\
TinyImageNet / ResNet-18 & 99,000 & 12,000 & 50,000 & 4,700 & 20,000 \\
PubFig83 / ResNet-18 & 6,970 & 1,950 & 4,000 & 2,800 & 3,680 \\
\bottomrule
\end{tabular}
\end{table}

\paragraph{Optimization and repeated runs.}
Vision models use cross-entropy and SGD with momentum $0.9$, weight decay $10^{-4}$, and batch size 128. Data-selection runs use initial learning rate $0.1$ for ResNet-18 and $10^{-3}$ for Swin-Tiny. They run for 300 epochs on CIFAR-10, 100 on TinyImageNet, and 200 on PubFig83, with tenfold learning-rate reductions at $75\%$ and $90\%$ of training. Scratch training uses the same architecture-specific initial learning rates. Fine-tuning/scratch comparisons run for 300 epochs on CIFAR-10 and 200 epochs on TinyImageNet and PubFig83 with the same relative milestones; fine-tuning uses initial learning rate $0.01$ for ResNet-18 and $10^{-4}$ for Swin-Tiny. Full-data controls use the same architecture, optimizer, learning-rate schedule, augmentation, epochs, and seeds as the corresponding High- and Low-Shapley experiments; only the training-set size changes. All reported three-seed vision results use seeds $\{0,1,2\}$ with fixed hyperparameters and report the sample standard deviation. Models within each comparison use corresponding seeds.

\paragraph{Membership inference and adversarial robustness.}
Vision MIA balances members from the model's actual training subset against non-members from its evaluation set and reports TPR at $0.1\%$ FPR. The three scalar scores are ground-truth confidence, prediction entropy, and modified prediction entropy. We evaluate untargeted $\ell_\infty$ PGD with cross-entropy, $\epsilon=8/255$, step size $\epsilon/20$, uniform random initialization, and 40 updates. For $N$ evaluated examples, the reported robustness score is
\[
R_{\mathrm{PGD}}=\frac{1}{N}\sum_{i=1}^{N}\sum_{t=0}^{39}
\mathbf{1}\{\widehat y_\theta(x_i^{(t)})=y_i\},
\]
where $\widehat y_\theta(x)$ is the model's predicted class. This is the average number of correctly classified iterates across the 40 pre-update checks; higher values indicate greater robustness. The procedure does not stop after the first misclassified iterate.

\paragraph{Model stealing.}
Attack images are drawn from the target distribution and excluded from target training. The data-selection experiments use 1,000 distinct images for CIFAR-10 and TinyImageNet and 500 for PubFig83; the fine-tuning/scratch experiments use 1,000. The surrogate matches the target architecture and minimizes mean $L_1$ distance between raw target and surrogate logits. We consider both random initialization and, for informed attacks, the same public pre-trained initialization used by the target. Surrogates train for 100 epochs with batch size 128. ResNet-18 uses SGD (learning rate $0.1$, momentum $0.9$, weight decay $5\times10^{-4}$, tenfold decay at epochs 75 and 90); Swin-Tiny uses AdamW (learning rate $5\times10^{-5}$, weight decay $0.05$, cosine decay). Reported query counts refer to distinct underlying images. Because augmented images are queried again during surrogate training rather than caching target logits, $Q$ images produce approximately $100Q$ image-level target evaluations.

\subsection{Structural Measurements}

\paragraph{Sharpness and calibration.}
We compute sharpness using Equation~\ref{equation:sharpness} over the complete test set. Perturbations start at zero and receive five normalized projected-gradient-ascent updates of size $2\times10^{-4}$. Each trainable parameter tensor is separately projected onto an $L_2$ ball of radius $10^{-3}$; original parameters are restored after each batch. Vision uses cross-entropy, while LLMs use shifted next-token cross-entropy. Vision ECE uses 15 equal-width confidence bins and maximum-class probability; LLM ECE uses ten bins and the mean probability of greedily generated tokens. LLM calibration is evaluated on GSM8K and the MATH 500 benchmark.

\paragraph{PCA, CKA, and pruning.}
PCA participation ratio is computed from centered penultimate-layer features over the complete vision evaluation set. Linear CKA compares the initial convolution and eight residual blocks of ResNet-18; the language-model analysis uses the final input-token representation from every Transformer layer. LLM CKA uses 500 test prompts per benchmark, sampled with seed 0 and shared across models; the MATH analysis draws these prompts from the full MATH test set. The maximum input length is 512. For pruning, we independently remove the smallest-magnitude 50\% of weights in every convolutional and linear layer, including classifiers, patch embeddings, attention projections, and MLPs. Biases, normalization and positional parameters are excluded, and accuracy is evaluated immediately without recovery fine-tuning.

\subsection{Language Models}

\paragraph{Checkpoints and membership inference.}
We evaluate \texttt{Qwen/Qwen2.5-Math-7B-Instruct}, \texttt{hkust-nlp/Qwen-2.5-7B-SimpleRL-Zoo}, and \texttt{PRIME-RL/Eurus-2-7B-PRIME}, together with the corresponding 32B Instruct and SimpleRL checkpoints. MIA candidates are drawn from difficulty levels 3--5 of MATH: the candidate-member source combines the training split and the first 4,500 test entries, while the remaining 500 test entries provide candidate non-members. Availability after filtering yields 339 examples per class. These source-defined groups provide the labels used throughout the LLM MIA evaluation. Answer likelihood is the mean conditional log probability of the reference solution; self-confidence is the mean conditional log probability of a greedily generated continuation; Min-$k$\% with $k=15$ averages the lowest-probability 15\% of retained reference-answer tokens. A balanced one-dimensional logistic regressor orients each score, and AUC and interpolated TPR are computed on this set. Reference sequences are truncated to 1,536 tokens; self-confidence uses prompts of at most 1,024 tokens and at most 512 generated tokens.

\paragraph{Robustness and jailbreak evaluation.}
Reasoning robustness is measured on RobustMath. Each question is placed in the user turn of a Qwen chat prompt with the system instruction to reason step by step and place the final answer in \texttt{\textbackslash boxed\{\}}. We greedily generate one response per question (temperature 0, top-$p=1$, seed 0) with at most 1,024 new tokens. Accuracy compares the extracted final answer with the reference using normalized numerical or symbolic equivalence; empty answers and grading timeouts are incorrect. AutoDAN-HGA is evaluated on 520 AdvBench prompts for at most 100 search iterations, using population size 16, elite fraction $0.05$, crossover probability $0.5$, five crossover points, mutation probability $0.01$, and seed 20. Responses use temperature $0.7$, top-$p=0.9$, and at most 64 new tokens. ASR uses all prompts, while average steps uses successful attacks and counts a success in the initial population as step zero.

\paragraph{IMDB adaptation and quantization.}
For the fine-tuning-stability evaluation, models undergo two epochs of causal language-model adaptation on the shuffled IMDB training split. We use 4-bit NF4 loading with bfloat16 computation and LoRA on the \texttt{q\_proj}, \texttt{k\_proj}, \texttt{v\_proj}, and \texttt{o\_proj} modules ($r=8$, $\alpha=16$, dropout $0.05$). The learning rate is $2\times10^{-4}$, per-device batch size is one, gradient accumulation is four, warmup is ten steps, and seed is 42. The adapter is merged into the bfloat16 base model before evaluation. The separate quantization sensitivity experiment loads the original checkpoints in 4-bit NF4 with float16 computation and uses matched decoding settings for the quantized and unquantized evaluations.

\endgroup
\end{document}